\documentclass[11pt]{article}

\usepackage[preprint]{acl}

\usepackage{times}
\usepackage{latexsym}

\usepackage[T1]{fontenc}

\usepackage[utf8]{inputenc}

\usepackage{microtype}

\usepackage{inconsolata}

\usepackage{graphicx}

\usepackage{CJKutf8}
\usepackage{tipa}
\usepackage{newtxmath}
\usepackage{amsmath}
\usepackage{makecell}
\usepackage{array, multirow}
\usepackage[table, x11names, dvipsnames, xcdraw]{xcolor}
\usepackage[normalem]{ulem}
\useunder{\uline}{\ul}{}
\usepackage{booktabs}
\usepackage{rotating}
\usepackage{multirow}
\usepackage{nicematrix}
\usepackage{color, colortbl}
\usepackage{hhline}
\usepackage{array}
\usepackage{pifont}
\usepackage{subfigure}     

\definecolor{Gray}{gray}{0.9}

\usepackage{enumitem}
\usepackage[super]{nth}
\usepackage{amsfonts}

\usepackage{pmboxdraw}

\usepackage{appendix}

\usepackage{cleveref}
\usepackage{hyperref}

\title{Beyond Surface Cues: Robust LLM-Generated Korean Text Detection
}

\author{
  \textbf{SungHo Kim}\textsuperscript{1},
  \textbf{SangKeun Lee}\textsuperscript{1,2}
\\
  \textsuperscript{1}Department of Artificial Intelligence, Korea University, Seoul, South Korea \\
  \textsuperscript{2}Department of Computer Science and Engineering, Korea University, Seoul, South Korea \\
\\
  \texttt{\{sungho3268, yalphy\}@korea.ac.kr}
}

\begin{document}

\begin{CJK}{UTF8}{mj}

\maketitle
\begin{abstract}

As large language models (LLMs) generate increasingly fluent text, the need for reliable detection of LLM-generated content has grown substantially. However, existing research remains heavily English-centric, leaving Korean largely underexplored. Moreover, current evaluations of Korean text detection often rely on superficial cues, leading to overly optimistic estimates of detection performance.
To address these limitations, we introduce \textsc{KoLing}, a bias-controlled adversarial benchmark comprising 50.9k instances that suppresses superficial shortcuts while preserving semantic content and separately evaluates LLM-generated and LLM-modified text detection. We further propose \textsc{KoLingProbe}, an interpretable detector grounded in theoretical linguistics that models Korean-specific properties across multiple linguistic levels.
Experiments show that \textsc{KoLingProbe} substantially outperforms strong baselines on all detection tasks while remaining robust to adversarial attacks. Our analyses further reveal that integrating complementary linguistic features across multiple levels yields both improved performance and interpretability.\footnote{The code will be released upon acceptance.}

\end{abstract}

\section{Introduction}

Detecting LLM-generated text has become increasingly important across settings like academic integrity, copyright protection, and synthetic-content monitoring~\cite{manipulation2022, pan-etal-2023-risk, orenstrakh-et-al-2024}. As LLMs produce increasingly fluent and human-like text, distinguishing machine-generated text from human writing grows more difficult~\cite{feng-etal-2024-bot, wan-etal-2025-social}, intensifying the need for reliable detection methods.
Most prior work in this domain, however, remains heavily English-centric, leaving Korean LLM-generated text detection largely underexplored despite its distinct linguistic characteristics~\cite{wu-etal-2025-survey, park-etal-2025-katfishnet}.

\newcommand{\capcolorbox}[2]{%
  {\setlength{\fboxsep}{0.5pt}\colorbox{#1}{#2}}%
}

\begin{table}[t]
\centering
\small
\resizebox{\columnwidth}{!}{
\begin{tabular}{@{}p{0.93\columnwidth}@{}}
    \toprule
    \textbf{Human-Written Text} \\
    \capcolorbox{violet!20}{성공율}이 낮다고 해도 따지고 보면 함께 해결할 \\ \capcolorbox{orange!20}{문제거든요}. \capcolorbox{blue!15}{근데} 서로 마음을 \capcolorbox{green!20}{알 길이 없}\capcolorbox{blue!15}{어요}. \\
    \midrule
    \textbf{LLM-Modified Text (GPT-4o)} \\
    \capcolorbox{violet!20}{성공률}이 낮더라도 함께 해결할 \capcolorbox{orange!20}{문제라고} \capcolorbox{orange!20}{생각} \\ \capcolorbox{orange!20}{합니다}. \capcolorbox{blue!15}{그러나} 마음을 \capcolorbox{green!20}{알 수 없}\capcolorbox{blue!15}{습니다}. \\
    \midrule
    \textbf{English Translated:} \textit{`Even if the success rate is low, it is a problem to solve together if you think about it. But there's no way to know how they really feel.'} \\
    \bottomrule
\end{tabular}
}
\caption{A human--LLM sentence pair with similar syntactic patterns. Colors indicate distinct linguistic features captured by \textsc{KoLingProbe}:
\capcolorbox{violet!20}{Phonological} (phonological alternation),
\capcolorbox{orange!20}{Morphological} (connective ending),
\capcolorbox{green!20}{Semantic} (idiom), and
\capcolorbox{blue!15}{Pragmatic} (discourse marker; polite ending) features.}
\vspace{-2px}
\label{tab:surface_vs_deep}
\end{table}

\begin{figure*}[t!]
  \centering
  \includegraphics[width=1.0\linewidth]{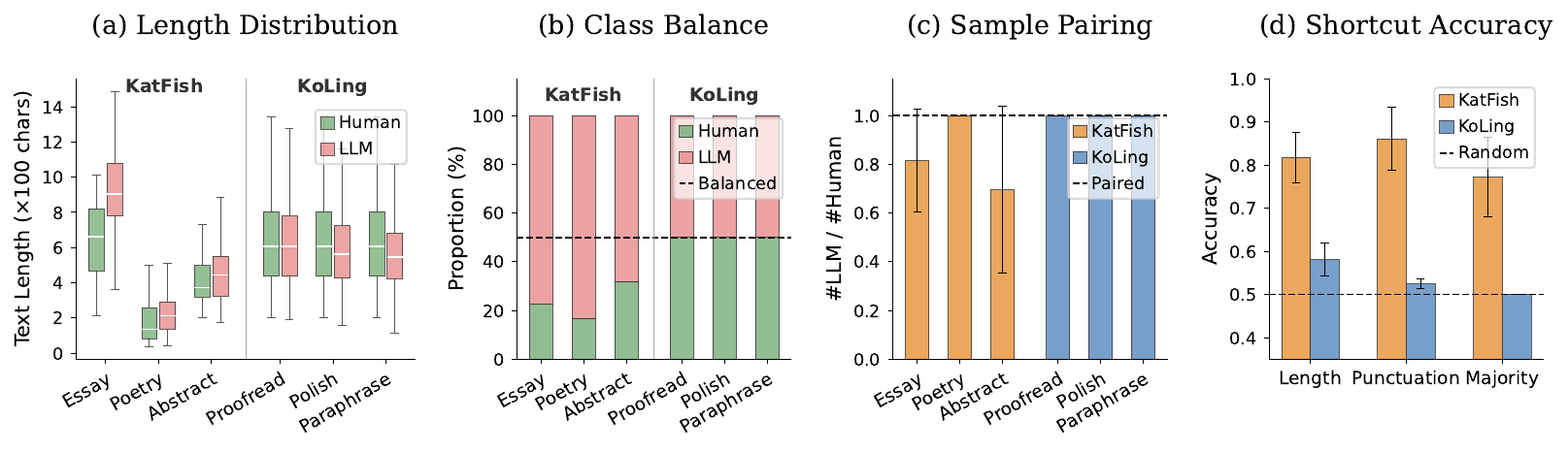}
  \caption{Comparison of dataset biases between KatFish and \textsc{KoLing}. (a) Length distributions; (b) Human vs. LLM class balance; (c) Paired human--LLM sample ratio per genre; (d) Shortcut classifier accuracy.}
  \label{fig:bias_comparison}
\end{figure*}

While a few studies have introduced Korean-specific benchmarks and detectors~\cite{park-etal-2025-katfishnet, go-etal-2025-xdac}, they rely primarily on simple syntactic patterns such as spacing and POS $n$-gram diversity, and thus fail when human and LLM texts share similar surface statistics (Table~\ref{tab:surface_vs_deep}). Distinguishing such cases instead requires modeling Korean-specific phenomena across multiple linguistic dimensions: phonological, morphological, syntactic, semantic, and pragmatic.

This limitation extends beyond detectors to the benchmarks used to evaluate them. As shown in Figure~\ref{fig:bias_comparison}, a representative Korean benchmark~\cite{park-etal-2025-katfishnet} retains substantial shortcut biases, length disparities, class imbalance, and few paired human--LLM samples, so even a simple classifier relying solely on comma-related syntactic cues performs deceptively well, without any deeper linguistic understanding. These artifacts cannot be reliably eliminated through post hoc balancing, since truncation or resampling distorts natural linguistic structure while leaving human and LLM texts unpaired~\cite{wu-etal-2025-survey}. Current evaluations thus substantially overestimate the detection capability of existing detectors.

We therefore rebuild both sides of the problem. We present a bias-controlled benchmark, \textsc{KoLing}, comprising 50.9k instances across three domains (essays, paper abstracts, and newspaper editorials) and the full authorship spectrum: human-written (HWT), LLM-modified (HLT), and LLM-generated (LGT). We also introduce an explainable detector \textsc{KoLingProbe}, integrating 47 features over the five levels of theoretical linguistics~\cite{Lyons_1968}: each level forms an independent sub-detector, and a lightweight meta-learner fuses their decisions, so every prediction is traceable to interpretable linguistic evidence. Our analyses unpack why this design works, which levels contribute, how they complement one another, and how far they transfer, and show that it stays reliable under typo noise that collapses fine-tuned detectors, entirely on CPU. Our contributions are as follows:
\begin{itemize}[leftmargin=1em, itemsep=0em]
    \item We introduce \textsc{KoLing}, a bias-controlled adversarial benchmark for Korean LLM-text detection covering three domains and the HWT/HLT/LGT authorship spectrum.
    \item We propose \textsc{KoLingProbe}, an explainable, theoretically grounded detector spanning five linguistic levels, achieving the best average binary and ternary performance entirely on CPU.
    \item Through extensive analyses, we show that the five linguistic levels contribute complementary, cross-domain-portable evidence, a design that also degrades gracefully under typo noise where fine-tuned detectors collapse.
\end{itemize}

\section{Detection Benchmark: \textsc{KoLing}}
\label{sec:koling}

\textsc{KoLing} is a \textbf{Ko}rean \textbf{Ling}uistic controlled adversarial benchmark designed around three principles: diverse source domains, controlled text generation, and systematic shortcut control.

\subsection{Data Generation}

Existing detection benchmarks primarily formulate detection as a binary human-vs.-LLM task~\cite{hc3-guo2023close, park-etal-2025-katfishnet}. In practice, however, LLMs are more often used to revise existing human writing than to generate text entirely from scratch~\cite{detectrlx2026}. To reflect this real-world usage, we construct three text categories representing increasing levels of LLM involvement.

\begin{figure}[t!]
  \centering
  \includegraphics[width=1.0\linewidth]{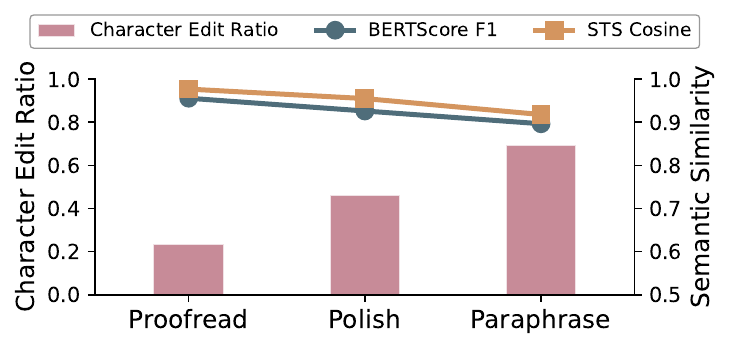}
  \caption{Surface variation (bars) and semantic preservation (lines) across \textsc{KoLing} modification levels.}
  \label{fig:modification_level}
\end{figure}

\begin{itemize}[leftmargin=15pt, itemsep=3pt, topsep=7pt]
    \item \textbf{HWT: Human-Written Text.} Original texts written without LLM assistance, collected from three genres spanning a broad domain: essays, academic paper abstracts, and newspaper editorials, from informal student writing to professionally edited prose (details in Appendix~\ref{app:source_data_details}).

    \item \textbf{HLT: LLM-Modified Text.} LLM modifies human-written texts rather than generating from scratch, suppressing generation artifacts while preserving human syntactic patterns, at three levels of increasing LLM intervention: \textbf{Proofread} (correct errors and awkward expressions, preserving style and structure), \textbf{Polish} (improve fluency and clarity), and \textbf{Paraphrase} (rewrite structures and vocabulary, preserving meaning). Figure~\ref{fig:modification_level} shows that the character edit ratio increases monotonically with modification level, whereas semantic similarity remains close to 0.90 even for Paraphrase; surface form varies, meaning is preserved.

    \item \textbf{LGT: LLM-Generated Text from Scratch.} LGT follows a single principle: the LLM receives the same input as the original human author, the full paper body for abstracts, the title for editorials, and the essay prompt, reading passage, and grade level (as a student persona) for essays. This prevents topical mismatch from becoming a shortcut; the two formal domains additionally enforce the human register (plain ``-다'' declarative). All prompts are in Appendix~\ref{app:prompts}.
\end{itemize}

\paragraph{Generators and splits.}

Each domain is populated using the same five LLMs: GPT-4o~\cite{gpt-4o}, Claude-Sonnet-4.6~\cite{claude-4.6}, Solar-Pro3~\cite{solar-pro3}, Qwen2.5-32B~\cite{qwen2025qwen25technicalreport}, and Mistral-Small-24B~\cite{mistral-3.1}, with human texts split 8:2 into train and test. Training-based detectors are trained on Claude-Sonnet-4.6 outputs, selected as a representative frontier model, and evaluated against all five generators (one in-distribution, four out-of-distribution); switching the training generator to GPT-4o or Qwen2.5-32B yields the same trends and rankings, confirming that this choice does not drive our findings (Appendix~\ref{sec:app_trainsource}).

\subsection{Shortcut Control}
As illustrated in Figure~\ref{fig:bias_comparison}, existing benchmarks exhibit exploitable surface biases in text length, class balance, and punctuation. We mitigate these during generation via prompts that constrain output length and encourage preservation of human syntactic cues, but prompt-level control alone cannot eliminate residual biases.

We therefore normalize punctuation distributions against the corresponding human texts and probabilistically subsample each class to match class-conditional length distributions within every split. These procedures substantially reduce residual length and punctuation cues, although they intentionally produce imbalanced class proportions. The resulting composition of \textsc{KoLing} is summarized in Table~\ref{tab:generated_dataset}.

\begin{figure*}[t!]
  \centering
  \includegraphics[width=.98\linewidth]{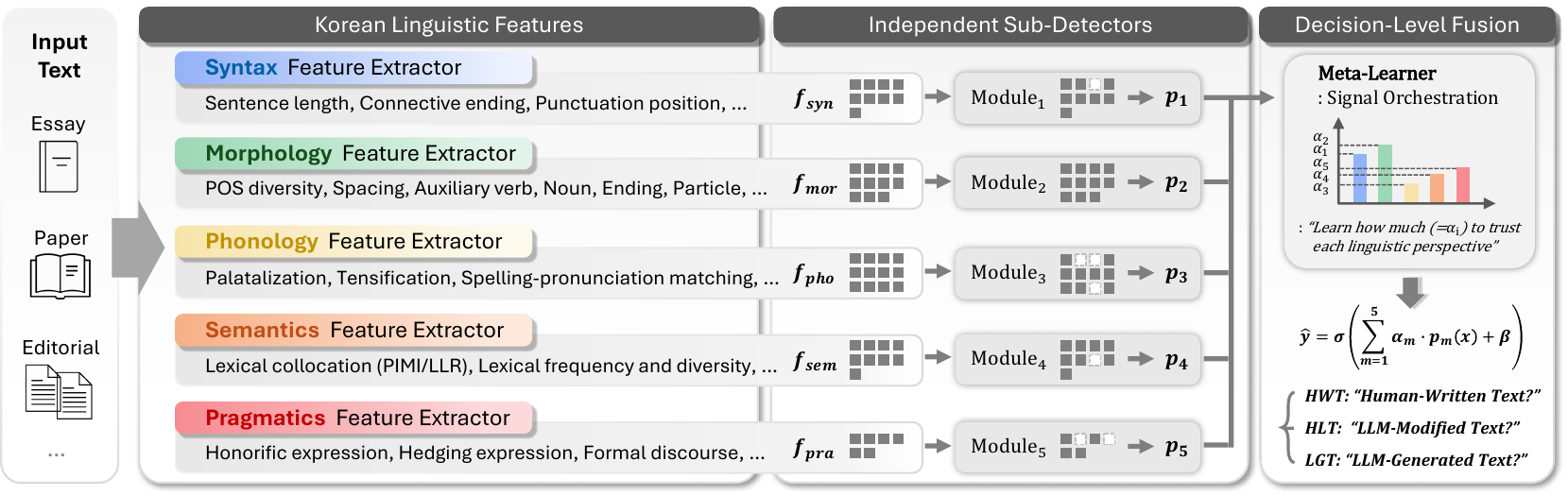}
  \caption{Overview of \textsc{KoLingProbe}: five independent linguistic sub-detectors each produce a detection probability, which a meta-learner fuses into the final decision.}
  \label{fig:detector_framework}
  \vspace{-7px}
\end{figure*}

\section{Detection Method: \textsc{KoLingProbe}}

\textsc{KoLingProbe} is a lightweight, CPU-only detector that organizes 47 Korean-specific features into five theoretically motivated linguistic levels (\S~\ref{sec:linguistic_sub-detectors}), trains a separate sub-detector for each, and fuses their predictions only at the decision level (\S~\ref{sub-detector_integration}; Figure~\ref{fig:detector_framework}). This design preserves interpretability while limiting interference among heterogeneous feature types.

\begin{table}[t!]
\centering
\resizebox{\columnwidth}{!}{
\begin{tabular}{l|rr|rr|rr|r}
    \toprule
    & \multicolumn{2}{c|}{\textbf{HLT-vs-HWT}} & \multicolumn{2}{c|}{\textbf{LGT-vs-HWT}} & \multicolumn{2}{c|}{\textbf{Ternary}} & \\
    \textbf{Domain} & Train & Test & Train & Test & Train & Test & \textbf{Total} \\
    \midrule
    Essay & 10220 & 3900 & 5566 & 1397 & 13056 & 4304 & 38443 \\
    Paper Abs. & 1089 & 262 & 338 & 79 & 1068 & 194 & 3030 \\
    Editorial & 3301 & 852 & 1006 & 202 & 3407 & 661 & 9429 \\
    \midrule
    \textbf{Total} & \textbf{14610} & \textbf{5014} & \textbf{6910} & \textbf{1678} & \textbf{17531} & \textbf{5159} & \textbf{50902} \\
    \bottomrule
\end{tabular}}
\caption{Statistics of the \textsc{KoLing} benchmark: evaluation instances per domain and task after shortcut control. The last column sums each row.}
\label{tab:generated_dataset}
\end{table}

\subsection{Linguistic Features}
\label{sec:linguistic_sub-detectors}

To move beyond the surface cues of prior Korean detectors~\cite{park-etal-2025-katfishnet, go-etal-2025-xdac}, we ground our features in the canonical five-level taxonomy of theoretical linguistics: phonology, morphology, syntax, semantics, and pragmatics~\cite{Lyons_1968, kim-etal-2025-polishing}, which decomposes linguistic phenomena systematically rather than through isolated stylistic heuristics. Drawing on Korean linguistic structure~\cite{sohn2001korean, Cho_2022} and error patterns of native usage~\cite{koo2022k, kim-etal-2024-kogec}, we define 47 Korean-specific features across the five levels (full descriptions in Appendix~\ref{app:feature_exploration}).

\paragraph{Phonology (12d).}
Korean spelling frequently diverges from pronunciation due to morphophonological rules. We use a G2P converter~\cite{park2019g2pk} with pre-compiled pronunciation dictionaries\footnote{Built from a large Korean corpus; see Appendix~\ref{app:feature_exploration}.} to count instances of five phonological rules: palatalization, tensification, liquid assimilation, nasal assimilation, and initial law (6d). Human writers produce more spelling-pronunciation mismatches (e.g., \begin{CJK}{UTF8}{mj}성공율\end{CJK}\textsuperscript{*} vs.\ \begin{CJK}{UTF8}{mj}성공률\end{CJK}),\footnote{\;`*' marks ungrammaticality.} while LLMs are orthographically conservative. We additionally extract six syllable structure features (e.g., CVC ratio, final consonant distributions) from the onset-nucleus-coda decomposition.

\paragraph{Morphology (11d).}
Korean is agglutinative: a single \emph{eojeol} (spacing unit) often packs multiple morphemes. Using the Kiwi analyzer~\cite{kiwi}, we extract spacing features (5d), POS-sequence diversity, dependent-noun and auxiliary-verb spacing ratios, and morphological distribution features (6d): josa entropy, rare josa ratio, eomi diversity, connective ending ratio, morphemes per eojeol, and function-to-content ratio. LLMs tend to regularize spacing conventions and underutilize rare case markers and verbal endings.

\paragraph{Syntax (9d).}
Our syntax sub-detector combines (i) the comma-usage features (5d) of KatFishNet~\cite{park-etal-2025-katfishnet}, including density, relative position, inclusion rate, segment length, and POS-bigram diversity around commas, and (ii) four new syntactic structure features: sentence-length statistics (mean, standard deviation, binned entropy) and clause density. LLMs tend to produce more uniform sentence lengths with lower entropy than human writers.

\paragraph{Semantics (9d).}
We build a dictionary of noun--predicate collocation statistics from Korean Wikipedia and score noun--josa--predicate triples, extracted via dependency parsing~\cite{qi-etal-2020-stanza}, with PMI and LLR statistics (5d). Human writers tend to favor more idiomatic collocations, whereas LLMs often produce more generic combinations.
Word-frequency features (4d) add log-frequency statistics and rare/hapax-word ratios.

\paragraph{Pragmatics (6d).}
We focus on honorifics: Korean's rich speech-level system, which LLMs often collapse into a single formal register. We classify sentences into four speech levels (formal polite, informal polite, plain/written, and casual) and compute the formal ratio and entropy (2d). We further measure hedging expression density (1d), discourse-marker density and formality ratio (2d), and sentence-ending diversity (1d). LLMs tend toward uniform formality, overuse formal discourse markers, and produce repetitive sentence endings.

\begin{table*}[t]
\centering
\resizebox{\textwidth}{!}{
\setlength{\tabcolsep}{2.5pt}
\begin{tabular}{l|ccc|ccc|ccc|ccc|ccc|ccc|ccc|ccc|ccc}
    \toprule
    \multirow{3}{*}{\textbf{Method}} & \multicolumn{12}{c|}{\textbf{HLT vs.\ HWT}} & \multicolumn{12}{c|}{\textbf{LGT vs.\ HWT}} & \multicolumn{3}{c}{\multirow{2}{*}{\textbf{Avg.}}} \\
    \cmidrule(lr){2-13} \cmidrule(lr){14-25}
    & \multicolumn{3}{c|}{Essay} & \multicolumn{3}{c|}{Paper Abs.} & \multicolumn{3}{c|}{Editorial} & \multicolumn{3}{c|}{Avg.} & \multicolumn{3}{c|}{Essay} & \multicolumn{3}{c|}{Paper Abs.} & \multicolumn{3}{c|}{Editorial} & \multicolumn{3}{c|}{Avg.} & \multicolumn{3}{c}{} \\
    \cmidrule(lr){2-4} \cmidrule(lr){5-7} \cmidrule(lr){8-10} \cmidrule(lr){11-13} \cmidrule(lr){14-16} \cmidrule(lr){17-19} \cmidrule(lr){20-22} \cmidrule(lr){23-25} \cmidrule(lr){26-28}
    & AUC & F1 & T@1 & AUC & F1 & T@1 & AUC & F1 & T@1 & AUC & F1 & T@1 & AUC & F1 & T@1 & AUC & F1 & T@1 & AUC & F1 & T@1 & AUC & F1 & T@1 & AUC & F1 & T@1 \\
    \midrule
    \rowcolor{gray!15} \multicolumn{28}{l}{\textit{Confidence-based}} \\
    ~~Log-Likelihood & 54.0 & 52.4 & 1.1 & 56.0 & 58.2 & 5.7 & 52.1 & 53.5 & 2.0 & 54.0 & 54.7 & 2.9 & 47.9 & 51.2 & 1.0 & 55.1 & 60.4 & 6.4 & 52.1 & 59.1 & 4.6 & 51.7 & 56.9 & 4.0 & 52.9 & 55.8 & 3.5 \\
    ~~Rank & 54.9 & 52.5 & 0.7 & 56.6 & 56.7 & 6.5 & 52.1 & 53.6 & 2.4 & 54.5 & 54.3 & 3.2 & 48.5 & 51.1 & 1.7 & 54.8 & 58.5 & 11.1 & 49.2 & 56.6 & 2.4 & 50.8 & 55.4 & 5.1 & 52.7 & 54.8 & 4.1 \\
    ~~Log-Rank & 54.1 & 52.8 & 1.0 & 55.6 & 58.3 & 5.7 & 52.2 & 53.6 & 1.8 & 54.0 & 54.9 & 2.8 & 48.0 & 50.6 & 1.5 & 55.9 & 64.0 & 6.4 & 50.5 & 58.1 & 2.9 & 51.5 & 57.6 & 3.6 & 52.7 & 56.2 & 3.2 \\
    ~~Entropy & 52.5 & 51.3 & 1.4 & 49.5 & 53.3 & 1.7 & 51.8 & 52.6 & 1.9 & 51.3 & 52.4 & 1.7 & 49.5 & 52.3 & 2.6 & 49.6 & 54.7 & 3.8 & 45.0 & 53.9 & 0.9 & 48.0 & 53.6 & 2.4 & 49.6 & 53.0 & 2.0 \\
    ~~LRR & 54.0 & 53.5 & 4.0 & 52.2 & 56.7 & 11.9 & 52.1 & 52.5 & 2.5 & 52.8 & 54.2 & 6.1 & 47.0 & 51.1 & 4.8 & 54.7 & 58.6 & 11.3 & 43.7 & 51.5 & 4.6 & 48.5 & 53.7 & 6.9 & 50.6 & 54.0 & 6.5 \\
    \rowcolor{gray!15} \multicolumn{28}{l}{\textit{Perturbation}} \\
    ~~NPR & 71.9 & 65.3 & 3.7 & 57.9 & 60.7 & 10.7 & 48.7 & 51.4 & 2.3 & 59.5 & 59.1 & 5.6 & 74.5 & 68.9 & 10.4 & 43.0 & 53.4 & 2.8 & 81.9 & 76.5 & 28.4 & 66.5 & 66.3 & 13.9 & 63.0 & 62.7 & 9.7 \\
    ~~DetectGPT & 62.2 & 58.6 & 4.0 & 52.7 & 54.7 & 5.8 & 52.7 & 53.5 & 0.9 & 55.9 & 55.6 & 3.6 & 59.1 & 57.6 & 8.1 & 49.7 & 58.0 & 2.6 & 82.6 & 79.6 & 32.9 & 63.8 & 65.1 & 14.5 & 59.8 & 60.3 & 9.0 \\
    ~~Fast-DetectGPT & 52.5 & 52.3 & 3.7 & 45.4 & 50.0 & 2.7 & 56.1 & 55.4 & 1.1 & 51.3 & 52.6 & 2.5 & 64.5 & 60.8 & 11.3 & 64.8 & 67.1 & 3.4 & 91.2 & 87.7 & 37.7 & 73.5 & 71.9 & 17.5 & 62.4 & 62.2 & 10.0 \\
    \rowcolor{gray!15} \multicolumn{28}{l}{\textit{LLM-based}} \\
    ~~Binoculars & 71.9 & 65.0 & 7.1 & 53.4 & 59.0 & 9.0 & 54.2 & 53.3 & 0.5 & 59.8 & 59.1 & 5.5 & 88.6 & 83.3 & \textbf{38.2} & 48.9 & 57.2 & 10.0 & 80.6 & 80.7 & 36.0 & 72.7 & 73.7 & 28.1 & 66.3 & 66.4 & 16.8 \\
    ~~LLM-Det (0-shot) & 60.0 & 58.8 & 1.8 & 50.8 & 51.4 & 0.4 & 56.5 & 54.3 & \textbf{4.0} & 55.8 & 54.8 & 2.1 & 67.9 & 63.1 & 4.0 & 57.0 & 58.5 & 2.0 & 67.8 & 66.3 & 15.8 & 64.2 & 62.6 & 7.3 & 60.0 & 58.7 & 4.7 \\
    ~~LLM-Det (10-shot) & 54.1 & 52.4 & 2.2 & 47.7 & 51.4 & 2.4 & 49.3 & 50.4 & 0.9 & 50.4 & 51.4 & 1.8 & 56.6 & 56.1 & 2.6 & 52.1 & 58.2 & 5.4 & 46.0 & 50.7 & 5.2 & 51.6 & 55.0 & 4.4 & 51.0 & 53.2 & 3.1 \\
    \rowcolor{gray!15} \multicolumn{28}{l}{\textit{Neural-based}} \\
    ~~XDAC-D & 65.5 & 61.9 & 0.0 & 63.3 & 62.9 & 0.0 & 59.6 & 58.5 & 0.0 & 62.8 & 61.1 & 0.0 & 68.5 & 69.4 & 0.0 & 58.8 & 58.0 & 0.0 & 67.7 & 69.1 & 0.0 & 65.0 & 65.5 & 0.0 & 63.9 & 63.3 & 0.0 \\
    ~~RepreGuard & 81.6 & 72.9 & \textbf{13.9} & 66.7 & \textbf{65.7} & 14.8 & 68.0 & 64.7 & 3.2 & 72.1 & 67.8 & \textbf{10.6} & 89.4 & 84.1 & 35.9 & 69.5 & 70.7 & 28.0 & 95.3 & 90.0 & 65.6 & 84.7 & 81.6 & 43.2 & 78.4 & 74.7 & 26.9 \\
    \rowcolor{gray!15} \multicolumn{28}{l}{\textit{Linguistic-Feature}} \\
    ~~KatFishNet & 60.9 & 57.0 & 1.5 & 50.0 & 42.6 & 0.0 & 52.9 & 52.0 & 2.0 & 54.6 & 50.5 & 1.2 & 73.7 & 66.4 & 11.6 & 53.1 & 61.0 & 0.0 & 91.9 & 86.2 & 40.9 & 72.9 & 71.2 & 17.5 & 63.8 & 60.9 & 9.3 \\
    \cellcolor[HTML]{EAECFF}\textbf{\textsc{KoLingProbe}} & \cellcolor[HTML]{EAECFF}\textbf{85.1} & \cellcolor[HTML]{EAECFF}\textbf{77.0} & \cellcolor[HTML]{EAECFF}12.8 & \cellcolor[HTML]{EAECFF}\textbf{66.9} & \cellcolor[HTML]{EAECFF}64.7 & \cellcolor[HTML]{EAECFF}\textbf{17.4} & \cellcolor[HTML]{EAECFF}\textbf{70.1} & \cellcolor[HTML]{EAECFF}\textbf{73.1} & \cellcolor[HTML]{EAECFF}0.0 & \cellcolor[HTML]{EAECFF}\textbf{74.0} & \cellcolor[HTML]{EAECFF}\textbf{71.6} & \cellcolor[HTML]{EAECFF}10.1 & \cellcolor[HTML]{EAECFF}\textbf{91.2} & \cellcolor[HTML]{EAECFF}\textbf{86.4} & \cellcolor[HTML]{EAECFF}22.8 & \cellcolor[HTML]{EAECFF}\textbf{74.0} & \cellcolor[HTML]{EAECFF}\textbf{73.9} & \cellcolor[HTML]{EAECFF}\textbf{47.6} & \cellcolor[HTML]{EAECFF}\textbf{99.6} & \cellcolor[HTML]{EAECFF}\textbf{99.1} & \cellcolor[HTML]{EAECFF}\textbf{98.5} & \cellcolor[HTML]{EAECFF}\textbf{88.3} & \cellcolor[HTML]{EAECFF}\textbf{86.5} & \cellcolor[HTML]{EAECFF}\textbf{56.3} & \cellcolor[HTML]{EAECFF}\textbf{81.1} & \cellcolor[HTML]{EAECFF}\textbf{79.0} & \cellcolor[HTML]{EAECFF}\textbf{33.2} \\
    \bottomrule
\end{tabular}}
\caption{\textbf{Binary} detection: AUROC (AUC), macro-F1 (F1), and TPR@FPR{=}1\% (T@1) in \%, averaged over 5 generators $\times$ 5 seeds. Avg.\ columns give per-task and overall averages; per-column best is \textbf{bolded}.}
\label{tab:exp_binary}
\end{table*}

\subsection{Training and Integration}
\label{individual_training}

Training one classifier on the concatenated 47 features would mix heterogeneous signals and invite cross-module interference~\cite{guyon2003introduction}.\footnote{Indeed, a single classifier over the concatenated features proves far more fragile under typo noise (Appendix~\ref{app:concat_ablation}).} We therefore adopt a two-stage architecture of lightweight logistic regressions: each linguistic level is modeled independently, and only the resulting decisions are combined.

\paragraph{Stage 1: Independent Sub-Detectors.}
Each linguistic level gets its own logistic regression classifier, a \emph{sub-detector}. Sub-detector $m$ receives only its own feature vector $\mathbf{x}_m \in \mathbb{R}^{d_m}$ and learns to predict the probability that a text is LLM-generated:
\begin{equation}
    \hat{p}_m(x) = \sigma\!\left(\mathbf{w}_m^\top \mathbf{x}_m + b_m\right)
\end{equation}
where $\sigma$ is the logistic function. Training each sub-detector in isolation prevents cross-level contamination. Prior to training, stability selection~\cite{meinshausen2009stabilityselection} identifies a robust feature subset per sub-detector; all sub-detectors run on CPU.

\paragraph{Stage 2: Decision-Level Fusion.}
\label{sub-detector_integration}
A logistic regression \emph{meta-learner} fuses the five sub-detector probabilities into the final decision:
\begin{equation}
    \hat{y}(x) = \sigma\!\left(\sum_{m=1}^{M} \alpha_m \cdot \hat{p}_m(x) + \beta\right)
\end{equation}
where $M{=}5$, $\alpha_m$ are the learned sub-detector weights, and $\beta$ is the bias. The meta-learner learns how much to trust each linguistic perspective; texts whose output exceeds 0.5 are predicted as LLM-generated. To prevent data leakage, sub-detector predictions used for training are generated via out-of-fold cross-validation. Further training details are provided in Appendix~\ref{app:training_details}.

\section{Experiments}
\label{sec:experiments}

\subsection{Experimental Settings}
\label{exp_settings}

\paragraph{Datasets.}
We evaluate all methods on \textsc{KoLing} on two binary tasks per domain, HLT-vs-HWT (LLM-modified) and LGT-vs-HWT (fully generated), and a ternary HWT/HLT/LGT task inspired by \citet{detectrlx2026} (\S\ref{sec:ternary}). Following \S\ref{sec:koling}, all results are averaged over five generators (one in-distribution, four out-of-distribution) and five seeds (42--46)\footnote{For readability; standard deviations are in Table~\ref{tab:app_std}.}.

\paragraph{Metrics.}
We report AUROC (AUC; threshold-free ranking quality), macro-F1 (F1; positive-class F1 would reward majority-class prediction on our imbalanced classes), and TPR@FPR{=}1\% (T@1; a stringent operating point where falsely accusing a human writer is costly).

\paragraph{Baselines.}
We compare \textsc{KoLingProbe} against 14 baselines spanning five categories:
\textbf{Confidence} (Log-Likelihood~\cite{solaiman2019release}, Rank, Entropy~\cite{gehrmann-etal-2019-gltr}, Log-Rank, LRR~\cite{su-etal-2023-detectllm}; scored with KoGPT2~\cite{kogpt2}, the standard open Korean GPT-2); \textbf{Perturbation} (NPR~\cite{su-etal-2023-detectllm}, DetectGPT~\cite{detectgpt2023}, Fast-DetectGPT~\cite{bao-et-al-fastdetectgpt-2024}; pko-T5~\cite{paust_pkot5_v1});
\textbf{LLM-based} (Binoculars~\cite{hans2024binoculars}; EXAONE-3.5-7.8B~\cite{an2024exaone35serieslarge} prompted as a detector, 0-/10-shot);
\textbf{Neural-based} (XDAC-D~\cite{go-etal-2025-xdac}; RepreGuard~\cite{chen-etal-2025-repreguard}, a frozen-encoder probe); and
\textbf{Linguistic-Feature} (KatFishNet~\cite{park-etal-2025-katfishnet}, syntax-only; our \textsc{KoLingProbe}, multi-level).
None of these update a large pre-trained backbone on the target task; fully \emph{fine-tuned} detectors such as XLM-RoBERTa operate under a different supervision regime and are analyzed separately in \S\ref{sec:exp_finetuned}. Full configurations are listed in Appendix~\ref{app:training_details}.\footnote{We also evaluated the grammar-based GECScore~\cite{wu-etal-2025-wrote}; it is near chance on \textsc{KoLing}. Korean lacks a strong open-source GEC model, so we report it only in the appendix tables (Appendix~\ref{app:koling_per_llm}).}

\subsection{Main Results: Binary Detection}
\label{main_results}

Table~\ref{tab:exp_binary} summarizes binary detection performance across both tasks and all domains. After controlling for surface-level shortcuts, the adversarial HLT-vs-HWT task clearly differentiates methods: confidence-based baselines drop to near-chance performance, whereas \textsc{KoLingProbe} achieves the highest AUROC in every domain--task setting and the best overall averages, outperforming its strongest frozen-feature baseline, RepreGuard, on both task averages. Furthermore, two consistent patterns emerge. First, LGT detection is uniformly easier than HLT detection across all domains, confirming that detecting LLM-modified text is the more challenging and practically relevant setting. Second, the Paper Abstract domain remains the most difficult for all methods, likely because its highly standardized writing style provides weaker discriminative signals; \S\ref{sec:transfer} attributes this to uniformly small feature effect sizes.\footnote{Per-generator breakdowns, essay-type and modification-level analyses, and inference efficiency comparisons are provided in Appendices~\ref{app:koling_per_llm}, \ref{app:variance}, \ref{app:essay_breakdown}, and~\ref{app:inference_time}.}

\subsection{Ternary Classification}
\label{sec:ternary}

\begin{table}[h!]
\centering
\small
\setlength{\tabcolsep}{4pt}
\resizebox{\columnwidth}{!}{
\begin{tabular}{l|ccc|c}
    \toprule
    \textbf{Method} & \textbf{Essay} & \textbf{Paper Abs.} & \textbf{Editorial} & \textbf{Avg.} \\
    \midrule
    \rowcolor{gray!15} \multicolumn{5}{l}{\textit{Confidence-based}} \\
    ~~Log-Likelihood & 32.6/51.7 & 38.5/53.1 & 35.1/52.7 & 35.4/52.5 \\
    ~~Rank & 32.6/47.9 & 38.4/52.1 & 35.0/48.1 & 35.3/49.4 \\
    ~~Log-Rank & 32.6/45.7 & 36.3/51.3 & 34.1/49.7 & 34.3/48.9 \\
    ~~Entropy & 33.8/43.9 & 33.5/42.0 & 34.0/38.2 & 33.8/41.4 \\
    ~~LRR & 34.2/44.1 & 33.4/43.1 & 34.4/39.8 & 34.0/42.3 \\
    \rowcolor{gray!15} \multicolumn{5}{l}{\textit{Perturbation}} \\
    ~~NPR & 45.4/54.7 & 35.2/41.2 & 47.1/52.9 & 42.6/49.6 \\
    ~~DetectGPT & 39.2/47.1 & 30.8/43.5 & 42.6/47.3 & 37.5/46.0 \\
    ~~Fast-DetectGPT & 39.0/48.3 & 32.2/45.2 & 52.8/55.6 & 41.3/49.7 \\
    \rowcolor{gray!15} \multicolumn{5}{l}{\textit{LLM-based}} \\
    ~~Binoculars & 55.2/60.6 & 37.2/41.7 & 48.1/54.6 & 46.8/52.3 \\
    ~~LLM-Det (0-shot) & 40.2/48.5 & 35.7/46.3 & 27.1/37.8 & 34.3/44.2 \\
    ~~LLM-Det (10-shot) & 35.3/47.2 & 33.4/43.0 & 37.9/48.3 & 35.5/46.2 \\
    \rowcolor{gray!15} \multicolumn{5}{l}{\textit{Neural-based}} \\
    ~~XDAC-D & 28.4/46.5 & 25.4/24.1 & 35.5/39.3 & 29.8/36.6 \\
    ~~RepreGuard & 49.2/61.9 & \textbf{42.3}/56.5 & 40.6/51.5 & 44.0/56.6 \\
    \rowcolor{gray!15} \multicolumn{5}{l}{\textit{Linguistic-Feature}} \\
    ~~KatFishNet & 33.8/56.6 & 24.6/\textbf{58.5} & 42.1/63.6 & 33.5/59.6 \\
    \cellcolor[HTML]{EAECFF}\textbf{\textsc{KoLingProbe}} & \cellcolor[HTML]{EAECFF}\textbf{57.8}/\textbf{66.1} & \cellcolor[HTML]{EAECFF}40.0/\textbf{58.5} & \cellcolor[HTML]{EAECFF}\textbf{72.9}/\textbf{77.5} & \cellcolor[HTML]{EAECFF}\textbf{56.9}/\textbf{67.4} \\
    \bottomrule
\end{tabular}}
\caption{\textbf{Ternary} HWT/HLT/LGT classification: macro-F1~/~Accuracy (\%). The last column averages the three domains; per-column best is \textbf{bolded}.}
\label{tab:exp_ternary}
\end{table}

\begin{figure}[t!]
  \centering
  \includegraphics[width=.95\linewidth]{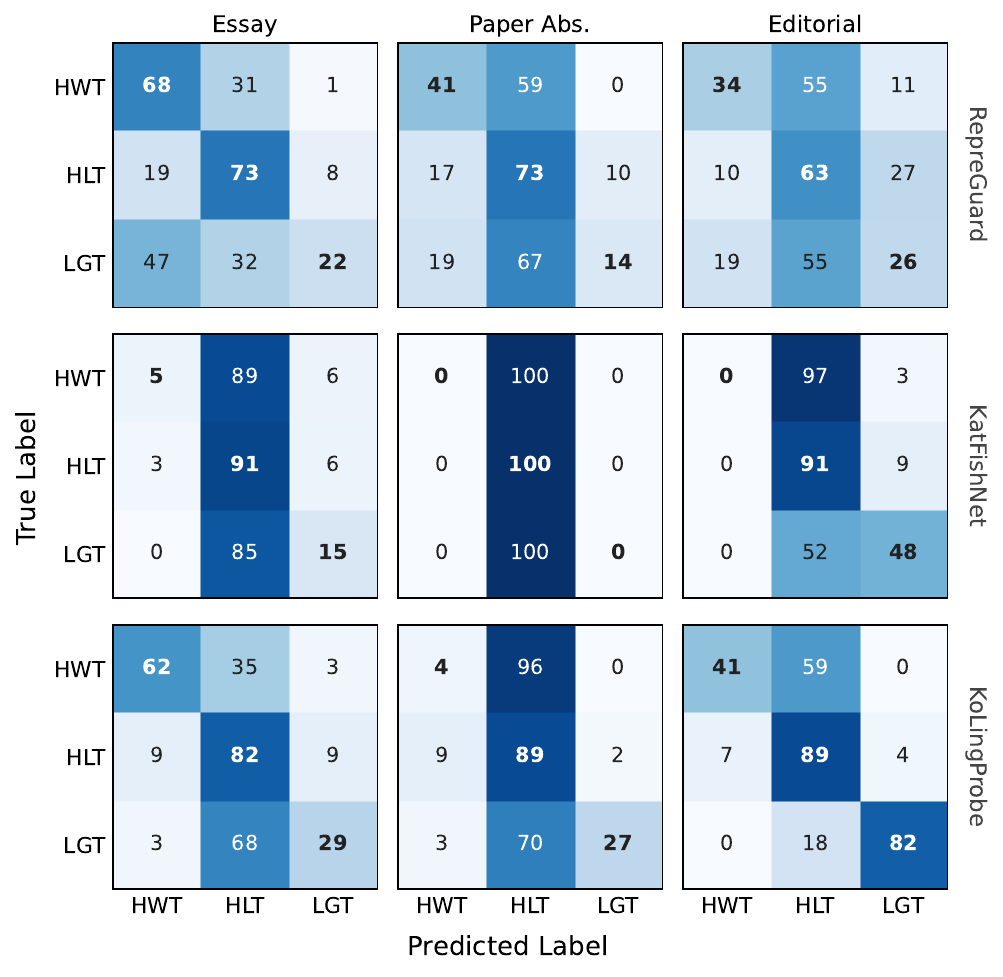}
  \caption{Ternary confusion matrices (row-normalized \%; rows = true) for the two native 3-class classifiers (\textsc{KatFishNet}, \textsc{KoLingProbe}) and the strongest single-score baseline (RepreGuard, threshold-mapped).}
  \label{fig:exp_confusion}
\end{figure}

In real deployments, a detector must often judge not only whether an LLM was involved but to what degree, for instance, when policies permit light editing but prohibit full generation~\cite{acl2023policy}. To test whether detectors can make this finer-grained distinction, we conduct an additional ternary HWT/HLT/LGT classification task. Table~\ref{tab:exp_ternary} reports the ternary classification results. Native 3-class classifiers predict three ways directly, while single-score baselines are mapped to three classes via train-calibrated 2-thresholds. Under this harder setting, most baselines collapse toward chance; \textsc{KoLingProbe} leads on essay, editorial, and the three-domain average, while the low-signal Paper Abstract domain remains challenging for every method.
Figure~\ref{fig:exp_confusion} shows confusion matrices for the two native
3-class classifiers and the strongest single-score baseline. The comma-only
KatFishNet collapses to the majority HLT class on both Paper Abstract and Editorial
(0\% HWT recall), and every method degenerates on the low-signal Paper Abstract
domain; on Editorial, however, \textsc{KoLingProbe} is the only method that keeps
all three classes separable.

\subsection{Robustness Comparison with Fine-tuned Detectors}
\label{sec:exp_finetuned}

Detectors that fine-tune a full pre-trained backbone on each task's training split operate under a heavier supervision regime than the baselines above, so we compare against the strongest such model, XLM-RoBERTa, separately. On clean text, they are remarkably strong: fine-tuned XLM-RoBERTa surpasses even \textsc{KoLingProbe} on the Essay domain.

\begin{figure}[h!]
  \centering
  \includegraphics[width=.95\columnwidth]{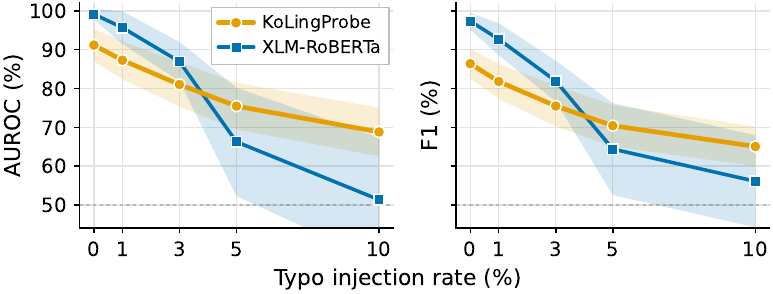}
  \caption{Essay LGT-vs-HWT detection as the typo injection rate increases (mean over 5 generators $\times$ 5 seeds; shaded: $\pm$1 std; dotted: chance).}
  \label{fig:app_typosweep}
\end{figure}

This strength, however, comes from fitting tightly to the training distribution, and it is brittle once that distribution is perturbed. Under character-level Korean typo attacks (jamo swaps, spacing noise, and initial-law violations injected into LLM-side test texts only; Appendix~\ref{app:typo_robustness}), Figure~\ref{fig:app_typosweep} shows XLM-RoBERTa collapsing to near chance once the typo rate passes 3\%, whereas \textsc{KoLingProbe}'s token-aggregate linguistic features degrade gracefully and overtake it from roughly 4\% onward. On the ternary task, XLM-RoBERTa likewise falls to majority-class prediction on Paper Abstract and Editorial (Macro-F1 25.4 and 32.5). Since real-world text is riddled with typos, this fragility makes fine-tuned detectors difficult to deploy in practice: high clean accuracy conceals a fundamental robustness problem.

\section{Interpretability}
\label{sec:interpretability}

\textsc{KoLingProbe}'s robustness is not incidental; it follows from the modular, linguistically grounded design. We examine which sub-detectors contribute and whether they are complementary.

\subsection{Sub-Detector Contribution Analysis}

\begin{figure}[h!]
  \centering
  \includegraphics[width=.95\linewidth]{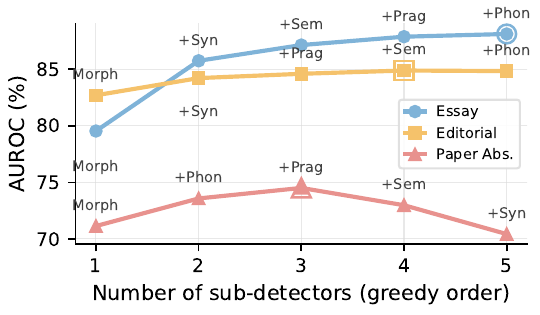}
  \caption{Greedy forward selection of \textsc{KoLingProbe} sub-detectors per domain.
  }
  \label{fig:greedy_domains}
\end{figure}

Figure~\ref{fig:greedy_domains} traces greedy forward selection per domain: starting from the best single module, we repeatedly add the module that most improves overall AUROC. Morphology is the strongest single module everywhere, consistent with Korean's agglutinative structure providing the richest signals. Beyond that shared start, the domains diverge: on the high-signal Essay domain every addition helps; Editorial saturates at four modules; and on the low-signal Paper Abstract domain performance peaks at three modules before further additions hurt: where discriminative signal is weak, extra feature dimensions overfit. The same pattern holds under F1. This diagnosis is possible only because each level is an independent sub-detector: the benchmark reveals not just how much signal a domain carries, but which linguistic levels carry it (all 31 combinations in Appendix~\ref{app:all_results_koling}).

\subsection{Sub-Detector Complementarity}

\begin{figure}[h!]
  \centering
  \includegraphics[width=.9\linewidth]{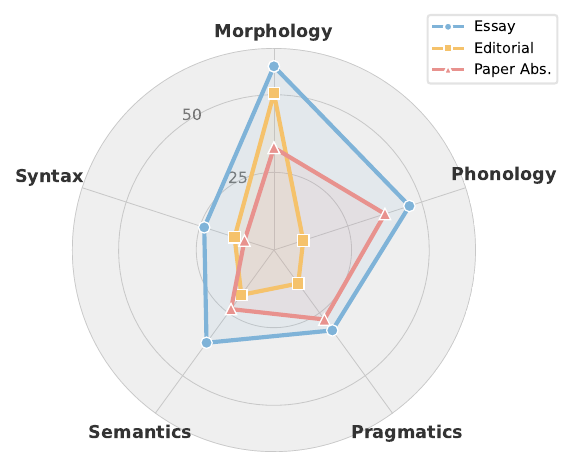}
  \caption{Rescue analysis per domain: the fraction of KatFishNet's misclassified samples that each \textsc{KoLingProbe} sub-detector alone classifies correctly.}
  \label{fig:sub-detector_rescue}
  \vspace{-5px}
\end{figure}

To test whether the five sub-detectors capture different error patterns, we collect the test samples misclassified by the comma-only KatFishNet and measure the fraction that each \textsc{KoLingProbe} sub-detector alone classifies correctly. Figure~\ref{fig:sub-detector_rescue} shows a similar shape in every domain: the syntax sub-detector rescues the least (it overlaps most with KatFishNet's comma-centered cues), the strongest rescuer is always a non-syntactic level, and at least one sub-detector corrects 48--81\% of KatFishNet's errors. Robust detection, in short, arises from complementary linguistic perspectives rather than reliance on any single dominant cue.

\section{Transferability}
\label{sec:transfer}

A deployed detector inevitably faces unseen domains, where corpus-specific cues no longer hold. If \textsc{KoLingProbe} captures general properties of Korean rather than corpus artifacts, its signal should transfer beyond the training distribution. We evaluate this at three levels: cross-domain feature consistency, cross-domain detector transfer, and external benchmark generalization.

\begin{figure}[t!]
  \centering
  \includegraphics[width=1.\linewidth]{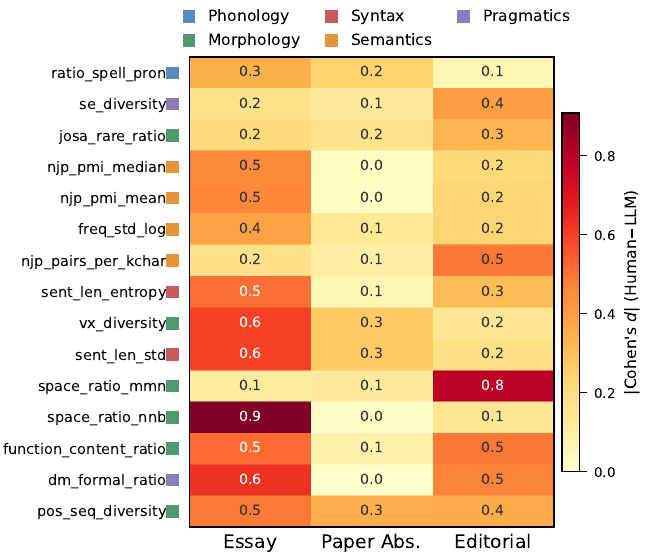}
  \caption{Cross-domain portability of \textsc{KoLingProbe}'s features: effect sizes ($|$Cohen's~$d|$, human vs.\ LLM) of the top-15 features across the three domains.}
  \label{fig:exp_portability}
  \vspace{-5px}
\end{figure}

\paragraph{Feature level.}
Figure~\ref{fig:exp_portability} plots the effect sizes of \textsc{KoLingProbe}'s most discriminative features (top 15 of 47 by mean $|$Cohen's~$d|$) across the three domains. Essay and editorial share several discriminative features (e.g., \texttt{dm\_formal\_ratio}), indicating that the linguistic signature of LLM-generated Korean is partly domain-general rather than an artifact of student writing. In contrast, paper abstracts exhibit uniformly small effect sizes, suggesting that their rigid conventions compress the stylistic space of both human and LLM writing, consistent with its lower detection performance. Feature-level interpretability therefore not only explains model predictions but also reveals when a domain contains sufficient signal for reliable detection.

\paragraph{Detector level.}
Appendix Table~\ref{tab:cross_domain_all} reports cross-domain transfer, training each detector on one domain and evaluating it on the other two. Among methods with the same supervision regime, \textsc{KoLingProbe} achieves the highest average transfer AUROC, outperforming RepreGuard on 9 of 12 domain pairs. Linguistic features are especially effective when transferring from Editorial and Paper Abstract: an editorial-trained \textsc{KoLingProbe} reaches 85.0 AUROC on Essay LGT and an academic-trained one 69.3 on Essay HLT, the best results among all methods, including fine-tuned XLM-RoBERTa, which transfers well only from Essay. These results suggest that linguistic features generalize in accordance with their effect sizes, even from relatively small training corpora.

\paragraph{Benchmark level.}
\label{sec:generalization}
Table~\ref{tab:katfish_results} compares all methods on KatFish~\cite{park-etal-2025-katfishnet}, a shortcut-rich benchmark with from-scratch generation. \textsc{KoLingProbe} achieves the best average AUROC (84.1), outperforming KatFishNet on its own benchmark and showing that multi-level linguistic modeling generalizes to both shortcut-rich and shortcut-controlled settings (Appendix~\ref{app:feature_effects}). Under post-shortcut control, its average AUROC drops by only 5.9, versus 13.2 for KatFishNet itself (Appendix~\ref{app:katfish_clean}).

\section{Related Work}
\paragraph{LLM-Generated Text Detection.} Benchmarks have scaled from early human--machine pair collections~\cite{uchendu-etal-2021-turingbench-benchmark, hc3-guo2023close} to multi-task, multilingual, in-the-wild, and adversarially robust testbeds~\cite{he2024mgtbench, wang-etal-2024-m4gt, li-etal-2024-mage, dugan-etal-2024-raid, wu2024detectrl}. For Korean, KatFish~\cite{park-etal-2025-katfishnet} is the only dedicated benchmark for long-form text, but as we show, it retains surface-level biases that enable shortcut-based detection. \textsc{KoLing} addresses this gap through systematic generation with explicit bias control.

\begin{table}[t!]
\centering
\renewcommand{\arraystretch}{0.95}
\resizebox{\columnwidth}{!}{
\begin{tabular}{ll|ccc|c}
    \toprule
    \multicolumn{2}{c|}{\textbf{Models}} & \textbf{Essay} & \textbf{Poetry} & \textbf{Abstract} & \textbf{Avg.} \\
    \midrule
    \multirow{5}{*}{Confidence-based} & Log-Likelihood & 51.58 & 43.32 & 64.11 & 53.00 \\
     & Rank & 53.00 & 40.78 & 67.61 & 53.79 \\
     & Log-Rank & 55.94 & 44.64 & 66.31 & 55.63 \\
     & Entropy & 51.09 & 47.76 & 61.57 & 53.47 \\
     & LRR & 64.45 & 46.65 & 66.18 & 59.10 \\
    \midrule
    \multirow{3}{*}{Perturbation} & NPR & 82.30 & 65.49 & 60.21 & 69.33 \\
     & DetectGPT & 57.40 & 45.86 & 44.01 & 49.09 \\
     & Fast-DetectGPT & 53.58 & 58.42 & 55.04 & 55.68 \\
    \midrule
    \multirow{3}{*}{LLM-based} & Binoculars & 96.43 & 63.06 & 68.61 & 76.03 \\
     & LLM-Det (0-shot) & 66.17 & 63.36 & 41.35 & 56.96 \\
     & LLM-Det (10-shot) & 82.61 & 59.31 & 57.12 & 66.34 \\
    \midrule
    \multirow{2}{*}{Neural-based} & XDAC-D & 85.09 & 49.65 & 55.94 & 63.56 \\
     & RepreGuard & 94.06 & 78.63 & 54.24 & 75.64 \\
    \midrule
    \multirow{2}{*}{Linguistic-Feature} & KatFishNet & 95.61 & 72.05 & 73.99 & 80.55 \\
     & \cellcolor[HTML]{EAECFF}\textbf{\textsc{KoLingProbe}} & \cellcolor[HTML]{EAECFF}\textbf{98.09} & \cellcolor[HTML]{EAECFF}\textbf{79.33} & \cellcolor[HTML]{EAECFF}\textbf{74.79} & \cellcolor[HTML]{EAECFF}\textbf{84.07} \\
    \bottomrule
\end{tabular}}
\caption{Detection performance (AUROC \%) on the KatFish benchmark. Best results per column are \textbf{bolded}.}
\label{tab:katfish_results}
\end{table}

\paragraph{Detectors.} Detection methods span confidence-based~\cite{gehrmann-etal-2019-gltr, su-etal-2023-detectllm}, perturbation-based~\cite{detectgpt2023, bao-et-al-fastdetectgpt-2024}, neural~\cite{solaiman2019release}, LLM-based~\cite{bhattacharjee2024fighting, hans2024binoculars}, representation-probing~\cite{chen-etal-2025-repreguard}, and grammar-based approaches~\cite{wu-etal-2025-wrote}, developed primarily for English around surface-level distributional differences~\cite{wu-etal-2025-survey}. For Korean, KatFishNet~\cite{park-etal-2025-katfishnet} (spacing, POS $n$-grams, commas) and XDAC-D~\cite{go-etal-2025-xdac} remain limited to such surface cues; \textsc{KoLingProbe} instead grounds detection in five linguistic levels.

\section{Conclusion}

We presented \textsc{KoLing}, a bias-controlled Korean benchmark spanning three domains and the HWT/HLT/LGT authorship spectrum, and \textsc{KoLingProbe}, an interpretable detector built on the five levels of theoretical linguistics. \textsc{KoLingProbe} achieves the best overall performance on both binary and ternary detection tasks. Our analyses show that this advantage stems from its design: the five linguistic sub-detectors provide complementary evidence, overcoming the limitations of syntax-only detection while yielding features that transfer across domains, even from relatively small sources, and to an external benchmark. Moreover, unlike fine-tuned neural detectors, which degrade substantially under typographical perturbations, \textsc{KoLingProbe} remains robust while requiring only CPU inference. We argue that Korean LLM-text detection should be evaluated not only by clean accuracy, but also by whether a detector remains \emph{interpretable}, \emph{robust}, and \emph{transferable} when surface cues are controlled, corrupted, or shifted.

\clearpage

\section*{Limitations}

Despite these contributions, several limitations remain. First, although \textsc{KoLing} covers three domains spanning a broad register spectrum, other domain categories inevitably exist, and extending the benchmark to them is left for future work. Second, the 47 features are heuristic operationalizations, grounded in Korean linguistics, but not exhaustive of the language's grammatical phenomena. Third, several phonological and semantic features rely on dictionaries precompiled from large corpora and may inherit their coverage biases. Fourth, the 47 features are Korean-specific, so the detector does not apply to other languages as-is; however, the five-level framework itself is language-agnostic, and porting \textsc{KoLingProbe} requires re-instantiating the per-level feature extractors rather than redesigning the architecture. Finally, we evaluate only modest character-level attacks (\S\ref{sec:exp_finetuned}); more sophisticated semantic-preserving attacks are left for future work.

\section*{Ethical Considerations}

\textsc{KoLing} is built from publicly released resources, the AI-Hub ``Essay Writing Evaluation'' and ``Document Summarization'' datasets and the KatFish source articles, used within their stated research terms; we retain only the released text and standard metadata, with no personally identifying information. Detection systems carry inherent risks of demographic and linguistic bias: writers whose patterns differ from the training distribution, such as non-native Korean or dialect speakers, may be disproportionately misclassified, and some features may partially correlate with educational level. We therefore caution against high-stakes use, such as academic-misconduct adjudication, without further fairness validation across diverse user groups. All LLM-generated texts are produced with publicly accessible APIs or open-weight models, and we release the benchmark, feature-extraction pipelines, and trained classifiers subject to the corresponding providers' licensing policies.


\bibliography{custom}

@book{Lyons_1968, place={Cambridge}, title={Introduction to Theoretical Linguistics}, publisher={Cambridge University Press}, author={Lyons, John}, year={1968}}

@inproceedings{gehrmann-etal-2019-gltr,
    title = "{GLTR}: Statistical Detection and Visualization of Generated Text",
    author = "Gehrmann, Sebastian  and
      Strobelt, Hendrik  and
      Rush, Alexander",
    editor = "Costa-juss{\`a}, Marta R.  and
      Alfonseca, Enrique",
    booktitle = "Proceedings of the 57th Annual Meeting of the Association for Computational Linguistics: System Demonstrations",
    month = jul,
    year = "2019",
    address = "Florence, Italy",
    publisher = "Association for Computational Linguistics",
    url = "https://aclanthology.org/P19-3019/",
    doi = "10.18653/v1/P19-3019",
    pages = "111--116"
}

@inproceedings{park-etal-2025-katfishnet,
    title = "{K}at{F}ish{N}et: Detecting {LLM}-Generated {K}orean Text through Linguistic Feature Analysis",
    author = "Park, Shinwoo  and
      Kim, Shubin  and
      Kim, Do-Kyung  and
      Han, Yo-Sub",
    editor = "Che, Wanxiang  and
      Nabende, Joyce  and
      Shutova, Ekaterina  and
      Pilehvar, Mohammad Taher",
    booktitle = "Proceedings of the 63rd Annual Meeting of the Association for Computational Linguistics (Volume 1: Long Papers)",
    month = jul,
    year = "2025",
    address = "Vienna, Austria",
    publisher = "Association for Computational Linguistics",
    url = "https://aclanthology.org/2025.acl-long.1030/",
    doi = "10.18653/v1/2025.acl-long.1030",
    pages = "21189--21222",
    ISBN = "979-8-89176-251-0",
}

@inproceedings{detectgpt2023,
author = {Mitchell, Eric and Lee, Yoonho and Khazatsky, Alexander and Manning, Christopher D. and Finn, Chelsea},
title = {DetectGPT: zero-shot machine-generated text detection using probability curvature},
year = {2023},
publisher = {JMLR.org},
booktitle = {Proceedings of the 40th International Conference on Machine Learning},
articleno = {1038},
numpages = {13},
location = {Honolulu, Hawaii, USA},
series = {ICML'23}
}

@inproceedings{go-etal-2025-xdac,
    title = "{XDAC}: {XAI}-Driven Detection and Attribution of {LLM}-Generated News Comments in {K}orean",
    author = "Go, Wooyoung  and
      Kim, Hyoungshick  and
      Oh, Alice  and
      Kim, Yongdae",
    editor = "Che, Wanxiang  and
      Nabende, Joyce  and
      Shutova, Ekaterina  and
      Pilehvar, Mohammad Taher",
    booktitle = "Proceedings of the 63rd Annual Meeting of the Association for Computational Linguistics (Volume 1: Long Papers)",
    month = jul,
    year = "2025",
    address = "Vienna, Austria",
    publisher = "Association for Computational Linguistics",
    url = "https://aclanthology.org/2025.acl-long.1108/",
    doi = "10.18653/v1/2025.acl-long.1108",
    pages = "22728--22750",
    ISBN = "979-8-89176-251-0",
}

@inproceedings{su-etal-2023-detectllm,
    title = "{D}etect{LLM}: Leveraging Log Rank Information for Zero-Shot Detection of Machine-Generated Text",
    author = "Su, Jinyan  and
      Zhuo, Terry  and
      Wang, Di  and
      Nakov, Preslav",
    editor = "Bouamor, Houda  and
      Pino, Juan  and
      Bali, Kalika",
    booktitle = "Findings of the Association for Computational Linguistics: EMNLP 2023",
    month = dec,
    year = "2023",
    address = "Singapore",
    publisher = "Association for Computational Linguistics",
    url = "https://aclanthology.org/2023.findings-emnlp.827/",
    doi = "10.18653/v1/2023.findings-emnlp.827",
    pages = "12395--12412",
}

@article{wu-etal-2025-survey,
    title = "A Survey on {LLM}-Generated Text Detection: Necessity, Methods, and Future Directions",
    author = "Wu, Junchao  and
      Yang, Shu  and
      Zhan, Runzhe  and
      Yuan, Yulin  and
      Chao, Lidia Sam  and
      Wong, Derek Fai",
    journal = "Computational Linguistics",
    volume = "51",
    number = "1",
    month = mar,
    year = "2025",
    address = "Cambridge, MA",
    publisher = "MIT Press",
    url = "https://aclanthology.org/2025.cl-1.8/",
    doi = "10.1162/coli_a_00549",
    pages = "275--338",
}

@inproceedings{wan-etal-2025-social,
    title = "How Do Social Bots Participate in Misinformation Spread? A Comprehensive Dataset and Analysis",
    author = "Wan, Herun  and
      Luo, Minnan  and
      Ma, Zihan  and
      Dai, Guang  and
      Zhao, Xiang",
    editor = "Christodoulopoulos, Christos  and
      Chakraborty, Tanmoy  and
      Rose, Carolyn  and
      Peng, Violet",
    booktitle = "Proceedings of the 2025 Conference on Empirical Methods in Natural Language Processing",
    month = nov,
    year = "2025",
    address = "Suzhou, China",
    publisher = "Association for Computational Linguistics",
    url = "https://aclanthology.org/2025.emnlp-main.1604/",
    doi = "10.18653/v1/2025.emnlp-main.1604",
    pages = "31493--31516",
    ISBN = "979-8-89176-332-6",
}

@inproceedings{feng-etal-2024-bot,
    title = "What Does the Bot Say? Opportunities and Risks of Large Language Models in Social Media Bot Detection",
    author = "Feng, Shangbin  and
      Wan, Herun  and
      Wang, Ningnan  and
      Tan, Zhaoxuan  and
      Luo, Minnan  and
      Tsvetkov, Yulia",
    editor = "Ku, Lun-Wei  and
      Martins, Andre  and
      Srikumar, Vivek",
    booktitle = "Proceedings of the 62nd Annual Meeting of the Association for Computational Linguistics (Volume 1: Long Papers)",
    month = aug,
    year = "2024",
    address = "Bangkok, Thailand",
    publisher = "Association for Computational Linguistics",
    url = "https://aclanthology.org/2024.acl-long.196/",
    doi = "10.18653/v1/2024.acl-long.196",
    pages = "3580--3601",
}

@article{manipulation2022,
author = {Zerback, Thomas and Töpfl, Florian},
title = {Forged Examples as Disinformation: The Biasing Effects of Political Astroturfing Comments on Public Opinion Perceptions and How to Prevent Them},
journal = {Political Psychology},
volume = {43},
number = {3},
pages = {399-418},
doi = {https://doi.org/10.1111/pops.12767},
url = {https://onlinelibrary.wiley.com/doi/abs/10.1111/pops.12767},
eprint = {https://onlinelibrary.wiley.com/doi/pdf/10.1111/pops.12767},
year = {2022}
}

@inproceedings{pan-etal-2023-risk,
    title = "On the Risk of Misinformation Pollution with Large Language Models",
    author = "Pan, Yikang  and
      Pan, Liangming  and
      Chen, Wenhu  and
      Nakov, Preslav  and
      Kan, Min-Yen  and
      Wang, William",
    editor = "Bouamor, Houda  and
      Pino, Juan  and
      Bali, Kalika",
    booktitle = "Findings of the Association for Computational Linguistics: EMNLP 2023",
    month = dec,
    year = "2023",
    address = "Singapore",
    publisher = "Association for Computational Linguistics",
    url = "https://aclanthology.org/2023.findings-emnlp.97/",
    doi = "10.18653/v1/2023.findings-emnlp.97",
    pages = "1389--1403",
}

@INPROCEEDINGS{orenstrakh-et-al-2024,
  author={Orenstrakh, Michael Sheinman and Karnalim, Oscar and Suárez, Carlos Aníbal and Liut, Michael},
  booktitle={2024 IEEE 48th Annual Computers, Software, and Applications Conference (COMPSAC)}, 
  title={Detecting LLM-Generated Text in Computing Education: Comparative Study for ChatGPT Cases}, 
  year={2024},
  volume={},
  number={},
  pages={121-126},
  doi={10.1109/COMPSAC61105.2024.00027}
}

@inproceedings{bao-et-al-fastdetectgpt-2024,
  author       = {Guangsheng Bao and
                  Yanbin Zhao and
                  Zhiyang Teng and
                  Linyi Yang and
                  Yue Zhang},
  title        = {Fast-DetectGPT: Efficient Zero-Shot Detection of Machine-Generated
                  Text via Conditional Probability Curvature},
  booktitle    = {The Twelfth International Conference on Learning Representations,
                  {ICLR} 2024, Vienna, Austria, May 7-11, 2024},
  publisher    = {OpenReview.net},
  year         = {2024},
  url          = {https://openreview.net/forum?id=Bpcgcr8E8Z},
  bibsource    = {dblp computer science bibliography, https://dblp.org}
}

@article{hc3-guo2023close,
  title={How close is chatgpt to human experts? comparison corpus, evaluation, and detection},
  author={Guo, Biyang and Zhang, Xin and Wang, Ziyuan and Jiang, Minqi and Nie, Jinran and Ding, Yuxuan and Yue, Jianwei and Wu, Yupeng},
  journal={arXiv preprint arXiv:2301.07597},
  year={2023}
}

@article{gpt-4o,
  title={Gpt-4o system card},
  author={Hurst, Aaron and Lerer, Adam and Goucher, Adam P and Perelman, Adam and Ramesh, Aditya and Clark, Aidan and Ostrow, AJ and Welihinda, Akila and Hayes, Alan and Radford, Alec and others},
  journal={arXiv preprint arXiv:2410.21276},
  year={2024}
}

@misc{claude-4.6,
  author       = {{Anthropic}},
  title        = {Introducing Claude Sonnet 4.6},
  year         = {2026},
  month        = feb,
  day          = {17},
  howpublished = {\url{https://www.anthropic.com/news/claude-sonnet-4-6}},
  note         = {Accessed: 2026-05-25}
}

@misc{solar-pro3,
  author       = {{Upstage}},
  title        = {Solar Pro 3: Better Reasoning at Production Scale},
  year         = {2026},
  month        = jan,
  day          = {26},
  howpublished = {\url{https://www.upstage.ai/blog/en/solar-pro-3-0127}},
  note         = {Accessed: 2026-05-25}
}

@misc{mistral-3.1,
  author       = {{Mistral AI}},
  title        = {Mistral Small 3.1},
  year         = {2025},
  howpublished = {\url{https://mistral.ai/news/mistral-small-3-1/}},
  note         = {Accessed: 2026-05-25}
}

@misc{qwen2025qwen25technicalreport,
      title={Qwen2.5 Technical Report}, 
      author={{Qwen Team}},
      year={2025},
      eprint={2412.15115},
      archivePrefix={arXiv},
      primaryClass={cs.CL},
      url={https://arxiv.org/abs/2412.15115}, 
}

@inproceedings{wang-etal-2024-m4gt,
    title = "{M}4{GT}-Bench: Evaluation Benchmark for Black-Box Machine-Generated Text Detection",
    author = "Wang, Yuxia  and
      Mansurov, Jonibek  and
      Ivanov, Petar  and
      Su, Jinyan  and
      Shelmanov, Artem  and
      Tsvigun, Akim  and
      Mohammed Afzal, Osama  and
      Mahmoud, Tarek  and
      Puccetti, Giovanni  and
      Arnold, Thomas  and
      Aji, Alham  and
      Habash, Nizar  and
      Gurevych, Iryna  and
      Nakov, Preslav",
    editor = "Ku, Lun-Wei  and
      Martins, Andre  and
      Srikumar, Vivek",
    booktitle = "Proceedings of the 62nd Annual Meeting of the Association for Computational Linguistics (Volume 1: Long Papers)",
    month = aug,
    year = "2024",
    address = "Bangkok, Thailand",
    publisher = "Association for Computational Linguistics",
    url = "https://aclanthology.org/2024.acl-long.218/",
    doi = "10.18653/v1/2024.acl-long.218",
    pages = "3964--3992",
}

@inproceedings{he2024mgtbench,
  title={Mgtbench: Benchmarking machine-generated text detection},
  author={He, Xinlei and Shen, Xinyue and Chen, Zeyuan and Backes, Michael and Zhang, Yang},
  booktitle={Proceedings of the 2024 on ACM SIGSAC Conference on Computer and Communications Security},
  pages={2251--2265},
  year={2024}
}

@inproceedings{kim-etal-2025-polishing,
    title = "Polishing Every Facet of the {GEM}: Testing Linguistic Competence of {LLM}s and Humans in {K}orean",
    author = "Kim, SungHo  and
      Kim, Nayeon  and
      Jeon, Taehee  and
      Lee, SangKeun",
    editor = "Che, Wanxiang  and
      Nabende, Joyce  and
      Shutova, Ekaterina  and
      Pilehvar, Mohammad Taher",
    booktitle = "Proceedings of the 63rd Annual Meeting of the Association for Computational Linguistics (Volume 1: Long Papers)",
    month = jul,
    year = "2025",
    address = "Vienna, Austria",
    publisher = "Association for Computational Linguistics",
    url = "https://aclanthology.org/2025.acl-long.492/",
    doi = "10.18653/v1/2025.acl-long.492",
    pages = "9955--9984",
    ISBN = "979-8-89176-251-0",
}

@article{koo2022k,
  title={K-nct: Korean neural grammatical error correction gold-standard test set using novel error type classification criteria},
  author={Koo, Seonmin and Park, Chanjun and Seo, Jaehyung and Lee, Seungjun and Moon, Hyeonseok and Lee, Jungseob and Lim, Heuiseok},
  journal={IEEE Access},
  volume={10},
  pages={118167--118175},
  year={2022},
  publisher={IEEE}
}

@inproceedings{kim-etal-2024-kogec,
    title = "{K}o{GEC} : {K}orean Grammatical Error Correction with Pre-trained Translation Models",
    author = "Kim, Taeeun  and
      Song, Youngsook  and
      Jeong, Semin",
    editor = "Oco, Nathaniel  and
      Dita, Shirley N.  and
      Borlongan, Ariane Macalinga  and
      Kim, Jong-Bok",
    booktitle = "Proceedings of the 38th Pacific Asia Conference on Language, Information and Computation",
    month = dec,
    year = "2024",
    address = "Tokyo, Japan",
    publisher = "Tokyo University of Foreign Studies",
    url = "https://aclanthology.org/2024.paclic-1.16/",
    pages = "170--180"
}

@book{sohn2001korean,
  title={The korean language},
  author={Sohn, Ho-Min},
  year={2001},
  publisher={Cambridge University Press}
}

@inbook{Cho_2022, 
  place={Cambridge}, 
  series={Cambridge Handbooks in Language and Linguistics}, 
  title={Politeness Strategies in Korean}, 
  booktitle={The Cambridge Handbook of Korean Linguistics}, 
  publisher={Cambridge University Press}, 
  author={Cho, Sungdai}, 
  editor={Cho, Sungdai and Whitman, JohnEditors}, 
  year={2022}, 
  pages={105–132}, 
  collection={Cambridge Handbooks in Language and Linguistics}
}

@article{bhattacharjee2024fighting,
  title={Fighting fire with fire: can ChatGPT detect AI-generated text?},
  author={Bhattacharjee, Amrita and Liu, Huan},
  journal={ACM SIGKDD Explorations Newsletter},
  volume={25},
  number={2},
  pages={14--21},
  year={2024},
  publisher={ACM New York, NY, USA}
}

@inproceedings{li-etal-2024-mage,
    title = "{MAGE}: Machine-generated Text Detection in the Wild",
    author = "Li, Yafu  and
      Li, Qintong  and
      Cui, Leyang  and
      Bi, Wei  and
      Wang, Zhilin  and
      Wang, Longyue  and
      Yang, Linyi  and
      Shi, Shuming  and
      Zhang, Yue",
    editor = "Ku, Lun-Wei  and
      Martins, Andre  and
      Srikumar, Vivek",
    booktitle = "Proceedings of the 62nd Annual Meeting of the Association for Computational Linguistics (Volume 1: Long Papers)",
    month = aug,
    year = "2024",
    address = "Bangkok, Thailand",
    publisher = "Association for Computational Linguistics",
    url = "https://aclanthology.org/2024.acl-long.3/",
    doi = "10.18653/v1/2024.acl-long.3",
    pages = "36--53",
}

@misc{paust_pkot5_v1,
  author       = {Park, Dennis},
  title        = {{pko-T5}: {PAUST} Korean {T5} for Text-to-Text Unified Framework},
  year         = {2022},
  month        = may,
  howpublished = {\url{https://github.com/paust-team/pko-t5}},
  note         = {Version 1.0; Accessed: 2026-05-25}
}

@misc{kogpt2,
  author       = {{SKT-AI}},
  title        = {{KoGPT2}: Korean {GPT-2} Pretrained Cased},
  year         = {2021},
  howpublished = {\url{https://github.com/skt-ai/kogpt2}},
  note         = {Version 2.0; Accessed: 2026-05-25}
}

@misc{an2024exaone35serieslarge,
      title={EXAONE 3.5: Series of Large Language Models for Real-world Use Cases}, 
      author={Soyoung An and Kyunghoon Bae and Eunbi Choi and Kibong Choi and Stanley Jungkyu Choi and Seokhee Hong and Junwon Hwang and Hyojin Jeon and Gerrard Jeongwon Jo and Hyunjik Jo and Jiyeon Jung and Yountae Jung and Hyosang Kim and Joonkee Kim and Seonghwan Kim and Soyeon Kim and Sunkyoung Kim and Yireun Kim and Yongil Kim and Youchul Kim and Edward Hwayoung Lee and Haeju Lee and Honglak Lee and Jinsik Lee and Kyungmin Lee and Woohyung Lim and Sangha Park and Sooyoun Park and Yongmin Park and Sihoon Yang and Heuiyeen Yeen and Hyeongu Yun},
      year={2024},
      eprint={2412.04862},
      archivePrefix={arXiv},
      primaryClass={cs.CL},
      url={https://arxiv.org/abs/2412.04862}, 
}

@misc{park2019g2pk,
  author       = {Park, Kyubyong},
  title        = {{g2pK}: Korean Grapheme-to-Phoneme Conversion},
  year         = {2019},
  howpublished = {\url{https://github.com/Kyubyong/g2pK}},
  note         = {Accessed: 2026-05-25}
}

@inproceedings{qi-etal-2020-stanza,
    title = "{S}tanza: A Python Natural Language Processing Toolkit for Many Human Languages",
    author = "Qi, Peng  and
      Zhang, Yuhao  and
      Zhang, Yuhui  and
      Bolton, Jason  and
      Manning, Christopher D.",
    editor = "Celikyilmaz, Asli  and
      Wen, Tsung-Hsien",
    booktitle = "Proceedings of the 58th Annual Meeting of the Association for Computational Linguistics: System Demonstrations",
    month = jul,
    year = "2020",
    address = "Online",
    publisher = "Association for Computational Linguistics",
    url = "https://aclanthology.org/2020.acl-demos.14/",
    doi = "10.18653/v1/2020.acl-demos.14",
    pages = "101--108",
}

@article{kiwi,
title = {Kiwi: Developing a Korean Morphological Analyzer  Based on Statistical Language Models and Skip-Bigram},
journal = {Korean Journal of Digital Humanities},
volume = {1},
number = {1},
page = {109-136},
year = {2024},
issn = {3058-311X},
doi = {https://doi.org/10.23287/KJDH.2024.1.1.6},
url = {https://kjdh/v.1/1/109/43508},
author = {Minchul Lee},
}

@inproceedings{park2014konlpy,
  title={KoNLPy: Korean natural language processing in Python},
  author={Park, Eunjeong L. and Cho, Sungzoon},
  booktitle={Proceedings of the 26th Annual Conference on Human \& Cognitive Language Technology},
  address={Chuncheon, Korea},
  month={October},
  year={2014}
}

@misc{meinshausen2009stabilityselection,
      title={Stability Selection}, 
      author={Nicolai Meinshausen and Peter Buehlmann},
      year={2009},
      eprint={0809.2932},
      archivePrefix={arXiv},
      primaryClass={stat.ME},
      url={https://arxiv.org/abs/0809.2932}, 
}

@article{guyon2003introduction,
  title={An introduction to variable and feature selection},
  author={Guyon, Isabelle and Elisseeff, Andr{\'e}},
  journal={Journal of machine learning research},
  volume={3},
  number={Mar},
  pages={1157--1182},
  year={2003}
}

@misc{solaiman2019release,
      title={Release Strategies and the Social Impacts of Language Models}, 
      author={Irene Solaiman and Miles Brundage and Jack Clark and Amanda Askell and Ariel Herbert-Voss and Jeff Wu and Alec Radford and Gretchen Krueger and Jong Wook Kim and Sarah Kreps and Miles McCain and Alex Newhouse and Jason Blazakis and Kris McGuffie and Jasmine Wang},
      year={2019},
      eprint={1908.09203},
      archivePrefix={arXiv},
      primaryClass={cs.CL},
      url={https://arxiv.org/abs/1908.09203}, 
}

@inproceedings{uchendu-etal-2021-turingbench-benchmark,
    title = "{TURINGBENCH}: A Benchmark Environment for {T}uring Test in the Age of Neural Text Generation",
    author = "Uchendu, Adaku  and
      Ma, Zeyu  and
      Le, Thai  and
      Zhang, Rui  and
      Lee, Dongwon",
    editor = "Moens, Marie-Francine  and
      Huang, Xuanjing  and
      Specia, Lucia  and
      Yih, Scott Wen-tau",
    booktitle = "Findings of the Association for Computational Linguistics: EMNLP 2021",
    month = nov,
    year = "2021",
    address = "Punta Cana, Dominican Republic",
    publisher = "Association for Computational Linguistics",
    url = "https://aclanthology.org/2021.findings-emnlp.172/",
    doi = "10.18653/v1/2021.findings-emnlp.172",
    pages = "2001--2016",
}

@inproceedings{hans2024binoculars,
author = {Hans, Abhimanyu and Schwarzschild, Avi and Cherepanova, Valeriia and Kazemi, Hamid and Saha, Aniruddha and Goldblum, Micah and Geiping, Jonas and Goldstein, Tom},
title = {Spotting LLMs with binoculars: zero-shot detection of machine-generated text},
year = {2024},
publisher = {JMLR.org},
booktitle = {Proceedings of the 41st International Conference on Machine Learning},
articleno = {698},
numpages = {19},
location = {Vienna, Austria},
series = {ICML'24}
}

@inproceedings{wu-etal-2025-wrote,
    title = "Who Wrote This? The Key to Zero-Shot {LLM}-Generated Text Detection Is {GECS}core",
    author = "Wu, Junchao  and
      Zhan, Runzhe  and
      Wong, Derek F.  and
      Yang, Shu  and
      Liu, Xuebo  and
      Chao, Lidia S.  and
      Zhang, Min",
    editor = "Rambow, Owen  and
      Wanner, Leo  and
      Apidianaki, Marianna  and
      Al-Khalifa, Hend  and
      Eugenio, Barbara Di  and
      Schockaert, Steven",
    booktitle = "Proceedings of the 31st International Conference on Computational Linguistics",
    month = jan,
    year = "2025",
    address = "Abu Dhabi, UAE",
    publisher = "Association for Computational Linguistics",
    url = "https://aclanthology.org/2025.coling-main.684/",
    pages = "10275--10292"
}

@article{chen-etal-2025-repreguard,
    title = "{R}epre{G}uard: Detecting {LLM}-Generated Text by Revealing Hidden Representation Patterns",
    author = "Chen, Xin  and
      Wu, Junchao  and
      Yang, Shu  and
      Zhan, Runzhe  and
      Wu, Zeyu  and
      Luo, Ziyang  and
      Wang, Di  and
      Yang, Min  and
      Chao, Lidia S.  and
      Wong, Derek F.",
    journal = "Transactions of the Association for Computational Linguistics",
    volume = "13",
    year = "2025",
    address = "Cambridge, MA",
    publisher = "MIT Press",
    url = "https://aclanthology.org/2025.tacl-1.81/",
    doi = "10.1162/tacl.a.61",
    pages = "1812--1831"
}

@inproceedings{detectrlx2026,
    title = "{D}etect{RL}-{X}: Towards Reliable Multilingual and Real-World {LLM}-Generated Text Detection",
    author = "Wu, Junchao  and
      Liu, Yefeng  and
      Zhu, Chenyu  and
      Zhang, Hao  and
      Wu, Zeyu  and
      Shi, Tianqi  and
      Du, Yichao  and
      Wang, Longyue  and
      Luo, Weihua  and
      Su, Jinsong  and
      Wong, Derek F.",
    editor = "Liakata, Maria  and
      Moreira, Viviane P.  and
      Zhang, Jiajun  and
      Jurgens, David",
    booktitle = "Proceedings of the 64th Annual Meeting of the {A}ssociation for {C}omputational {L}inguistics (Volume 1: Long Papers)",
    month = jul,
    year = "2026",
    address = "San Diego, California, United States",
    publisher = "Association for Computational Linguistics",
    url = "https://aclanthology.org/2026.acl-long.1773/",
    doi = "10.18653/v1/2026.acl-long.1773",
    pages = "38247--38294",
    ISBN = "979-8-89176-390-6"
}

@inproceedings{wu2024detectrl,
  title={Detect{RL}: {B}enchmarking {LLM}-{G}enerated {T}ext {D}etection in {R}eal-{W}orld {S}cenarios},
  author={Wu, Junchao and Zhan, Runzhe and Wong, Derek F and Yang, Shu and Yang, Xinyi and Yuan, Yulin and Chao, Lidia S},
  booktitle={The Thirty-eight Conference on Neural Information Processing Systems Datasets and Benchmarks Track},
  year={2024}
}

@inproceedings{dugan-etal-2024-raid,
    title = "{RAID}: A Shared Benchmark for Robust Evaluation of Machine-Generated Text Detectors",
    author = "Dugan, Liam  and
      Hwang, Alyssa  and
      Trhl{\'i}k, Filip  and
      Zhu, Andrew  and
      Ludan, Josh Magnus  and
      Xu, Hainiu  and
      Ippolito, Daphne  and
      Callison-Burch, Chris",
    editor = "Ku, Lun-Wei  and
      Martins, Andre  and
      Srikumar, Vivek",
    booktitle = "Proceedings of the 62nd Annual Meeting of the Association for Computational Linguistics (Volume 1: Long Papers)",
    month = aug,
    year = "2024",
    address = "Bangkok, Thailand",
    publisher = "Association for Computational Linguistics",
    url = "https://aclanthology.org/2024.acl-long.674/",
    doi = "10.18653/v1/2024.acl-long.674",
    pages = "12463--12492",
}

@misc{acl2023policy,
    title = {{ACL} 2023 Policy on {AI} Writing Assistance},
    author = {{ACL}},
    year = {2023},
    howpublished = {\url{https://2023.aclweb.org/blog/ACL-2023-policy/}}
}

\clearpage
\appendix

\section{Source Data Details}
\label{app:source_data_details}

\begin{table}[h!]
\centering
\resizebox{\columnwidth}{!}{\begin{tabular}{l|l|rr}
    \toprule
    \textbf{Domain} & \textbf{Source} & \textbf{Avg. Len} & \textbf{Std. Len} \\
    \midrule
    Essay & AI-Hub Essay Evaluation & 629 & 251 \\
    Paper Abs. & KatFish (KIISE 2016--2018) & 407 & 120 \\
    Editorial & AI-Hub Doc.\ Summarization & 1068 & 251 \\
    \bottomrule
\end{tabular}}
\caption{Human-written source texts per \textsc{KoLing} domain. Text length is measured in characters. All domains use an 8:2 train--test split.}
\label{tab:source_domains}
\end{table}

Table~\ref{tab:source_domains} summarizes the human-written source texts of the three \textsc{KoLing} domains, which span a register spectrum from student writing (Essay) through academic prose (Paper Abstract) to professional journalism (Editorial).

\begin{table}[h!]
\centering
\resizebox{\columnwidth}{!}{\begin{tabular}{l|rrr|rr}
    \toprule
    \textbf{Writing Type} & \textbf{Total} & \textbf{Train} & \textbf{Test} & \textbf{Avg. Len} & \textbf{Std. Len} \\
    \midrule
    Creative & 300 & 240 & 60 & 671 & 303 \\
    Problem-Solving & 300 & 240 & 60 & 730 & 236 \\
    Expository & 300 & 240 & 60 & 504 & 213 \\
    Argumentative & 300 & 240 & 60 & 625 & 209 \\
    Pro-Con & 300 & 240 & 60 & 613 & 223 \\
    \midrule
    \textbf{Total} & \textbf{1500} & \textbf{1200} & \textbf{300} & \textbf{629} & \textbf{251} \\
    \bottomrule
\end{tabular}}
\caption{Statistics of the source human essay dataset used in \textsc{KoLing}. Text length is measured in characters.}
\label{tab:source_essays}
\end{table}

\paragraph{Essay.}
The essay domain is built from the AI-Hub ``Essay Writing Evaluation (에세이 글 평가)'' dataset, a Korean educational corpus introduced for automatic essay scoring and related educational services.\footnote{This research used datasets from 'The Open AI Dataset Project (AI-Hub, S. Korea)'. All data information can be accessed through 'AI-Hub (www.aihub.or.kr)'. The official dataset is publicly accessible through \href{https://aihub.or.kr/aihubdata/data/view.do?currMenu=115&topMenu=100&dataSetSn=545}{this page}.} According to the dataset description, the full resource contains essays written by elementary, middle, and high school students, paired with expert scoring information and rubric-based annotations across 50 prompts. AI-Hub reports more than 50k source essays in the full release, collected to support essay evaluation research and EduTech applications. We do not use the full corpus: we sample 1{,}500 human-written essays and rebalance them into five writing types used throughout our benchmark (Creative, Problem-Solving, Expository, Argumentative, Pro-Con; Table~\ref{tab:source_essays}). This controlled subset design helps equalize genre coverage before LLM modification, so that benchmark difficulty is shaped primarily by the intervention levels rather than by source-data imbalance. Each essay retains its original prompt materials (topic statement, reading passage, and writer grade level), which we reuse as generation input for the LGT class (Appendix~\ref{app:prompts}).

\paragraph{Paper Abstract.}
We reuse the human source articles released with KatFish~\cite{park-etal-2025-katfishnet}: Korean computer-science papers (KIISE, 2016--2018) with author-written abstracts and full paper bodies. We use only the human abstracts (as HWT) and the paper bodies (as grounding input for LGT generation); every LLM-side text is generated anew by our pipeline, so no KatFish-generated text enters \textsc{KoLing}. Because KatFish also serves as a cross-benchmark in \S\ref{sec:transfer}, we deliberately reuse only its human source material to avoid circularity.

\paragraph{Editorial.}
We draw newspaper editorials from the AI-Hub ``Document Summarization Text (문서요약 텍스트)'' dataset. Because the released reference summaries are short ($\sim$120 characters), we use the full article bodies (mean 1{,}068 characters) as HWT and the editorial titles as the topic input for LGT generation. Editorials are professionally written and consistently use the formal declarative register (`-다' style), providing a register contrast to essays.

\section{Generation Prompts}
\label{app:prompts}

\paragraph{HLT (modification).}
Table~\ref{tab:generation_prompts} presents the system message and user prompts for the three modification levels (Proofread, Polish, Paraphrase) used to construct the HLT class. All prompts are identical across LLM backends (GPT-4o~\cite{gpt-4o}, Claude-Sonnet-4.6~\cite{claude-4.6}, Solar-Pro3~\cite{solar-pro3}, Qwen2.5-32B-Instruct~\cite{qwen2025qwen25technicalreport}, Mistral-Small-24B-Instruct~\cite{mistral-3.1}). Generation temperature is set to 0.7 for all levels. For the two formal domains (Paper Abstract, Editorial), whose human texts consistently use the plain declarative register (`-다' style), we append a register-preservation clause to every modification prompt so that models do not drift into the polite register: \begin{CJK}{UTF8}{mj}``반드시 원문과 동일한 문장 종결 방식을 유지하세요. 원문이 `-다' 평서체이면 결과도 모든 문장을 `-다' 평서체로 쓰고, 존댓말(`-습니다', `-요')로 절대 바꾸지 마세요.''\end{CJK} (\textit{``Always keep the same sentence-ending style as the original. If the original uses the plain `-다' declarative, write every sentence in the same declarative style and never switch to the polite `-습니다'/`-요' register.''})

\paragraph{LGT (from-scratch generation).}
The LGT class follows a single principle: \emph{the LLM receives the same input a human author received}. Table~\ref{tab:lgt_prompts} lists the three domain-specific prompts. For Paper Abstract, the model writes an abstract grounded in the full paper body; for Editorial, it writes an editorial from the title alone; for Essay, it adopts a student persona and writes from the same topic statement, reading passage, and grade level given to the student. Each prompt targets the original human text's length, and the two formal domains enforce the `-다' declarative register in the prompt itself.

\begin{table*}[t]
\centering
\small
\begin{tabular}{@{}p{0.14\textwidth} p{0.80\textwidth}@{}}
\toprule
\textbf{Input Regime} & \textbf{Prompt} \\
\midrule
\textbf{System Message}
&
\textbf{[KO]}
당신은 한국어 글쓰기 전문가입니다. 주어진 지시에 따라 글을 작성하세요. 작성한 글만 출력하고, 부가 설명이나 메타 정보, 제목 표지는 포함하지 마세요.
\newline\newline
\textbf{[EN]}
\textit{You are a Korean writing expert. Write a text according to the given instructions. Output only the written text, without any additional explanation, meta-information, or title page.} \\
\midrule
\textbf{Grounded} \newline (Paper Abs.)
&
\textbf{[KO]}
다음은 한 학술 논문의 본문입니다(초록 제외). 이 논문을 대표하는 초록(abstract)을 작성해 주세요. 연구의 배경, 목적, 방법, 결과가 드러나도록 학술 논문 초록의 문어체로 작성하되, 반드시 `-다'로 끝나는 평서체(예: `제안한다', `분석하였다')로 쓰고 존댓말(`-습니다', `-요')은 절대 사용하지 마세요. 분량은 한국어 기준 약 \{length\}자 내외로 맞춰 주세요. \newline 본문: \{body\}
\newline\newline
\textbf{[EN]}
\textit{The following is the body of an academic paper (abstract excluded). Write an abstract that represents this paper. Use the formal written style of academic abstracts, covering the background, objective, method, and results. Always use the plain `-다' declarative ending and never the polite register. Keep the length to about \{length\} Korean characters.} \newline \textit{Body: \{body\}} \\
\midrule
\textbf{Topic} \newline (Editorial)
&
\textbf{[KO]}
다음 제목으로 신문 사설을 작성해 주세요. 하나의 논지를 명확히 세우고 근거를 들어 설득력 있게 전개하는 신문 사설의 문어체로 작성하되, 반드시 `-다'로 끝나는 평서체로 쓰고 존댓말(`-습니다', `-요')은 절대 사용하지 마세요. 분량은 한국어 기준 약 \{length\}자 내외로 맞춰 주세요. \newline 제목: \{title\}
\newline\newline
\textbf{[EN]}
\textit{Write a newspaper editorial under the following title. Establish one clear thesis and develop it persuasively with supporting arguments, in the formal written style of newspaper editorials. Always use the plain `-다' declarative ending and never the polite register. Keep the length to about \{length\} Korean characters.} \newline \textit{Title: \{title\}} \\
\midrule
\textbf{Persona} \newline (Essay)
&
\textbf{[KO]}
당신은 \{grade\} 학생입니다. 다음 논제와 제시문을 바탕으로 `\{type\}'을 작성해 주세요. 제시문의 내용을 참고하되 자신의 생각을 담아 쓰고, 분량은 한국어 기준 약 \{length\}자 내외로 맞춰 주세요. \newline 논제: \{topic\} \newline 제시문: \{passage\}
\newline\newline
\textbf{[EN]}
\textit{You are a \{grade\} student. Based on the following topic and reading passage, write a \{type\} essay. Refer to the passage but express your own thoughts, and keep the length to about \{length\} Korean characters.} \newline \textit{Topic: \{topic\}} \newline \textit{Passage: \{passage\}} \\
\bottomrule
\end{tabular}
\caption{System message and domain-specific user prompts for LGT (from-scratch) generation. \{length\} is set to the original human text's character length; \{type\} is the essay writing type (e.g., argumentative); the grounded/topic/persona input regimes mirror what the human author of each domain received.}
\label{tab:lgt_prompts}
\end{table*}

\section{Feature Explanation}
\label{app:feature_exploration}

Table~\ref{tab:feature_detail} provides the full specification of all 47 linguistic features used in \textsc{KoLingProbe}. Features are grouped by their theoretical linguistic category and sub-category. Each feature is computed at the text level, taking a single essay as input and producing a scalar value. For phonological features, we use a grapheme-to-phoneme (G2P) converter and pre-compiled pronunciation dictionaries built from a large Korean corpus. Morphological and syntactic features are extracted using the Kiwi morphological analyzer~\cite{kiwi}. Semantic collocation features rely on a noun--josa--predicate co-occurrence dictionary constructed from Korean Wikipedia, scored using PMI and LLR statistics. Pragmatic features are based on curated lexicons of speech-level markers, hedging expressions, and discourse markers.

\section{Training Details}
\label{app:training_details}

This section provides the full training details for \textsc{KoLingProbe} described in \S~\ref{individual_training}.

\paragraph{Stability Selection.}
Before training each sub-detector, we apply stability selection~\cite{meinshausen2009stabilityselection} to identify a robust feature subset. For each sub-detector $m$ with feature dimensionality $d_m$, we run nested cross-validation: the outer loop uses 5-fold stratified splits across multiple random seeds, and the inner loop uses 3-fold CV to select the best ElasticNet hyperparameters ($C \in \{0.1, 0.3, 1.0, 3.0\}$, $\ell_1$-ratio $\in \{0.0, 0.2, 0.5, 0.8\}$) by maximizing AUROC. After training ElasticNet on each outer fold with the best hyperparameters, we record which features have non-zero coefficients ($|\theta_j| > 10^{-12}$). Only features selected in ${\geq}70\%$ of all outer folds across all seeds are retained, yielding a binary mask $\mathbf{s}_m \in \{0,1\}^{d_m}$. This procedure ensures that selected features are consistently informative regardless of data perturbation.

\paragraph{Sub-Detector Training.}
Each sub-detector is a logistic regression with ElasticNet penalty, trained on its stability-selected features. We use 5-fold stratified cross-validation to generate \emph{out-of-fold} (OOF) predictions for every training sample: in each fold, the model is trained on 4 folds and produces probability predictions $\hat{p}_m(x_i) \in [0, 1]$ for the held-out fold. After completing all folds, every training sample has exactly one OOF prediction from a fold where it served as validation data. For test-time prediction, a final model is trained on the full training set.

\paragraph{Meta-Learner Training.}
The OOF predictions from all five sub-detectors are concatenated into a matrix $\mathbf{P} \in \mathbb{R}^{n \times 5}$, where each row contains one sample's five sub-detector probabilities. A logistic regression meta-learner ($C{=}1.0$, solver: LBFGS) is trained on $\mathbf{P}$ to learn the optimal combination weights $\alpha_1, \ldots, \alpha_5$. Because $\mathbf{P}$ consists entirely of OOF predictions, no sample's own training-fold predictions influence the meta-learner, preventing data leakage.

\paragraph{Hyperparameters.}
Stability selection: outer folds = 5, inner folds = 3, $\tau = 0.7$, solver = SAGA, max iterations = 4,000. Sub-detectors and meta-learner: max iterations = 4,000.
All experiments are repeated over 5 random seeds (42--46), and we report the mean.


\paragraph{Hardware.}
All \textsc{KoLingProbe} stages, feature extraction, sub-detector training, meta-learner training, and inference, run on CPU only and require no GPU resources. Baselines that depend on neural inference or fine-tuning (Confidence-based scoring, Perturbation, LLM-as-Detector, XLM-RoBERTa, XDAC-D) use a single NVIDIA RTX A5000 (24\,GB) GPU.

\paragraph{Baseline Hyperparameters.}
\emph{Confidence-based} methods (Log-Likelihood, Rank, Log-Rank, Entropy, LRR) use KoGPT2 (\texttt{skt/kogpt2-base-v2}) as the scoring model with no additional tuning. These methods were originally defined over GPT-2 scores; KoGPT2 is the de facto standard open Korean counterpart, so this choice gives the baselines a scorer native to the target language. \emph{Perturbation} baselines (NPR, DetectGPT, Fast-DetectGPT) use \texttt{paust/pko-t5-base} for mask-fill with \texttt{n\_perturbations}=50, \texttt{span\_length}=2, and \texttt{anchor\_window}=3; Fast-DetectGPT additionally uses a Hutchinson estimator with $\sigma=0.01$ and $n=10$ samples. \emph{LLM-as-Detector} uses EXAONE-3.5-7.8B-Instruct: the 0-shot setting extracts next-token probabilities greedily, and the 10-shot setting samples five balanced human/LLM demonstrations per query with random ordering. \emph{Binoculars} uses Llama-3.1-8B as the observer and Llama-3.1-8B-Instruct as the performer (shared tokenizer), scoring each text by the ratio of observer log-perplexity to cross-perplexity at max sequence length 512. \emph{GECScore} applies the Korean grammatical-error-correction model \texttt{j5ng/et5-typos-corrector} with greedy decoding and uses the normalized edit distance between the input and its correction as the detection score. \emph{RepreGuard} is adapted as a frozen-encoder representation probe: the backbone \texttt{Hello-SimpleAI/chatgpt-detector-roberta} is kept frozen, hidden states are mean-pooled at the most discriminative layer (selected among $\{L/3, L/2, 2L/3, L{-}1, L\}$ by 5-fold cross-validated AUROC averaged over 5 seeds), and a logistic-regression probe ($C{=}1.0$) is trained on the pooled representations, the same supervised protocol as \textsc{KoLingProbe}. \emph{XLM-RoBERTa} (\texttt{FacebookAI/xlm-roberta-base}) is fine-tuned on each task's training split with the HuggingFace \texttt{Trainer} default optimizer (AdamW), learning rate $2\!\times\!10^{-5}$, batch size 32, max sequence length 512, and 3 epochs over 5 random seeds. \emph{XDAC-D} is used as a pretrained Korean detector without further fine-tuning. \emph{KatFishNet} uses its comma-usage feature module with Kkma-based preprocessing~\cite{park2014konlpy} and the same lightweight logistic-regression protocol as \textsc{KoLingProbe}.

\section{Efficiency}
\label{app:inference_time}

\begin{figure}[h!]
  \centering
  \includegraphics[width=1.0\linewidth]{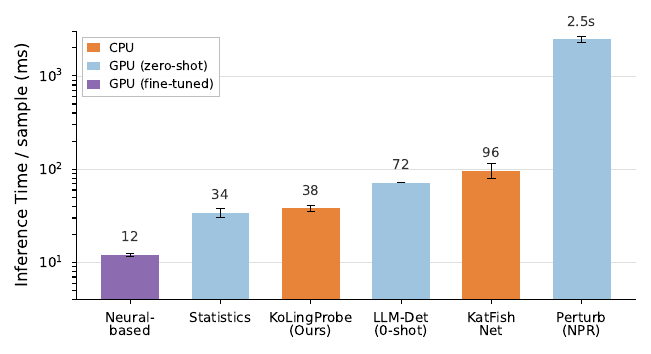}
  \caption{Per-sample inference latency (single sample, batch{=}1; log scale).
  Colors denote CPU, GPU zero-shot, and GPU task-fine-tuned methods.
  \textsc{KoLingProbe} (38\,ms, CPU) needs no GPU and is far faster than the strongest
  zero-shot competitors (e.g., NPR 2{,}476\,ms).}
  \label{fig:inference_time}
\end{figure}

Figure~\ref{fig:inference_time} compares per-sample inference latency at batch size~1, grouping methods by hardware and task-training regime. Perturbation-based methods are the slowest: NPR requires generating 50 perturbed texts via a T5 model on GPU, taking 2{,}476\,ms per sample. Fine-tuned neural detectors achieve fast inference through a single forward pass, but require GPU resources and a one-time task fine-tuning pass.

\textsc{KoLingProbe} (38\,ms) is 2.5$\times$ faster than KatFishNet (96\,ms) despite extracting features from five modules rather than one. This is because KatFishNet relies on Kkma~\cite{park2014konlpy}, a Java-based morphological analyzer, whereas \textsc{KoLingProbe} uses Kiwi~\cite{kiwi}, a C++-based analyzer that offers both high accuracy and efficiency. The entire \textsc{KoLingProbe} pipeline, from feature extraction to the final meta-learner decision, runs on CPU with lightweight logistic regression classifiers, making it a practical solution for low-resource and large-scale deployment scenarios.

Three caveats apply to this comparison: (i) latency is measured at batch{=}1, and GPU throughput improves substantially with batching; (ii) the reported times exclude the one-time GPU \emph{fine-tuning} cost incurred by the fine-tuned detectors (e.g., XLM-RoBERTa); and (iii) hardware differs across methods by design, so the figure reflects deployment cost rather than raw computational speed.

\clearpage

\begin{table*}[t]
\centering
\small
\begin{tabular}{@{}p{0.14\textwidth} p{0.80\textwidth}@{}}
\toprule
\textbf{Level} & \textbf{Prompt} \\
\midrule
\textbf{System Message}
&
\textbf{[KO]}
당신은 한국어 글쓰기 전문가입니다. 주어진 지시에 따라 텍스트를 변환하세요. 변환된 글만 출력하고, 부가 설명이나 메타 정보는 포함하지 마세요.
\newline\newline
\textbf{[EN]}
\textit{You are a Korean writing expert. Transform the text according to the given instructions. Output only the transformed text without any additional explanation or meta-information.} \\
\midrule
\textbf{Proofread} \newline (Level~1)
&
\textbf{[KO]}
다음 글의 맞춤법, 문법 오류를 교정하고, 반복적이거나 부적절한 단어를 교체해 주세요. 원문의 문체와 어조를 유지하고, 전체 내용과 구조는 변경하지 마세요. 분량은 원문과 비슷하게 유지하세요. \newline 원문: \{text\}
\newline\newline
\textbf{[EN]}
\textit{Correct any spelling and grammar errors in the following text and replace any repetitive or inappropriate words. Maintain the original style and tone, and do not change the overall content or structure. Keep the length similar to the original.} \newline \textit{Original: \{text\}} \\
\midrule
\textbf{Polish} \newline (Level~2)
&
\textbf{[KO]}
다음 글의 문장들을 유창하고 명확하게 다듬어 주세요. 전체 내용과 구조는 유지하되, 표현을 개선해 주세요. 원문의 문체와 어조를 유지하고, 분량은 원문과 비슷하게 유지하세요. \newline 원문: \{text\}
\newline\newline
\textbf{[EN]}
\textit{Refine the sentences in the following text to be fluent and clear. Maintain the overall content and structure, but improve the expressions. Preserve the original style and tone, and keep the length similar to the original.} \newline \textit{Original: \{text\}} \\
\midrule
\textbf{Paraphrase} \newline (Level~3)
&
\textbf{[KO]}
다음 글을 같은 의미를 전달하되, 문장 구조와 표현을 완전히 다르게 재작성해 주세요. 원문의 단어와 문장 구조를 그대로 사용하지 말고, 다른 어휘와 문형으로 바꿔 주세요. 전체 내용과 논지는 유지하되, 분량은 원문과 비슷하게 유지하세요. \newline 원문: \{text\}
\newline\newline
\textbf{[EN]}
\textit{Rewrite the following text to convey the same meaning, but with completely different sentence structures and expressions. Do not reuse the original words or sentence patterns; use different vocabulary and syntactic forms. Maintain the overall content and argument, and keep the length similar to the original.} \newline \textit{Original: \{text\}} \\
\bottomrule
\end{tabular}
\caption{System message and user prompts for each modification level used in the \textsc{KoLing} benchmark. \{text\} is replaced with the original human-written essay.}
\label{tab:generation_prompts}
\end{table*}

\clearpage

\begin{table*}[t]
\centering
\resizebox{\textwidth}{!}{\begin{tabular}{p{1.8cm}p{1.8cm}lp{11.5cm}}
    \toprule
    Category & Sub-category & Feature & Description \\
    \midrule

    \multirow{12}{*}{\makecell[l]{Phonology\\(12d)}}
     & \multirow{6}{*}{\makecell[l]{G2P\\Rules\\(6d)}}
       & \texttt{ratio\_spell\_pron}       & Ratio of content words whose spelling matches their G2P-converted pronunciation. \\
     & & \texttt{n\_palatalization}         & Count of words exhibiting palatalization (ㄷ/ㅌ + ㅣ $\rightarrow$ ㅈ/ㅊ). \\
     & & \texttt{n\_tensification}          & Count of words exhibiting tensification (lenis $\rightarrow$ fortis after obstruent codas). \\
     & & \texttt{n\_liquid\_assimilation}   & Count of words with liquid assimilation (ㄴ+ㄹ or ㄹ+ㄴ $\rightarrow$ ㄹㄹ). \\
     & & \texttt{n\_nasal\_assimilation}    & Count of words with nasal place assimilation. \\
     & & \texttt{n\_initial\_law}           & Count of initial-law (두음법칙) violations in spelling. \\
    \cmidrule(lr){2-4}
     & \multirow{6}{*}{\makecell[l]{Syllable\\Structure\\(6d)}}
       & \texttt{cvc\_rate}                & Proportion of syllables with a coda (CVC structure). \\
     & & \texttt{complex\_coda\_rate}       & Proportion of syllables with a complex (double) coda. \\
     & & \texttt{sonorant\_coda\_rate}      & Proportion of syllables ending with a sonorant coda (ㄴ, ㄹ, ㅇ, ㅁ). \\
     & & \texttt{p\_coda\_n}               & Proportion of syllables ending with coda ㄴ. \\
     & & \texttt{p\_coda\_l}               & Proportion of syllables ending with coda ㄹ. \\
     & & \texttt{mean\_syll\_per\_word}     & Average number of Hangul syllables per whitespace-delimited word. \\
    \midrule

    \multirow{11}{*}{\makecell[l]{Morphology\\(11d)}}
     & \multirow{5}{*}{\makecell[l]{Spacing\\(5d)}}
       & \texttt{pos\_seq\_diversity}       & Type-token ratio of POS tag sequences, measuring syntactic pattern diversity. \\
     & & \texttt{space\_ratio\_mmn}         & Correct spacing ratio for modifier + noun (MM+N) patterns. \\
     & & \texttt{space\_ratio\_nnb}         & Correct spacing ratio for noun + dependent noun (N+NB) patterns. \\
     & & \texttt{space\_ratio\_vx}          & Correct spacing ratio for verb + auxiliary (V+VX) patterns. \\
     & & \texttt{vx\_diversity}             & Type-token ratio of verb + auxiliary predicate combinations. \\
    \cmidrule(lr){2-4}
     & \multirow{6}{*}{\makecell[l]{Morpho-\\logical\\(6d)}}
       & \texttt{josa\_entropy}            & Shannon entropy of case marker (josa) type distribution. \\
     & & \texttt{josa\_rare\_ratio}         & Ratio of rare case markers (complementizer, adnominal, quotative) to all case markers. \\
     & & \texttt{eomi\_diversity}           & Type-token ratio of verbal ending (eomi) forms. \\
     & & \texttt{eomi\_connective\_ratio}   & Ratio of connective endings among all verbal endings. \\
     & & \texttt{morph\_per\_eojeol}        & Average number of morphemes per eojeol (spacing unit). \\
     & & \texttt{func\_content\_ratio}      & Ratio of functional morphemes (josa, eomi, affixes) to content morphemes (nouns, verbs). \\
    \midrule

    \multirow{9}{*}{\makecell[l]{Syntax\\(9d)}}
     & \multirow{5}{*}{\makecell[l]{Comma\\Usage\\(5d)}}
       & \texttt{comma\_inclusion\_rate}    & Proportion of sentences containing at least one comma. \\
     & & \texttt{comma\_usage\_rate}        & Average ratio of commas to total morphemes per sentence. \\
     & & \texttt{comma\_rel\_position}      & Average relative position of commas within a sentence (0--1). \\
     & & \texttt{comma\_seg\_length}        & Average segment length between consecutive commas. \\
     & & \texttt{comma\_pos\_diversity}     & Type-token ratio of POS bigrams surrounding commas. \\
    \cmidrule(lr){2-4}
     & \multirow{4}{*}{\makecell[l]{Syntactic\\Structure\\(4d)}}
       & \texttt{sent\_len\_mean}          & Mean sentence length in eojeol (word) count. \\
     & & \texttt{sent\_len\_std}            & Standard deviation of sentence lengths. \\
     & & \texttt{sent\_len\_entropy}        & Shannon entropy of binned sentence lengths ([1--5], [6--10], [11--20], [21--40], [41+]). \\
     & & \texttt{clause\_density}           & Average number of connective endings per sentence, measuring clausal complexity. \\
    \midrule

    \multirow{9}{*}{\makecell[l]{Semantics\\(9d)}}
     & \multirow{5}{*}{\makecell[l]{Collo-\\cation\\(5d)}}
       & \texttt{njp\_pairs\_per\_kchar}   & Density of noun-josa-predicate collocations per 1K characters. \\
     & & \texttt{njp\_pmi\_mean}            & Mean Pointwise Mutual Information of noun-josa-predicate pairs. \\
     & & \texttt{njp\_pmi\_median}          & Median PMI of noun-josa-predicate pairs. \\
     & & \texttt{njp\_llr\_mean}            & Mean Log-Likelihood Ratio of noun-josa-predicate pairs. \\
     & & \texttt{njp\_llr\_median}          & Median LLR of noun-josa-predicate pairs. \\
    \cmidrule(lr){2-4}
     & \multirow{4}{*}{\makecell[l]{Word\\Frequency\\(4d)}}
       & \texttt{freq\_mean\_log}          & Mean of log-transformed token frequencies based on Korean Wikipedia. \\
     & & \texttt{freq\_std\_log}            & Standard deviation of log-transformed token frequencies. \\
     & & \texttt{freq\_low\_ratio}          & Ratio of rare tokens (below the 10th percentile in corpus frequency). \\
     & & \texttt{freq\_hapax\_ratio}        & Ratio of hapax legomena and out-of-vocabulary tokens. \\
    \midrule

    \multirow{6}{*}{\makecell[l]{Pragmatics\\(6d)}}
     & \multirow{2}{*}{\makecell[l]{Speech\\Level (2d)}}
       & \texttt{sl\_formal\_ratio}        & Ratio of sentences using the formal speech level (\textit{hapsyo-che}). \\
     & & \texttt{sl\_entropy}               & Shannon entropy over the distribution of speech levels (5 categories). \\
    \cmidrule(lr){2-4}
     & \makecell[l]{Hedging\\(1d)}
       & \texttt{hedge\_ratio}             & Average count of epistemic hedging expressions (e.g., \textit{-kes kath-}, \textit{ama}) per sentence. \\
    \cmidrule(lr){2-4}
     & \multirow{2}{*}{\makecell[l]{Discourse\\Markers (2d)}}
       & \texttt{dm\_density}              & Density of discourse markers per 1K characters. \\
     & & \texttt{dm\_formal\_ratio}         & Ratio of formal discourse markers among all discourse markers. \\
    \cmidrule(lr){2-4}
     & \makecell[l]{Sent.\\Endings (1d)}
       & \texttt{se\_diversity}            & Type-token ratio of sentence-final morphological forms (last 3 syllables). \\

    \bottomrule
\end{tabular}}
\caption{Detailed description of all 47 linguistic features. Each feature is computed at the text level.}
\label{tab:feature_detail}
\end{table*}

\clearpage

\section{Which Linguistic Features Matter? Feature-Level Effect Sizes on KatFish and \textsc{KoLing}}
\label{app:feature_effects}
\begin{figure}[t!]
    \centering
    \subfigure[Top 10 Features (KatFish)]{\includegraphics[width=0.49\textwidth]{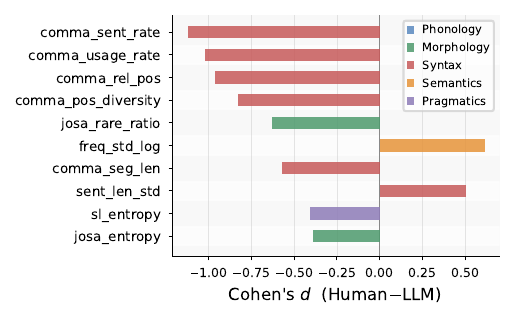}}
    \subfigure[Top 10 Features (\textsc{KoLing})]{\includegraphics[width=0.49\textwidth]{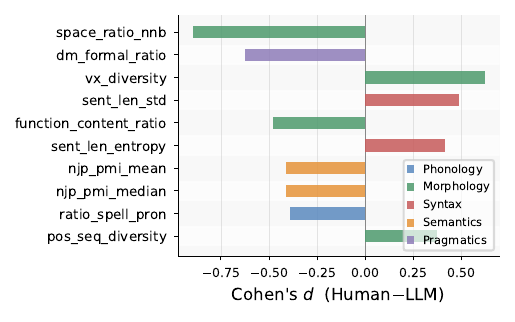}}
    \caption{Top-10 features by Cohen's $d$ on (a) KatFish and (b) \textsc{KoLing}. KatFish is dominated by syntax, while \textsc{KoLing} draws on all five sub-detectors. All features shown are significant at $p < .001$ (Welch's $t$-test).}
    \label{fig:cohen_comparison}
    \vspace{-2px}
\end{figure}

Figure~\ref{fig:cohen_comparison} compares the top-10 discriminative features by Cohen's $d$ on each benchmark. On KatFish, the ranking is dominated by syntactic features such as comma inclusion, which explains why surface-level methods perform well there. On \textsc{KoLing}, by contrast, all five linguistic categories appear in the top-10, indicating that bias control reduces reliance on a narrow set of surface cues and exposes more diverse human--LLM differences.

The strongest effects include \texttt{space\_ratio\_nnb} and \texttt{vx\_diversity} (morphology), reflecting spacing regularization and limited auxiliary-verb variation in LLM-generated texts, and \texttt{dm\_formal\_ratio} (pragmatics), capturing formal discourse-marker overuse. Syntactic (\texttt{sent\_len\_std}), semantic (\texttt{njp\_pmi\_mean}), and phonological (\texttt{ratio\_spell\_pron}) features also remain salient. Together, these patterns suggest that, unlike KatFish, \textsc{KoLing} requires evidence across all five linguistic levels rather than a single syntactic shortcut.

\section{Concatenated Features vs.\ Hierarchical Fusion}
\label{app:concat_ablation}

We compare \textsc{KoLingProbe}'s decision-level fusion against a single logistic regression trained on the concatenated 47-dimensional feature vector (\emph{Concat-47}), using the identical pipeline: the same stability selection (applied to the concatenated vector), hyperparameter grid, training source, and 5-generator $\times$ 5-seed protocol. Table~\ref{tab:concat_ablation} reports AUROC on clean text and under the character-level typo attack of \S\ref{sec:exp_finetuned} (Essay domain); the same pattern holds under macro-F1.

On clean text the two designs are comparable: Concat-47 is slightly higher on overall average (82.2 vs.\ 81.1), while the fusion leads on Editorial (both tasks) and the Paper Abstract HLT task, where a monolithic 47-dimensional classifier is more prone to overfitting. The difference emerges under corruption: as the typo rate grows, Concat-47 declines steeply and collapses to chance on LGT at 10\% injection (46.5 AUROC), whereas the fusion degrades gracefully and retains 68.8, because corruption entering one linguistic level cannot propagate through the other sub-detectors' decisions. On the harder HLT task both designs eventually collapse at 10\%. Beyond robustness, only the modular design supports the per-level analyses of \S\ref{sec:interpretability}, which are undefined for a monolithic classifier.

\begin{table}[h!]
\centering
\resizebox{\columnwidth}{!}{
\begin{tabular}{l|ccc|ccc|c}
    \toprule
    \multirow{2}{*}{\textbf{Design}} & \multicolumn{3}{c|}{\textbf{HLT vs.\ HWT}} & \multicolumn{3}{c|}{\textbf{LGT vs.\ HWT}} & \multirow{2}{*}{\textbf{Avg.}} \\
    & Essay & Abs. & Edit. & Essay & Abs. & Edit. & \\
    \midrule
    Concat-47 & \textbf{87.6} & 64.9 & 69.2 & \textbf{93.1} & \textbf{80.6} & 97.8 & \textbf{82.2} \\
    Fusion (ours) & 85.1 & \textbf{66.9} & \textbf{70.1} & 91.2 & 74.0 & \textbf{99.6} & 81.1 \\
    \midrule
    \multicolumn{8}{c}{\textit{Under typo attack (Essay); columns = injection rate}} \\
    \midrule
    & \multicolumn{3}{c|}{HLT: 1\% ~/~ 3\% ~/~ 5\% ~/~ 10\%} & \multicolumn{3}{c|}{LGT: 1\% ~/~ 3\% ~/~ 5\% ~/~ 10\%} & \\
    \midrule
    Concat-47 & \multicolumn{3}{c|}{\textbf{83.8} / \textbf{75.1} / \textbf{66.6} / \textbf{46.7}} & \multicolumn{3}{c|}{\textbf{89.3} / 80.0 / 69.1 / 46.5} & \\
    Fusion (ours) & \multicolumn{3}{c|}{79.1 / 67.6 / 58.1 / 41.8} & \multicolumn{3}{c|}{87.3 / \textbf{81.0} / \textbf{75.5} / \textbf{68.8}} & \\
    \bottomrule
\end{tabular}}
\caption{Concatenated 47-feature classifier vs.\ hierarchical fusion (AUROC \%, same protocol as the main tables). Top: clean \textsc{KoLing}. Bottom: Essay under increasing typo injection.}
\label{tab:concat_ablation}
\end{table}

\section{Typo Attack Construction}
\label{app:typo_robustness}

We construct typo-attacked test sets by perturbing every LLM-side (HLT/LGT) test instance with three types of Korean character-level noise, while keeping training data and human-written texts unchanged: (i) adjacent-consonant jamo swaps (e.g., \begin{CJK}{UTF8}{mj}ㅈ\end{CJK}↔\begin{CJK}{UTF8}{mj}ㅊ\end{CJK}, \begin{CJK}{UTF8}{mj}ㄱ\end{CJK}↔\begin{CJK}{UTF8}{mj}ㅋ\end{CJK}), (ii) random space insertion/deletion, and (iii) initial-law violations (\begin{CJK}{UTF8}{mj}률\end{CJK}↔\begin{CJK}{UTF8}{mj}율\end{CJK}, \begin{CJK}{UTF8}{mj}례\end{CJK}↔\begin{CJK}{UTF8}{mj}예\end{CJK}). Each noise type is applied sequentially at the target rate (0--10\% in the sweep of \S\ref{sec:exp_finetuned}) with a fixed random seed, and every detector is trained on the original clean training set and evaluated on the perturbed test sets. This simulates an adversary who masks LLM-generated text by injecting human-like surface noise.

\section{Shortcut-Controlled KatFish}
\label{app:katfish_clean}

\begin{table}[h!]
\centering
\resizebox{\columnwidth}{!}{
\begin{tabular}{l|ccc|c}
    \toprule
    \textbf{Method} & \textbf{Essay} & \textbf{Poetry} & \textbf{Abstract} & \textbf{Avg.} \\
    \midrule
    Log-Likelihood & 51.6 $\to$ 48.3 & 43.3 $\to$ 46.9 & 64.1 $\to$ 56.7 & 53.0 $\to$ 50.7 \\
    Binoculars & 96.4 $\to$ 93.8 & 63.1 $\to$ 54.7 & 68.6 $\to$ 68.0 & 76.0 $\to$ 72.2 \\
    KatFishNet & 95.6 $\to$ 81.3 & 72.1 $\to$ 56.8 & 74.0 $\to$ 64.9 & 80.6 $\to$ 67.7 \\
    \cellcolor[HTML]{EAECFF}\textbf{\textsc{KoLingProbe}} & \cellcolor[HTML]{EAECFF}98.1 $\to$ \textbf{93.7} & \cellcolor[HTML]{EAECFF}79.3 $\to$ \textbf{74.6} & \cellcolor[HTML]{EAECFF}74.8 $\to$ \textbf{66.4} & \cellcolor[HTML]{EAECFF}84.1 $\to$ \textbf{78.2} \\
    \bottomrule
\end{tabular}}
\caption{AUROC (\%) on KatFish, original $\to$ shortcut-controlled (length- and comma-matched, class-balanced subsets). Best shortcut-controlled result per column is \textbf{bolded}.}
\label{tab:katfish_clean}
\end{table}

To verify that \textsc{KoLingProbe}'s lead on KatFish (Table~\ref{tab:katfish_results}) is not itself shortcut-driven, we apply \textsc{KoLing}-style shortcut control to KatFish: within every test split, human and LLM samples are matched on text length (quartile bins) and comma count, with classes balanced by subsampling per bin (per seed). Table~\ref{tab:katfish_clean} re-evaluates representative methods on the matched subsets. \textsc{KoLingProbe} loses only 5.9 AUROC on average and stays ahead in every genre, whereas the comma-based KatFishNet collapses by 13.2; the training-free methods also fall behind. Because KatFish separates classes strongly by length and commas, the matched subsets are necessarily small, so we read these as trend evidence over five seeds rather than precise point estimates.

\section{All Results for Sub-Detector Combinations of \textsc{KoLingProbe} on \textsc{KoLing}}
\label{app:all_results_koling}

Table~\ref{tab:ablation_koling} presents the exhaustive ablation over all 31 possible sub-detector combinations ($2^5-1$) per domain and task (same protocol as the main tables). While Section~\ref{sec:interpretability} reports greedy forward selection (Figure~\ref{fig:greedy_domains}), this table enables a more granular analysis of how each linguistic level interacts with others. Key observations: (1) Morphology is by far the strongest individual sub-detector in every domain (average 77.8 vs.\ 60.5--65.6 for the other single modules), consistent with Korean's agglutinative structure providing rich detection signals. (2) The full five-module model attains the best global average (81.1) and the best Essay-HLT score, but most per-column optima are three- or four-module combinations that differ by domain, mirroring the greedy analysis, where low-signal domains favor smaller ensembles. (3) On Editorial LGT, a single module (Morphology, 95.6) nearly saturates the task, whereas the adversarial HLT tasks benefit most from the full complement.

\begin{table}[t!]
\centering
\resizebox{1.0\columnwidth}{!}{
\setlength{\tabcolsep}{5pt}
\begin{tabular}{l|cc|cc|cc|c}
    \toprule
    \multirow{2.5}{*}{\textbf{Sub-Detectors}} & \multicolumn{2}{c|}{\textbf{Essay}} & \multicolumn{2}{c|}{\textbf{Paper Abs.}} & \multicolumn{2}{c|}{\textbf{Editorial}} & \multirow{2.5}{*}{\textbf{Avg.}} \\
    \cmidrule(lr){2-3} \cmidrule(lr){4-5} \cmidrule(lr){6-7}
    & HLT & LGT & HLT & LGT & HLT & LGT & \\
    \midrule
    \multicolumn{8}{l}{\textit{1-Sub-Detector}} \\
    \midrule
    Phon & 68.0 & 68.8 & 58.0 & 72.9 & 55.9 & 39.6 & 60.5 \\
    Morph & 79.9 & 79.2 & 65.8 & 76.5 & 69.8 & 95.6 & 77.8 \\
    Syn & 65.5 & 76.5 & 50.8 & 43.4 & 50.9 & 79.6 & 61.1 \\
    Sem & 64.0 & 72.2 & 49.5 & 59.1 & 57.2 & 91.9 & 65.6 \\
    Prag & 72.7 & 51.1 & 52.3 & 67.6 & 56.9 & 76.7 & 62.9 \\
    \midrule
    \multicolumn{8}{l}{\textit{2-Sub-Detectors}} \\
    \midrule
    Phon+Morph & 80.5 & 80.2 & 66.5 & 80.7 & 69.9 & 95.6 & 78.9 \\
    Phon+Syn & 72.4 & 84.1 & 58.0 & 54.3 & 52.1 & 74.1 & 65.8 \\
    Phon+Sem & 71.3 & 75.9 & 54.9 & 68.3 & 58.2 & 84.8 & 68.9 \\
    Phon+Prag & 75.3 & 66.2 & 58.4 & 80.5 & 57.9 & 68.9 & 67.9 \\
    Morph+Syn & 81.1 & 90.4 & 65.8 & 67.7 & 69.7 & 98.7 & 78.9 \\
    Morph+Sem & 81.0 & 84.6 & 66.1 & 74.1 & 70.1 & 96.9 & 78.8 \\
    Morph+Prag & 83.6 & 78.5 & 65.9 & 78.2 & 70.3 & 97.8 & 79.1 \\
    Syn+Sem & 70.3 & 83.8 & 49.5 & 51.2 & 54.2 & 94.5 & 67.2 \\
    Syn+Prag & 74.8 & 75.1 & 52.8 & 54.7 & 54.9 & 85.4 & 66.3 \\
    Sem+Prag & 75.0 & 68.9 & 49.7 & 63.7 & 58.2 & 95.2 & 68.5 \\
    \midrule
    \multicolumn{8}{l}{\textit{3-Sub-Detectors}} \\
    \midrule
    Phon+Morph+Syn & 81.7 & 90.9 & 66.5 & 71.3 & 69.7 & 98.6 & 79.8 \\
    Phon+Morph+Sem & 81.6 & 84.5 & 66.6 & 76.0 & 70.2 & 97.0 & 79.3 \\
    Phon+Morph+Prag & 83.8 & 79.3 & 66.6 & \textbf{82.5} & 70.3 & 97.7 & 80.0 \\
    Phon+Syn+Sem & 74.6 & 86.8 & 55.0 & 59.6 & 55.0 & 91.7 & 70.5 \\
    Phon+Syn+Prag & 77.4 & 82.3 & 58.3 & 66.0 & 55.7 & 81.4 & 70.2 \\
    Phon+Sem+Prag & 77.4 & 73.5 & 55.5 & 74.2 & 59.1 & 92.4 & 72.0 \\
    Morph+Syn+Sem & 82.1 & 92.2 & 66.0 & 65.4 & 69.9 & 99.3 & 79.2 \\
    Morph+Syn+Prag & 84.2 & 89.1 & 65.9 & 71.9 & 70.0 & 99.2 & 80.1 \\
    Morph+Sem+Prag & 84.4 & 82.9 & 66.2 & 75.3 & 70.4 & 98.8 & 79.7 \\
    Syn+Sem+Prag & 76.9 & 81.6 & 49.6 & 57.4 & 56.7 & 97.0 & 69.9 \\
    \midrule
    \multicolumn{8}{l}{\textit{4-Sub-Detectors}} \\
    \midrule
    Phon+Morph+Syn+Sem & 82.5 & \textbf{92.3} & 66.6 & 69.8 & 69.9 & 99.1 & 80.0 \\
    Phon+Morph+Syn+Prag & 84.4 & 89.7 & 66.5 & 76.0 & 70.0 & 99.0 & 80.9 \\
    Phon+Morph+Sem+Prag & 84.5 & 83.1 & 66.9 & 79.1 & \textbf{70.4} & 98.7 & 80.5 \\
    Phon+Syn+Sem+Prag & 79.0 & 85.4 & 55.5 & 65.6 & 57.2 & 95.6 & 73.0 \\
    Morph+Syn+Sem+Prag & 84.9 & 90.9 & 66.2 & 69.3 & 70.1 & \textbf{99.7} & 80.2 \\
    \midrule
    \multicolumn{8}{l}{\textit{5-Sub-Detectors (Full Model)}} \\
    \midrule
    \cellcolor[HTML]{EAECFF}Phon+Morph+Syn+Sem+Prag & \cellcolor[HTML]{EAECFF}\textbf{85.1} & \cellcolor[HTML]{EAECFF}91.2 & \cellcolor[HTML]{EAECFF}\textbf{66.9} & \cellcolor[HTML]{EAECFF}74.0 & \cellcolor[HTML]{EAECFF}70.1 & \cellcolor[HTML]{EAECFF}99.6 & \cellcolor[HTML]{EAECFF}\textbf{81.1} \\
    \bottomrule
\end{tabular}}
\caption{AUROC (\%) for all 31 sub-detector combinations on \textsc{KoLing}. The per-column best is \textbf{bolded}. Sub-Detectors: Phon = Phonology, Morph = Morphology, Syn = Syntax, Sem = Semantics, Prag = Pragmatics.}
\label{tab:ablation_koling}
\end{table}

\clearpage
\clearpage
\section{Results per Generator}
\label{app:koling_per_llm}
Tables~\ref{tab:app_pergen_essay}--\ref{tab:app_pergen_editorial} break down the main binary results (Table~\ref{tab:exp_binary}) by generation model for each domain. Claude-Sonnet-4.6 is the in-distribution generator for training-dependent methods; the remaining four are out-of-distribution. \textsc{KoLingProbe}'s advantage holds across generators rather than being driven by the in-distribution split. All numbers in this and the following appendices are re-aggregations of the same runs as the main tables; no new training is involved.

\begin{table*}[t]
\centering
\small
\resizebox{\textwidth}{!}{
\setlength{\tabcolsep}{3pt}
\begin{tabular}{l|ccccc|ccccc|c}
    \toprule
    \multirow{2}{*}{\textbf{Method}} & \multicolumn{5}{c|}{\textbf{HLT vs.\ HWT}} & \multicolumn{5}{c|}{\textbf{LGT vs.\ HWT}} & \multirow{2}{*}{\textbf{Avg.}} \\
    \cmidrule(lr){2-6} \cmidrule(lr){7-11}
    & Claude & GPT-4o & Qwen & Mistral & Solar & Claude & GPT-4o & Qwen & Mistral & Solar & \\
    \midrule
    \rowcolor{gray!15} \multicolumn{12}{l}{\textit{Confidence-based}} \\
    ~~Log-Likelihood & 54.8/50.8/0.9 & 51.0/51.2/0.4 & 50.8/51.7/0.7 & 50.1/50.7/0.5 & 63.2/57.6/2.9 & 44.6/48.2/0.0 & 44.3/49.1/0.0 & 47.4/50.8/0.7 & 38.2/44.0/0.0 & 64.8/64.1/4.1 & 50.9/51.8/1.0 \\
    ~~Rank & 52.8/50.5/0.3 & 51.8/51.0/0.4 & 52.6/51.7/0.7 & 52.5/50.7/0.7 & 65.0/58.7/1.4 & 37.8/44.0/0.0 & 47.2/50.0/0.9 & 50.0/50.9/2.9 & 43.6/46.3/0.0 & 64.1/64.5/4.5 & 51.7/51.8/1.2 \\
    ~~Log-Rank & 55.2/51.7/0.7 & 51.2/51.7/0.8 & 51.1/52.7/0.5 & 50.6/51.0/0.5 & 62.6/57.0/2.2 & 40.2/45.2/0.0 & 44.4/48.5/0.9 & 49.9/50.9/0.4 & 41.1/44.7/0.0 & 64.2/63.7/6.4 & 51.1/51.7/1.2 \\
    ~~Entropy & 53.5/51.6/1.7 & 50.9/51.4/1.0 & 51.2/50.2/1.8 & 50.7/49.2/0.5 & 56.2/54.1/1.8 & 33.6/38.8/0.8 & 46.3/50.1/0.9 & 59.4/58.7/4.7 & 51.5/55.4/1.7 & 56.4/58.7/5.0 & 51.0/51.8/2.0 \\
    ~~LRR & 54.9/52.7/4.2 & 52.7/54.0/2.8 & 52.3/53.7/3.8 & 51.8/52.8/2.8 & 58.3/54.5/6.5 & 32.1/38.4/1.5 & 44.4/48.6/2.7 & 53.5/56.1/5.4 & 47.1/52.3/3.4 & 57.6/60.2/10.9 & 50.5/52.3/4.4 \\
    \rowcolor{gray!15} \multicolumn{12}{l}{\textit{Perturbation}} \\
    ~~NPR & 76.1/68.9/2.5 & 76.5/69.0/4.4 & 67.6/62.5/6.7 & 69.5/62.4/2.5 & 69.9/63.4/2.4 & 57.8/54.9/1.1 & 85.8/80.1/17.3 & 74.2/68.5/19.1 & 76.0/68.7/5.7 & 78.5/72.4/9.1 & 73.2/67.1/7.1 \\
    ~~DetectGPT & 65.0/62.3/2.0 & 63.6/58.8/4.5 & 57.6/55.1/3.8 & 59.4/56.8/3.6 & 65.4/60.2/5.8 & 58.2/57.2/2.3 & 60.5/56.8/4.5 & 57.7/58.0/18.0 & 45.4/48.2/2.8 & 73.5/67.6/12.7 & 60.6/58.1/6.0 \\
    ~~Fast-DetectGPT & 54.1/52.5/2.1 & 47.8/50.5/2.5 & 55.3/53.4/5.9 & 57.4/54.8/4.8 & 47.7/50.5/2.9 & 67.9/62.9/14.7 & 62.8/58.9/5.9 & 60.0/56.0/14.0 & 70.0/67.3/13.1 & 61.9/59.1/8.6 & 58.5/56.6/7.4 \\
    \rowcolor{gray!15} \multicolumn{12}{l}{\textit{LLM-based}} \\
    ~~Binoculars & 71.5/64.7/5.6 & 76.3/68.3/9.8 & 74.3/66.5/\textbf{9.3} & 74.3/67.3/7.6 & 62.9/58.4/3.2 & 70.0/65.2/1.5 & \textbf{95.6}/\textbf{88.8}/\textbf{37.3} & \textbf{93.1}/\textbf{89.5}/\textbf{61.2} & \textbf{98.1}/\textbf{93.7}/\textbf{71.6} & 86.4/79.4/19.5 & 80.2/74.2/22.7 \\
    ~~LLM-Det (0-shot) & 66.6/63.5/2.9 & 59.8/58.7/2.0 & 55.8/55.7/1.0 & 57.9/57.2/1.7 & 59.8/58.8/1.5 & 76.2/70.0/3.8 & 69.3/63.9/4.1 & 62.9/59.1/4.3 & 63.5/59.7/3.4 & 67.8/62.9/4.5 & 64.0/61.0/2.9 \\
    ~~LLM-Det (10-shot) & 49.6/49.3/2.8 & 54.2/52.4/2.1 & 57.8/55.2/2.9 & 51.3/50.8/0.7 & 57.6/54.3/2.5 & 53.8/54.3/2.6 & 54.7/54.0/5.5 & 57.0/56.9/0.4 & 60.1/60.4/3.4 & 57.4/55.1/0.9 & 55.4/54.3/2.4 \\
    \rowcolor{gray!15} \multicolumn{12}{l}{\textit{Grammar-based}} \\
    ~~GECScore & 46.9/49.6/0.1 & 53.1/52.4/1.1 & 52.5/51.6/0.8 & 49.3/50.4/0.9 & 50.7/51.5/0.8 & 52.3/53.7/0.4 & 45.6/47.0/1.4 & 55.3/54.7/2.5 & 41.6/45.3/0.0 & 54.2/54.2/1.8 & 50.2/51.0/1.0 \\
    \rowcolor{gray!15} \multicolumn{12}{l}{\textit{Neural-based}} \\
    ~~XDAC-D & 62.1/58.0/0.0 & 66.8/63.9/0.0 & 65.7/62.0/0.0 & 64.8/60.5/0.0 & 68.2/64.9/0.0 & 44.0/46.3/0.0 & 81.1/80.9/0.0 & 73.5/74.4/0.0 & 73.5/72.6/0.0 & 70.2/72.6/0.0 & 67.0/65.6/0.0 \\
    ~~RepreGuard & 87.2/78.2/19.9 & 80.0/71.1/\textbf{9.9} & 77.6/68.7/8.3 & 79.4/71.2/\textbf{10.4} & 83.8/75.3/20.8 & \textbf{99.9}/\textbf{98.4}/\textbf{94.4} & 89.9/82.3/18.6 & 80.3/75.1/17.6 & 89.4/83.3/27.3 & 87.6/81.7/21.4 & 85.5/78.5/\textbf{24.9} \\
    \rowcolor{gray!15} \multicolumn{12}{l}{\textit{Linguistic-Feature}} \\
    ~~KatFishNet & 65.0/59.1/1.6 & 60.2/57.0/1.0 & 60.2/56.6/1.0 & 58.8/56.0/0.7 & 60.1/56.2/3.3 & 72.1/66.0/9.8 & 75.8/67.3/10.5 & 74.1/66.6/18.0 & 75.7/67.6/9.1 & 70.9/64.5/10.5 & 67.3/61.7/6.5 \\
    \cellcolor[HTML]{EAECFF}\textbf{\textsc{KoLingProbe}} & \cellcolor[HTML]{EAECFF}\textbf{90.4}/\textbf{81.2}/\textbf{20.4} & \cellcolor[HTML]{EAECFF}\textbf{85.4}/\textbf{78.5}/7.9 & \cellcolor[HTML]{EAECFF}\textbf{80.5}/\textbf{74.3}/6.3 & \cellcolor[HTML]{EAECFF}\textbf{82.7}/\textbf{75.4}/6.8 & \cellcolor[HTML]{EAECFF}\textbf{86.2}/\textbf{75.6}/\textbf{22.5} & \cellcolor[HTML]{EAECFF}97.1/92.4/49.9 & \cellcolor[HTML]{EAECFF}89.2/84.2/7.3 & \cellcolor[HTML]{EAECFF}87.9/83.2/13.5 & \cellcolor[HTML]{EAECFF}86.5/82.9/9.9 & \cellcolor[HTML]{EAECFF}\textbf{95.2}/\textbf{89.3}/\textbf{33.5} & \cellcolor[HTML]{EAECFF}\textbf{88.1}/\textbf{81.7}/17.8 \\
    \bottomrule
\end{tabular}}
\caption{Per-generator binary results on the \textbf{Essay} domain: AUROC/F1/TPR@FPR{=}1\% (\%; 5-seed average for seeded methods). ``Claude'' is the in-distribution generator (training-dependent methods are trained on Claude-Sonnet-4.6 outputs); the other four are OOD. The last column averages the ten generator--task cells. Per column and metric, best is \textbf{bolded}.}
\label{tab:app_pergen_essay}
\end{table*}

\begin{table*}[t]
\centering
\small
\resizebox{\textwidth}{!}{
\setlength{\tabcolsep}{3pt}
\begin{tabular}{l|ccccc|ccccc|c}
    \toprule
    \multirow{2}{*}{\textbf{Method}} & \multicolumn{5}{c|}{\textbf{HLT vs.\ HWT}} & \multicolumn{5}{c|}{\textbf{LGT vs.\ HWT}} & \multirow{2}{*}{\textbf{Avg.}} \\
    \cmidrule(lr){2-6} \cmidrule(lr){7-11}
    & Claude & GPT-4o & Qwen & Mistral & Solar & Claude & GPT-4o & Qwen & Mistral & Solar & \\
    \midrule
    \rowcolor{gray!15} \multicolumn{12}{l}{\textit{Confidence-based}} \\
    ~~Log-Likelihood & 58.3/58.1/4.3 & 53.3/58.5/6.5 & 56.5/60.2/4.2 & 53.1/54.0/7.8 & 58.8/60.5/5.6 & 59.4/57.6/18.2 & 62.8/70.0/8.3 & 41.1/49.9/5.6 & 53.3/61.1/0.0 & 58.7/63.2/0.0 & 55.5/59.3/6.0 \\
    ~~Rank & 57.4/57.2/6.5 & 58.1/58.7/6.5 & 57.2/56.7/6.2 & 52.4/54.0/7.8 & 57.7/56.9/5.6 & 55.8/57.6/18.2 & 58.9/59.0/8.3 & 44.8/52.3/11.1 & 55.9/58.1/7.7 & 58.7/65.3/10.0 & 55.7/57.6/8.8 \\
    ~~Log-Rank & 56.4/57.4/4.3 & 54.3/59.9/6.5 & 56.1/57.8/4.2 & 53.1/57.2/7.8 & 58.3/58.8/5.6 & 55.2/60.6/18.2 & 62.2/72.7/8.3 & 45.9/52.0/5.6 & 55.4/65.7/0.0 & 60.7/68.8/0.0 & 55.8/61.1/6.0 \\
    ~~Entropy & 51.0/54.8/2.2 & 47.7/53.0/0.0 & 46.4/49.5/4.2 & 54.2/56.1/2.0 & 47.9/53.3/0.0 & 58.8/57.6/9.1 & 56.1/58.3/0.0 & 44.4/53.0/0.0 & 35.4/44.7/0.0 & 53.3/59.9/10.0 & 49.5/54.0/2.8 \\
    ~~LRR & 50.5/57.4/8.7 & 53.6/60.1/\textbf{17.4} & 55.8/56.2/\textbf{16.7} & 47.6/53.0/5.9 & 53.5/56.7/11.1 & 44.2/53.6/0.0 & 55.0/57.1/16.7 & 63.0/60.7/16.7 & 66.2/65.7/\textbf{23.1} & 45.3/56.0/0.0 & 53.5/57.6/11.6 \\
    \rowcolor{gray!15} \multicolumn{12}{l}{\textit{Perturbation}} \\
    ~~NPR & 56.0/58.4/6.5 & 58.7/58.6/13.0 & 61.8/65.3/\textbf{16.7} & 59.6/56.9/\textbf{11.8} & 53.4/\textbf{64.4}/5.6 & 34.5/46.2/0.0 & 42.8/50.8/8.3 & 62.6/68.3/5.6 & 47.2/58.1/0.0 & 28.0/43.9/0.0 & 50.5/57.1/6.8 \\
    ~~DetectGPT & 47.3/49.7/2.2 & 55.0/56.0/15.2 & 59.2/59.0/0.0 & 52.5/54.0/7.8 & 49.6/54.6/3.7 & 43.0/52.7/0.0 & 41.1/51.6/0.0 & \textbf{77.8}/\textbf{78.7}/5.6 & 49.2/53.5/7.7 & 37.3/53.3/0.0 & 51.2/56.3/4.2 \\
    ~~Fast-DetectGPT & 45.4/48.3/0.0 & 40.3/48.5/0.0 & 46.8/53.0/0.0 & 50.9/53.6/9.8 & 43.5/46.6/3.7 & 67.9/65.3/9.1 & 75.6/72.7/0.0 & 52.2/57.2/0.0 & \textbf{73.8}/\textbf{78.5}/7.7 & 54.7/61.8/0.0 & 55.1/58.5/3.0 \\
    \rowcolor{gray!15} \multicolumn{12}{l}{\textit{LLM-based}} \\
    ~~Binoculars & 50.8/57.6/4.3 & 53.6/58.8/8.7 & 57.4/63.9/\textbf{16.7} & 55.1/58.3/\textbf{11.8} & 50.2/56.6/3.7 & 37.0/45.8/0.0 & 59.4/62.9/16.7 & 59.6/60.7/\textbf{33.3} & 44.6/56.9/0.0 & 44.0/59.4/0.0 & 51.2/58.1/9.5 \\
    ~~LLM-Det (0-shot) & 53.3/52.3/0.0 & 52.2/53.6/0.0 & 49.1/51.9/0.0 & 51.0/49.9/2.0 & 48.2/49.1/0.0 & 72.1/69.0/0.0 & 47.8/55.0/0.0 & 43.0/47.7/0.0 & 66.2/64.1/0.0 & 56.0/56.6/10.0 & 53.9/54.9/1.2 \\
    ~~LLM-Det (10-shot) & 49.0/51.9/0.0 & 48.8/50.5/2.2 & 55.1/56.6/4.2 & 53.4/54.0/3.9 & 32.1/44.2/1.9 & 58.8/61.5/9.1 & 40.8/55.6/0.0 & 47.4/56.6/0.0 & 59.7/60.7/7.7 & 53.7/56.6/10.0 & 49.9/54.8/3.9 \\
    \rowcolor{gray!15} \multicolumn{12}{l}{\textit{Grammar-based}} \\
    ~~GECScore & 51.9/55.2/10.9 & 53.6/58.8/6.5 & 56.7/57.9/\textbf{16.7} & 49.3/58.3/3.9 & 50.8/52.3/9.3 & 45.5/49.9/0.0 & 34.4/44.1/0.0 & 63.0/60.5/22.2 & 37.9/45.8/7.7 & 22.0/47.6/10.0 & 46.5/53.0/8.7 \\
    \rowcolor{gray!15} \multicolumn{12}{l}{\textit{Neural-based}} \\
    ~~XDAC-D & 65.8/66.5/0.0 & 60.4/61.0/0.0 & \textbf{67.8}/\textbf{66.8}/0.0 & 62.0/61.0/0.0 & 60.2/59.3/0.0 & 59.4/57.1/0.0 & 62.8/59.0/0.0 & 52.0/53.9/0.0 & 59.0/60.3/0.0 & 60.7/59.9/0.0 & 61.0/60.5/0.0 \\
    ~~RepreGuard & 74.8/70.0/23.9 & \textbf{70.6}/\textbf{72.2}/15.2 & 61.4/59.6/12.5 & \textbf{65.5}/\textbf{68.5}/5.9 & 61.3/58.3/16.7 & 77.6/72.0/36.4 & \textbf{77.2}/\textbf{80.5}/25.0 & 50.0/54.4/11.1 & 55.9/64.1/7.7 & 86.7/82.6/60.0 & 68.1/68.2/21.4 \\
    \rowcolor{gray!15} \multicolumn{12}{l}{\textit{Linguistic-Feature}} \\
    ~~KatFishNet & 50.0/42.2/0.0 & 50.0/42.2/0.0 & 50.0/42.5/0.0 & 50.0/42.9/0.0 & 50.0/43.2/0.0 & 54.5/60.6/0.0 & 54.4/62.9/0.0 & 41.9/53.0/0.0 & 49.7/60.7/0.0 & 64.7/67.9/0.0 & 51.5/51.8/0.0 \\
    \cellcolor[HTML]{EAECFF}\textbf{\textsc{KoLingProbe}} & \cellcolor[HTML]{EAECFF}\textbf{79.5}/\textbf{73.9}/\textbf{33.0} & \cellcolor[HTML]{EAECFF}63.8/66.1/10.0 & \cellcolor[HTML]{EAECFF}59.3/57.6/14.2 & \cellcolor[HTML]{EAECFF}64.9/63.1/11.0 & \cellcolor[HTML]{EAECFF}\textbf{66.9}/62.8/\textbf{18.9} & \cellcolor[HTML]{EAECFF}\textbf{89.9}/\textbf{87.6}/\textbf{72.7} & \cellcolor[HTML]{EAECFF}64.1/64.9/\textbf{33.3} & \cellcolor[HTML]{EAECFF}63.0/62.0/31.1 & \cellcolor[HTML]{EAECFF}64.2/64.3/\textbf{23.1} & \cellcolor[HTML]{EAECFF}\textbf{88.7}/\textbf{90.5}/\textbf{78.0} & \cellcolor[HTML]{EAECFF}\textbf{70.4}/\textbf{69.3}/\textbf{32.5} \\
    \bottomrule
\end{tabular}}
\caption{Per-generator binary results on the \textbf{Paper Abs.} domain; format and protocol as in Table~\ref{tab:app_pergen_essay}.}
\label{tab:app_pergen_academic}
\end{table*}

\begin{table*}[t]
\centering
\small
\resizebox{\textwidth}{!}{
\setlength{\tabcolsep}{3pt}
\begin{tabular}{l|ccccc|ccccc|c}
    \toprule
    \multirow{2}{*}{\textbf{Method}} & \multicolumn{5}{c|}{\textbf{HLT vs.\ HWT}} & \multicolumn{5}{c|}{\textbf{LGT vs.\ HWT}} & \multirow{2}{*}{\textbf{Avg.}} \\
    \cmidrule(lr){2-6} \cmidrule(lr){7-11}
    & Claude & GPT-4o & Qwen & Mistral & Solar & Claude & GPT-4o & Qwen & Mistral & Solar & \\
    \midrule
    \rowcolor{gray!15} \multicolumn{12}{l}{\textit{Confidence-based}} \\
    ~~Log-Likelihood & 43.6/49.9/1.8 & 58.4/56.5/1.3 & 56.5/56.2/3.2 & 52.4/52.6/2.6 & 49.8/52.2/1.2 & 57.6/59.8/13.5 & 46.9/56.9/0.0 & 67.4/68.5/4.5 & 46.2/56.0/0.0 & 42.4/54.2/4.8 & 52.1/56.3/3.3 \\
    ~~Rank & 43.0/50.2/2.4 & 58.6/55.4/1.9 & 57.3/56.2/1.3 & 51.8/53.6/2.6 & 49.9/52.5/3.8 & 52.7/56.6/5.4 & 44.2/52.4/0.0 & 63.3/63.3/6.8 & 47.5/56.7/0.0 & 38.3/54.3/0.0 & 50.7/55.1/2.4 \\
    ~~Log-Rank & 44.0/50.2/3.0 & 58.6/56.5/0.6 & 56.4/56.2/1.9 & 52.0/52.8/2.6 & 50.0/52.5/0.6 & 55.2/59.8/5.4 & 45.9/56.9/0.0 & 66.5/67.7/4.5 & 45.6/56.0/0.0 & 39.4/50.0/4.8 & 51.4/55.9/2.3 \\
    ~~Entropy & 42.8/46.7/2.4 & 58.9/57.6/1.9 & 56.1/56.5/1.9 & 50.8/51.6/2.6 & 50.2/50.5/0.6 & 45.2/53.8/0.0 & 38.9/48.3/0.0 & 62.3/67.6/4.5 & 44.0/53.1/0.0 & 34.4/46.5/0.0 & 48.4/53.2/1.4 \\
    ~~LRR & 43.4/47.5/2.4 & 59.3/56.5/2.5 & 56.0/54.3/2.5 & 51.5/52.0/\textbf{3.3} & 50.3/52.2/1.9 & 47.3/53.3/2.7 & 37.6/46.0/2.8 & 61.1/61.9/6.8 & 41.3/47.8/5.9 & 31.4/48.5/4.8 & 47.9/52.0/3.6 \\
    \rowcolor{gray!15} \multicolumn{12}{l}{\textit{Perturbation}} \\
    ~~NPR & 47.2/49.4/1.2 & 47.4/52.4/0.6 & 48.3/50.7/2.5 & 50.9/53.0/0.7 & 49.7/51.4/\textbf{6.2} & 87.4/81.8/37.8 & 74.9/69.7/16.7 & 81.8/76.9/47.7 & 77.4/71.6/20.6 & 88.1/82.3/19.0 & 65.3/63.9/15.3 \\
    ~~DetectGPT & 51.9/52.4/1.2 & 51.9/56.1/0.6 & 52.4/52.5/0.6 & 55.4/54.9/1.3 & 51.8/51.8/0.6 & 66.3/68.6/5.4 & 90.0/86.1/44.4 & 84.7/83.7/45.5 & 88.0/81.2/50.0 & 84.0/78.5/19.0 & 67.6/66.6/16.9 \\
    ~~Fast-DetectGPT & 56.9/54.9/1.2 & 54.2/55.0/0.0 & 57.1/57.4/1.3 & 61.2/56.3/\textbf{3.3} & 51.2/53.7/0.0 & 92.8/89.3/29.7 & 91.5/87.8/30.6 & 86.8/83.5/34.1 & 95.8/92.1/55.9 & 88.9/85.9/38.1 & 73.6/71.6/19.4 \\
    \rowcolor{gray!15} \multicolumn{12}{l}{\textit{LLM-based}} \\
    ~~Binoculars & 55.3/53.2/0.0 & 58.7/55.8/1.3 & 57.3/54.4/1.3 & 50.1/52.0/0.0 & 49.5/50.9/0.0 & 47.8/52.2/0.0 & 94.4/93.8/27.8 & 89.5/88.9/72.7 & 98.4/93.7/79.4 & 73.0/74.8/0.0 & 67.4/67.0/18.2 \\
    ~~LLM-Det (0-shot) & 52.8/51.5/3.6 & 59.0/57.2/\textbf{3.8} & 59.3/55.4/\textbf{7.0} & 57.0/55.0/\textbf{3.3} & 54.5/52.7/2.5 & 50.3/54.9/2.7 & 65.3/65.5/8.3 & 79.8/72.5/25.0 & 80.6/71.8/38.2 & 62.9/66.7/4.8 & 62.1/60.3/9.9 \\
    ~~LLM-Det (10-shot) & 47.5/48.8/3.0 & 53.1/50.5/1.3 & 54.7/53.6/0.0 & 45.0/50.7/0.0 & 46.2/48.4/0.0 & 48.3/50.7/2.7 & 40.6/45.5/8.3 & 57.2/56.6/4.5 & 43.4/48.7/5.9 & 40.5/52.1/4.8 & 47.6/50.6/3.1 \\
    \rowcolor{gray!15} \multicolumn{12}{l}{\textit{Grammar-based}} \\
    ~~GECScore & 53.2/54.1/1.2 & 57.1/54.7/1.3 & 56.1/55.7/1.3 & 52.9/53.0/1.3 & 53.3/52.2/0.6 & 57.1/58.2/0.0 & 49.7/57.0/5.6 & 68.0/64.8/2.3 & 57.5/57.8/2.9 & 51.9/54.9/0.0 & 55.7/56.2/1.6 \\
    \rowcolor{gray!15} \multicolumn{12}{l}{\textit{Neural-based}} \\
    ~~XDAC-D & 60.2/58.2/0.0 & 64.2/62.4/0.0 & 61.3/58.4/0.0 & 59.2/58.8/0.0 & 52.9/54.9/0.0 & 53.9/57.3/0.0 & 73.2/72.8/0.0 & 72.2/73.2/0.0 & 77.9/77.6/0.0 & 61.4/64.7/0.0 & 63.6/63.8/0.0 \\
    ~~RepreGuard & 70.7/65.4/\textbf{5.4} & 70.6/65.7/3.2 & 69.1/66.4/2.5 & 67.7/66.7/2.6 & 61.7/59.3/2.5 & \textbf{100.0}/\textbf{100.0}/\textbf{100.0} & 90.6/81.8/52.8 & 94.1/87.3/61.4 & 94.5/88.9/47.1 & 97.3/92.1/66.7 & 81.6/77.4/34.4 \\
    \rowcolor{gray!15} \multicolumn{12}{l}{\textit{Linguistic-Feature}} \\
    ~~KatFishNet & 58.7/56.0/3.6 & 49.4/51.1/1.3 & 51.1/50.1/1.9 & 50.6/49.6/2.0 & 54.7/53.3/1.2 & 86.0/79.0/27.0 & 93.4/87.8/50.0 & 88.8/83.5/54.5 & 95.7/90.5/44.1 & 95.6/90.1/28.6 & 72.4/69.1/21.4 \\
    \cellcolor[HTML]{EAECFF}\textbf{\textsc{KoLingProbe}} & \cellcolor[HTML]{EAECFF}\textbf{75.8}/\textbf{77.0}/0.0 & \cellcolor[HTML]{EAECFF}\textbf{71.2}/\textbf{76.3}/0.0 & \cellcolor[HTML]{EAECFF}\textbf{71.0}/\textbf{73.6}/0.0 & \cellcolor[HTML]{EAECFF}\textbf{68.7}/\textbf{73.4}/0.0 & \cellcolor[HTML]{EAECFF}\textbf{63.9}/\textbf{64.9}/0.0 & \cellcolor[HTML]{EAECFF}\textbf{100.0}/\textbf{100.0}/\textbf{100.0} & \cellcolor[HTML]{EAECFF}\textbf{99.1}/\textbf{98.5}/\textbf{97.2} & \cellcolor[HTML]{EAECFF}\textbf{98.7}/\textbf{97.2}/\textbf{95.5} & \cellcolor[HTML]{EAECFF}\textbf{100.0}/\textbf{100.0}/\textbf{100.0} & \cellcolor[HTML]{EAECFF}\textbf{100.0}/\textbf{100.0}/\textbf{100.0} & \cellcolor[HTML]{EAECFF}\textbf{84.8}/\textbf{86.1}/\textbf{49.3} \\
    \bottomrule
\end{tabular}}
\caption{Per-generator binary results on the \textbf{Editorial} domain; format and protocol as in Table~\ref{tab:app_pergen_essay}.}
\label{tab:app_pergen_editorial}
\end{table*}

\section{Per-Seed Variance}
\label{app:variance}
Table~\ref{tab:app_std} reports the main binary table with each cell as mean$\pm$std over the five generator test splits (and five random seeds, 42--46, for seeded methods), quantifying the variability behind the averaged numbers in Table~\ref{tab:exp_binary}. The dominant source of dispersion is generator-to-generator variation rather than seed noise.

\begin{table*}[t]
\centering
\resizebox{\textwidth}{!}{
\setlength{\tabcolsep}{2pt}
\begin{tabular}{l|ccc|ccc|ccc|ccc|ccc|ccc|ccc}
    \toprule
    \multirow{3}{*}{\textbf{Method}} & \multicolumn{9}{c|}{\textbf{HLT vs.\ HWT}} & \multicolumn{9}{c|}{\textbf{LGT vs.\ HWT}} & \multicolumn{3}{c}{\multirow{2}{*}{\textbf{Avg.}}} \\
    \cmidrule(lr){2-10} \cmidrule(lr){11-19}
    & \multicolumn{3}{c|}{Essay} & \multicolumn{3}{c|}{Paper Abs.} & \multicolumn{3}{c|}{Editorial} & \multicolumn{3}{c|}{Essay} & \multicolumn{3}{c|}{Paper Abs.} & \multicolumn{3}{c|}{Editorial} & \multicolumn{3}{c}{} \\
    \cmidrule(lr){2-4} \cmidrule(lr){5-7} \cmidrule(lr){8-10} \cmidrule(lr){11-13} \cmidrule(lr){14-16} \cmidrule(lr){17-19} \cmidrule(lr){20-22}
    & AUC & F1 & T@1 & AUC & F1 & T@1 & AUC & F1 & T@1 & AUC & F1 & T@1 & AUC & F1 & T@1 & AUC & F1 & T@1 & AUC & F1 & T@1 \\
    \midrule
    \rowcolor{gray!15} \multicolumn{22}{l}{\textit{Confidence-based}} \\
    ~~Log-Likelihood & 54.0\,{\scriptsize$\pm$4.9} & 52.4\,{\scriptsize$\pm$2.6} & 1.1\,{\scriptsize$\pm$0.9} & 56.0\,{\scriptsize$\pm$2.4} & 58.2\,{\scriptsize$\pm$2.3} & 5.7\,{\scriptsize$\pm$1.4} & 52.1\,{\scriptsize$\pm$5.2} & 53.5\,{\scriptsize$\pm$2.5} & 2.0\,{\scriptsize$\pm$0.8} & 47.9\,{\scriptsize$\pm$9.0} & 51.2\,{\scriptsize$\pm$6.8} & 1.0\,{\scriptsize$\pm$1.6} & 55.1\,{\scriptsize$\pm$7.6} & 60.4\,{\scriptsize$\pm$6.6} & 6.4\,{\scriptsize$\pm$6.7} & 52.1\,{\scriptsize$\pm$9.2} & 59.1\,{\scriptsize$\pm$5.0} & 4.6\,{\scriptsize$\pm$4.9} & 52.8 & 55.8 & 3.5 \\
    ~~Rank & 54.9\,{\scriptsize$\pm$5.0} & 52.5\,{\scriptsize$\pm$3.1} & 0.7\,{\scriptsize$\pm$0.4} & 56.6\,{\scriptsize$\pm$2.1} & 56.7\,{\scriptsize$\pm$1.5} & 6.5\,{\scriptsize$\pm$0.7} & 52.1\,{\scriptsize$\pm$5.6} & 53.6\,{\scriptsize$\pm$2.1} & 2.4\,{\scriptsize$\pm$0.8} & 48.5\,{\scriptsize$\pm$8.8} & 51.1\,{\scriptsize$\pm$7.1} & 1.7\,{\scriptsize$\pm$1.8} & 54.8\,{\scriptsize$\pm$5.2} & 58.5\,{\scriptsize$\pm$4.1} & 11.1\,{\scriptsize$\pm$3.8} & 49.2\,{\scriptsize$\pm$8.5} & 56.6\,{\scriptsize$\pm$3.7} & 2.4\,{\scriptsize$\pm$3.0} & 52.7 & 54.8 & 4.1 \\
    ~~Log-Rank & 54.1\,{\scriptsize$\pm$4.6} & 52.8\,{\scriptsize$\pm$2.1} & 1.0\,{\scriptsize$\pm$0.6} & 55.6\,{\scriptsize$\pm$1.8} & 58.3\,{\scriptsize$\pm$1.0} & 5.7\,{\scriptsize$\pm$1.4} & 52.2\,{\scriptsize$\pm$5.1} & 53.6\,{\scriptsize$\pm$2.4} & 1.8\,{\scriptsize$\pm$1.0} & 48.0\,{\scriptsize$\pm$8.8} & 50.6\,{\scriptsize$\pm$6.9} & 1.5\,{\scriptsize$\pm$2.4} & 55.9\,{\scriptsize$\pm$5.7} & 64.0\,{\scriptsize$\pm$7.2} & 6.4\,{\scriptsize$\pm$6.7} & 50.5\,{\scriptsize$\pm$9.5} & 58.1\,{\scriptsize$\pm$5.8} & 2.9\,{\scriptsize$\pm$2.4} & 52.7 & 56.2 & 3.2 \\
    ~~Entropy & 52.5\,{\scriptsize$\pm$2.1} & 51.3\,{\scriptsize$\pm$1.6} & 1.4\,{\scriptsize$\pm$0.5} & 49.5\,{\scriptsize$\pm$2.8} & 53.3\,{\scriptsize$\pm$2.2} & 1.7\,{\scriptsize$\pm$1.6} & 51.8\,{\scriptsize$\pm$5.5} & 52.6\,{\scriptsize$\pm$4.0} & 1.9\,{\scriptsize$\pm$0.7} & 49.5\,{\scriptsize$\pm$9.1} & 52.3\,{\scriptsize$\pm$7.5} & 2.6\,{\scriptsize$\pm$1.9} & 49.6\,{\scriptsize$\pm$8.6} & 54.7\,{\scriptsize$\pm$5.5} & 3.8\,{\scriptsize$\pm$4.7} & 45.0\,{\scriptsize$\pm$9.5} & 53.9\,{\scriptsize$\pm$7.4} & 0.9\,{\scriptsize$\pm$1.8} & 49.6 & 53.0 & 2.0 \\
    ~~LRR & 54.0\,{\scriptsize$\pm$2.4} & 53.5\,{\scriptsize$\pm$0.7} & 4.0\,{\scriptsize$\pm$1.4} & 52.2\,{\scriptsize$\pm$2.8} & 56.7\,{\scriptsize$\pm$2.3} & 11.9\,{\scriptsize$\pm$4.5} & 52.1\,{\scriptsize$\pm$5.4} & 52.5\,{\scriptsize$\pm$3.0} & 2.5\,{\scriptsize$\pm$0.4} & 47.0\,{\scriptsize$\pm$8.8} & 51.1\,{\scriptsize$\pm$7.4} & 4.8\,{\scriptsize$\pm$3.3} & 54.7\,{\scriptsize$\pm$8.9} & 58.6\,{\scriptsize$\pm$4.2} & 11.3\,{\scriptsize$\pm$9.5} & 43.7\,{\scriptsize$\pm$10.1} & 51.5\,{\scriptsize$\pm$5.7} & 4.6\,{\scriptsize$\pm$1.6} & 50.6 & 54.0 & 6.5 \\
    \rowcolor{gray!15} \multicolumn{22}{l}{\textit{Perturbation}} \\
    ~~NPR & 71.9\,{\scriptsize$\pm$3.6} & 65.3\,{\scriptsize$\pm$3.1} & 3.7\,{\scriptsize$\pm$1.7} & 57.9\,{\scriptsize$\pm$2.9} & 60.7\,{\scriptsize$\pm$3.5} & 10.7\,{\scriptsize$\pm$4.2} & 48.7\,{\scriptsize$\pm$1.4} & 51.4\,{\scriptsize$\pm$1.3} & 2.3\,{\scriptsize$\pm$2.1} & 74.5\,{\scriptsize$\pm$9.2} & 68.9\,{\scriptsize$\pm$8.2} & 10.4\,{\scriptsize$\pm$6.8} & 43.0\,{\scriptsize$\pm$11.8} & 53.4\,{\scriptsize$\pm$8.9} & 2.8\,{\scriptsize$\pm$3.5} & 81.9\,{\scriptsize$\pm$5.3} & 76.5\,{\scriptsize$\pm$5.1} & 28.4\,{\scriptsize$\pm$12.2} & 63.0 & 62.7 & 9.7 \\
    ~~DetectGPT & 62.2\,{\scriptsize$\pm$3.1} & 58.6\,{\scriptsize$\pm$2.5} & 4.0\,{\scriptsize$\pm$1.3} & 52.7\,{\scriptsize$\pm$4.2} & 54.7\,{\scriptsize$\pm$3.0} & 5.8\,{\scriptsize$\pm$5.4} & 52.7\,{\scriptsize$\pm$1.4} & 53.5\,{\scriptsize$\pm$1.7} & 0.9\,{\scriptsize$\pm$0.3} & 59.1\,{\scriptsize$\pm$8.9} & 57.6\,{\scriptsize$\pm$6.2} & 8.1\,{\scriptsize$\pm$6.2} & 49.7\,{\scriptsize$\pm$14.6} & 58.0\,{\scriptsize$\pm$10.4} & 2.6\,{\scriptsize$\pm$3.3} & 82.6\,{\scriptsize$\pm$8.4} & 79.6\,{\scriptsize$\pm$6.0} & 32.9\,{\scriptsize$\pm$17.5} & 59.8 & 60.3 & 9.0 \\
    ~~Fast-DetectGPT & 52.5\,{\scriptsize$\pm$4.0} & 52.3\,{\scriptsize$\pm$1.7} & 3.7\,{\scriptsize$\pm$1.4} & 45.4\,{\scriptsize$\pm$3.5} & 50.0\,{\scriptsize$\pm$2.8} & 2.7\,{\scriptsize$\pm$3.8} & 56.1\,{\scriptsize$\pm$3.3} & 55.4\,{\scriptsize$\pm$1.3} & 1.1\,{\scriptsize$\pm$1.2} & 64.5\,{\scriptsize$\pm$3.8} & 60.8\,{\scriptsize$\pm$3.9} & 11.3\,{\scriptsize$\pm$3.4} & 64.8\,{\scriptsize$\pm$9.7} & 67.1\,{\scriptsize$\pm$7.6} & 3.4\,{\scriptsize$\pm$4.1} & 91.2\,{\scriptsize$\pm$3.1} & 87.7\,{\scriptsize$\pm$2.9} & 37.7\,{\scriptsize$\pm$9.6} & 62.4 & 62.2 & 10.0 \\
    \rowcolor{gray!15} \multicolumn{22}{l}{\textit{LLM-based}} \\
    ~~Binoculars & 71.9\,{\scriptsize$\pm$4.7} & 65.0\,{\scriptsize$\pm$3.5} & 7.1\,{\scriptsize$\pm$2.4} & 53.4\,{\scriptsize$\pm$2.7} & 59.0\,{\scriptsize$\pm$2.5} & 9.0\,{\scriptsize$\pm$4.8} & 54.2\,{\scriptsize$\pm$3.8} & 53.3\,{\scriptsize$\pm$1.7} & 0.5\,{\scriptsize$\pm$0.6} & 88.6\,{\scriptsize$\pm$10.1} & 83.3\,{\scriptsize$\pm$10.2} & 38.2\,{\scriptsize$\pm$25.8} & 48.9\,{\scriptsize$\pm$9.1} & 57.2\,{\scriptsize$\pm$6.0} & 10.0\,{\scriptsize$\pm$13.3} & 80.6\,{\scriptsize$\pm$18.5} & 80.7\,{\scriptsize$\pm$15.9} & 36.0\,{\scriptsize$\pm$34.3} & 66.3 & 66.4 & 16.8 \\
    ~~LLM-Det (0-shot) & 60.0\,{\scriptsize$\pm$3.6} & 58.8\,{\scriptsize$\pm$2.6} & 1.8\,{\scriptsize$\pm$0.6} & 50.8\,{\scriptsize$\pm$1.9} & 51.4\,{\scriptsize$\pm$1.6} & 0.4\,{\scriptsize$\pm$0.8} & 56.5\,{\scriptsize$\pm$2.5} & 54.3\,{\scriptsize$\pm$2.0} & 4.0\,{\scriptsize$\pm$1.5} & 67.9\,{\scriptsize$\pm$4.8} & 63.1\,{\scriptsize$\pm$3.9} & 4.0\,{\scriptsize$\pm$0.4} & 57.0\,{\scriptsize$\pm$10.9} & 58.5\,{\scriptsize$\pm$7.4} & 2.0\,{\scriptsize$\pm$4.0} & 67.8\,{\scriptsize$\pm$11.4} & 66.3\,{\scriptsize$\pm$6.3} & 15.8\,{\scriptsize$\pm$13.7} & 60.0 & 58.7 & 4.7 \\
    ~~LLM-Det (10-shot) & 54.1\,{\scriptsize$\pm$3.3} & 52.4\,{\scriptsize$\pm$2.2} & 2.2\,{\scriptsize$\pm$0.8} & 47.7\,{\scriptsize$\pm$8.2} & 51.4\,{\scriptsize$\pm$4.2} & 2.4\,{\scriptsize$\pm$1.5} & 49.3\,{\scriptsize$\pm$3.9} & 50.4\,{\scriptsize$\pm$1.9} & 0.9\,{\scriptsize$\pm$1.2} & 56.6\,{\scriptsize$\pm$2.2} & 56.1\,{\scriptsize$\pm$2.4} & 2.6\,{\scriptsize$\pm$1.8} & 52.1\,{\scriptsize$\pm$7.1} & 58.2\,{\scriptsize$\pm$2.4} & 5.4\,{\scriptsize$\pm$4.4} & 46.0\,{\scriptsize$\pm$6.3} & 50.7\,{\scriptsize$\pm$3.7} & 5.2\,{\scriptsize$\pm$1.9} & 51.0 & 53.2 & 3.1 \\
    \rowcolor{gray!15} \multicolumn{22}{l}{\textit{Grammar-based}} \\
    ~~GECScore & 50.5\,{\scriptsize$\pm$2.3} & 51.1\,{\scriptsize$\pm$1.0} & 0.8\,{\scriptsize$\pm$0.3} & 52.5\,{\scriptsize$\pm$2.6} & 56.5\,{\scriptsize$\pm$2.5} & 9.4\,{\scriptsize$\pm$4.3} & 54.5\,{\scriptsize$\pm$1.7} & 53.9\,{\scriptsize$\pm$1.2} & 1.1\,{\scriptsize$\pm$0.3} & 49.8\,{\scriptsize$\pm$5.3} & 51.0\,{\scriptsize$\pm$4.0} & 1.2\,{\scriptsize$\pm$0.9} & 40.6\,{\scriptsize$\pm$13.5} & 49.6\,{\scriptsize$\pm$5.8} & 8.0\,{\scriptsize$\pm$8.2} & 56.8\,{\scriptsize$\pm$6.3} & 58.5\,{\scriptsize$\pm$3.3} & 2.2\,{\scriptsize$\pm$2.1} & 50.8 & 53.4 & 3.8 \\
    \rowcolor{gray!15} \multicolumn{22}{l}{\textit{Neural-based}} \\
    ~~XDAC-D & 65.5\,{\scriptsize$\pm$2.1} & 61.9\,{\scriptsize$\pm$2.5} & 0.0\,{\scriptsize$\pm$0.0} & 63.3\,{\scriptsize$\pm$3.0} & 62.9\,{\scriptsize$\pm$3.1} & 0.0\,{\scriptsize$\pm$0.0} & 59.6\,{\scriptsize$\pm$3.7} & 58.5\,{\scriptsize$\pm$2.4} & 0.0\,{\scriptsize$\pm$0.0} & 68.5\,{\scriptsize$\pm$12.7} & 69.4\,{\scriptsize$\pm$11.9} & 0.0\,{\scriptsize$\pm$0.0} & 58.8\,{\scriptsize$\pm$3.6} & 58.0\,{\scriptsize$\pm$2.4} & 0.0\,{\scriptsize$\pm$0.0} & 67.7\,{\scriptsize$\pm$8.8} & 69.1\,{\scriptsize$\pm$7.2} & 0.0\,{\scriptsize$\pm$0.0} & 63.9 & 63.3 & 0.0 \\
    ~~RepreGuard & 81.6\,{\scriptsize$\pm$3.4} & 72.9\,{\scriptsize$\pm$3.4} & 13.9\,{\scriptsize$\pm$5.4} & 66.7\,{\scriptsize$\pm$5.3} & 65.7\,{\scriptsize$\pm$5.7} & 14.8\,{\scriptsize$\pm$5.9} & 68.0\,{\scriptsize$\pm$3.3} & 64.7\,{\scriptsize$\pm$2.7} & 3.2\,{\scriptsize$\pm$1.1} & 89.4\,{\scriptsize$\pm$6.3} & 84.1\,{\scriptsize$\pm$7.7} & 35.9\,{\scriptsize$\pm$29.4} & 69.5\,{\scriptsize$\pm$14.0} & 70.7\,{\scriptsize$\pm$10.5} & 28.0\,{\scriptsize$\pm$19.0} & 95.3\,{\scriptsize$\pm$3.2} & 90.0\,{\scriptsize$\pm$6.0} & 65.6\,{\scriptsize$\pm$18.5} & 78.4 & 74.7 & 26.9 \\
    \rowcolor{gray!15} \multicolumn{22}{l}{\textit{Linguistic-Feature}} \\
    ~~KatFishNet & 60.9\,{\scriptsize$\pm$2.1} & 57.0\,{\scriptsize$\pm$1.1} & 1.5\,{\scriptsize$\pm$1.0} & 50.0\,{\scriptsize$\pm$0.0} & 42.6\,{\scriptsize$\pm$0.4} & 0.0\,{\scriptsize$\pm$0.0} & 52.9\,{\scriptsize$\pm$3.4} & 52.0\,{\scriptsize$\pm$2.4} & 2.0\,{\scriptsize$\pm$0.9} & 73.7\,{\scriptsize$\pm$1.9} & 66.4\,{\scriptsize$\pm$1.1} & 11.6\,{\scriptsize$\pm$3.3} & 53.1\,{\scriptsize$\pm$7.4} & 61.0\,{\scriptsize$\pm$4.8} & 0.0\,{\scriptsize$\pm$0.0} & 91.9\,{\scriptsize$\pm$3.9} & 86.2\,{\scriptsize$\pm$4.4} & 40.9\,{\scriptsize$\pm$11.2} & 63.7 & 60.9 & 9.3 \\
    \cellcolor[HTML]{EAECFF}\textbf{\textsc{KoLingProbe}} & \cellcolor[HTML]{EAECFF}85.1\,{\scriptsize$\pm$3.4} & \cellcolor[HTML]{EAECFF}77.0\,{\scriptsize$\pm$2.5} & \cellcolor[HTML]{EAECFF}12.8\,{\scriptsize$\pm$7.1} & \cellcolor[HTML]{EAECFF}66.9\,{\scriptsize$\pm$6.8} & \cellcolor[HTML]{EAECFF}64.7\,{\scriptsize$\pm$5.4} & \cellcolor[HTML]{EAECFF}17.4\,{\scriptsize$\pm$8.5} & \cellcolor[HTML]{EAECFF}70.1\,{\scriptsize$\pm$3.9} & \cellcolor[HTML]{EAECFF}73.1\,{\scriptsize$\pm$4.3} & \cellcolor[HTML]{EAECFF}0.0\,{\scriptsize$\pm$0.0} & \cellcolor[HTML]{EAECFF}91.2\,{\scriptsize$\pm$4.2} & \cellcolor[HTML]{EAECFF}86.4\,{\scriptsize$\pm$3.8} & \cellcolor[HTML]{EAECFF}22.8\,{\scriptsize$\pm$16.4} & \cellcolor[HTML]{EAECFF}74.0\,{\scriptsize$\pm$12.6} & \cellcolor[HTML]{EAECFF}73.9\,{\scriptsize$\pm$12.5} & \cellcolor[HTML]{EAECFF}47.6\,{\scriptsize$\pm$23.0} & \cellcolor[HTML]{EAECFF}99.6\,{\scriptsize$\pm$0.5} & \cellcolor[HTML]{EAECFF}99.1\,{\scriptsize$\pm$1.1} & \cellcolor[HTML]{EAECFF}98.5\,{\scriptsize$\pm$1.9} & \cellcolor[HTML]{EAECFF}\textbf{81.1} & \cellcolor[HTML]{EAECFF}\textbf{79.0} & \cellcolor[HTML]{EAECFF}\textbf{33.2} \\
    \bottomrule
\end{tabular}}
\caption{Main binary table with standard deviations: AUROC (AUC), F1, and TPR@FPR{=}1\% (T@1), each reported as mean$\pm$std over the 5 generator test splits (and 5 seeds for seeded methods). Same methods and training protocol as Table~\ref{tab:exp_binary}; the rightmost block averages the six domain--task means. Per column and metric, best is \textbf{bolded}.}
\label{tab:app_std}
\end{table*}

\section{Essay Writing-Type and Modification-Level Breakdowns}
\label{app:essay_breakdown}
Table~\ref{tab:app_essaytype} breaks the Essay-domain results down by writing type, and Table~\ref{tab:app_hltlevel} breaks the HLT-vs-HWT task down by modification level. Detection generally becomes easier as the LLM intervention strengthens (Proofread $\to$ Polish $\to$ Paraphrase), consistent with lighter edits leaving more of the human signal intact, Proofread remains the most adversarial setting in every domain.

\begin{table*}[t]
\centering
\small
\resizebox{\textwidth}{!}{
\setlength{\tabcolsep}{3pt}
\begin{tabular}{l|ccccc|ccccc}
    \toprule
    \multirow{2}{*}{\textbf{Method}} & \multicolumn{5}{c|}{\textbf{HLT vs.\ HWT}} & \multicolumn{5}{c}{\textbf{LGT vs.\ HWT}} \\
    \cmidrule(lr){2-6} \cmidrule(lr){7-11}
    & Creative & Prob.-Solv. & Expository & Argument. & Pro-Con & Creative & Prob.-Solv. & Expository & Argument. & Pro-Con \\
    \midrule
    \rowcolor{gray!15} \multicolumn{11}{l}{\textit{Confidence-based}} \\
    ~~Log-Likelihood & 54.4/54.2/0.3 & 56.8/57.9/3.1 & 52.5/52.4/1.8 & 57.8/56.7/7.8 & 51.6/54.4/0.9 & 38.8/47.8/0.0 & 58.5/62.1/2.4 & 38.8/46.6/0.3 & 55.7/60.1/4.6 & 46.9/53.9/0.3 \\
    ~~Rank & 54.4/54.6/0.6 & 58.1/57.7/5.9 & 53.9/51.7/3.0 & 58.2/58.0/8.2 & 52.2/54.3/6.2 & 38.4/46.1/0.4 & 60.3/62.8/6.1 & 39.3/45.0/1.6 & 56.2/58.5/9.9 & 48.0/53.1/1.6 \\
    ~~Log-Rank & 52.8/53.3/1.4 & 56.1/58.9/1.2 & 52.3/52.0/5.5 & 59.4/57.0/14.7 & 53.7/55.2/1.2 & 36.0/44.3/0.9 & 59.2/61.5/3.0 & 38.3/45.7/3.2 & 57.9/61.7/13.7 & 47.0/52.8/1.2 \\
    ~~Entropy & 47.2/53.0/0.5 & 53.6/56.7/1.7 & 48.0/49.2/1.4 & 59.2/54.6/12.7 & 56.9/56.0/1.3 & 35.1/46.2/0.0 & 62.0/65.4/5.4 & 42.2/50.6/2.4 & 60.1/61.4/17.1 & 47.1/51.8/2.9 \\
    ~~LRR & 49.6/53.1/5.3 & 54.6/59.0/3.1 & 49.7/50.3/6.2 & 60.6/55.8/21.2 & 58.7/59.0/3.9 & 34.1/44.8/0.4 & 56.8/61.6/5.3 & 38.3/47.4/4.3 & 57.7/61.7/16.1 & 46.3/52.1/5.9 \\
    \rowcolor{gray!15} \multicolumn{11}{l}{\textit{Perturbation}} \\
    ~~NPR & 70.6/65.3/5.8 & 71.7/66.3/5.0 & 69.8/64.1/4.0 & 72.7/69.6/1.7 & 75.4/69.7/7.9 & 71.3/69.0/11.3 & 74.4/70.0/6.0 & 74.7/71.6/15.2 & 75.6/72.2/4.4 & 77.7/73.9/23.0 \\
    ~~DetectGPT & 62.0/60.5/2.9 & 60.7/59.7/7.1 & 63.2/60.2/3.0 & 66.5/63.5/2.8 & 60.0/58.0/5.0 & 56.7/58.3/6.4 & 56.1/60.0/3.9 & 61.9/61.5/10.4 & 55.5/57.2/9.4 & 66.2/65.4/8.5 \\
    ~~Fast-DetectGPT & 54.0/54.0/4.4 & 49.9/53.0/2.8 & 52.4/53.4/5.6 & 50.9/53.0/3.9 & 55.1/55.6/4.0 & 74.9/71.2/21.9 & 58.3/58.6/3.9 & 64.7/64.3/14.8 & 61.6/61.8/10.6 & 63.6/64.0/6.9 \\
    \rowcolor{gray!15} \multicolumn{11}{l}{\textit{LLM-based}} \\
    ~~Binoculars & 71.6/65.4/6.0 & 67.7/63.2/20.2 & 69.3/65.3/4.0 & 74.0/70.1/3.9 & 78.8/68.1/31.2 & 89.8/86.2/54.4 & 82.0/78.4/47.1 & 91.9/86.9/\textbf{53.5} & 91.7/87.3/34.7 & 87.9/84.3/15.8 \\
    ~~LLM-Det (0-shot) & 51.3/53.9/3.4 & 59.9/58.8/2.4 & 59.5/59.9/1.1 & 65.0/63.0/6.5 & 69.9/63.9/6.2 & 59.4/58.1/8.4 & 65.4/62.2/3.6 & 65.8/64.9/2.2 & 79.7/73.7/20.7 & 74.2/67.1/13.3 \\
    ~~LLM-Det (10-shot) & 56.0/54.9/4.2 & 55.4/53.6/3.8 & 48.4/49.7/1.9 & 58.1/57.3/1.9 & 53.0/52.4/2.2 & 59.7/59.9/1.3 & 56.5/58.5/4.4 & 52.9/55.4/0.4 & 54.9/57.4/2.6 & 62.1/60.3/2.1 \\
    \rowcolor{gray!15} \multicolumn{11}{l}{\textit{Grammar-based}} \\
    ~~GECScore & 51.3/52.9/0.4 & 52.9/53.0/0.4 & 50.9/52.8/1.5 & 50.0/52.2/1.0 & 48.0/51.5/0.9 & 48.9/54.1/1.1 & 54.6/60.1/0.6 & 46.6/51.0/3.0 & 51.7/53.8/5.5 & 46.5/54.4/0.4 \\
    \rowcolor{gray!15} \multicolumn{11}{l}{\textit{Neural-based}} \\
    ~~XDAC-D & 58.0/56.4/0.0 & 67.7/65.2/0.0 & 55.7/53.6/0.0 & 72.8/69.7/0.0 & 69.4/65.9/0.0 & 59.5/62.3/0.0 & 70.3/68.8/0.0 & 64.8/69.4/0.0 & 72.7/73.2/0.0 & 75.8/76.1/0.0 \\
    ~~RepreGuard & 79.5/73.7/7.3 & 79.1/71.8/\textbf{30.8} & 80.6/72.0/\textbf{21.3} & 83.2/74.5/\textbf{24.3} & 85.9/77.6/\textbf{42.5} & 90.5/86.1/\textbf{57.8} & \textbf{86.6}/\textbf{84.7}/\textbf{53.2} & 89.4/86.0/31.0 & 90.7/87.1/\textbf{36.8} & \textbf{90.4}/84.6/\textbf{63.2} \\
    \rowcolor{gray!15} \multicolumn{11}{l}{\textit{Linguistic-Feature}} \\
    ~~KatFishNet & 56.1/52.3/2.2 & 61.1/59.0/4.5 & 66.2/60.3/0.4 & 59.3/56.9/3.9 & 62.1/58.8/0.4 & 77.4/70.3/14.3 & 72.5/65.7/12.7 & 74.4/64.6/26.2 & 76.2/72.3/2.1 & 66.7/63.2/10.6 \\
    \cellcolor[HTML]{EAECFF}\textbf{\textsc{KoLingProbe}} & \cellcolor[HTML]{EAECFF}\textbf{84.3}/\textbf{78.4}/\textbf{17.0} & \cellcolor[HTML]{EAECFF}\textbf{82.0}/\textbf{74.6}/15.2 & \cellcolor[HTML]{EAECFF}\textbf{83.1}/\textbf{79.5}/9.1 & \cellcolor[HTML]{EAECFF}\textbf{86.6}/\textbf{81.0}/6.1 & \cellcolor[HTML]{EAECFF}\textbf{90.9}/\textbf{80.3}/38.3 & \cellcolor[HTML]{EAECFF}\textbf{93.4}/\textbf{90.8}/14.4 & \cellcolor[HTML]{EAECFF}83.0/79.2/16.2 & \cellcolor[HTML]{EAECFF}\textbf{94.2}/\textbf{89.3}/41.9 & \cellcolor[HTML]{EAECFF}\textbf{91.8}/\textbf{88.3}/6.7 & \cellcolor[HTML]{EAECFF}90.2/\textbf{87.1}/19.0 \\
    \bottomrule
\end{tabular}}
\caption{Binary results on the Essay domain by writing type: AUROC/F1/TPR@FPR{=}1\% (\%; averaged over 5 generators, and 5 seeds for seeded methods). Types: Creative(글짓기), Problem-Solving(대안제시), Expository(설명글), Argumentative(주장), Pro-Con(찬성반대). Per column and metric, best is \textbf{bolded}.}
\label{tab:app_essaytype}
\end{table*}

\begin{table*}[t]
\centering
\small
\resizebox{\textwidth}{!}{
\setlength{\tabcolsep}{3pt}
\begin{tabular}{l|ccc|ccc|ccc}
    \toprule
    \multirow{2}{*}{\textbf{Method}} & \multicolumn{3}{c|}{\textbf{Essay}} & \multicolumn{3}{c|}{\textbf{Paper Abs.}} & \multicolumn{3}{c}{\textbf{Editorial}} \\
    \cmidrule(lr){2-4} \cmidrule(lr){5-7} \cmidrule(lr){8-10}
    & Proofread & Polish & Paraphrase & Proofread & Polish & Paraphrase & Proofread & Polish & Paraphrase \\
    \midrule
    \rowcolor{gray!15} \multicolumn{10}{l}{\textit{Confidence-based}} \\
    ~~Log-Likelihood & 51.4/52.1/0.4 & 52.2/52.9/0.8 & 58.5/58.3/2.1 & 53.3/58.0/9.3 & 55.2/59.2/5.7 & 58.9/61.9/2.2 & 49.4/51.3/0.4 & 50.9/53.1/1.5 & 56.2/58.3/4.3 \\
    ~~Rank & 53.1/52.9/0.5 & 52.7/52.5/0.7 & 59.1/57.8/0.9 & 55.7/57.5/9.3 & 55.9/57.8/3.6 & 57.8/61.5/7.5 & 50.1/52.0/2.1 & 51.0/53.3/1.7 & 55.6/58.0/3.4 \\
    ~~Log-Rank & 52.6/53.0/0.6 & 51.8/52.2/1.1 & 58.0/57.1/1.3 & 53.8/57.7/9.3 & 54.6/58.9/5.7 & 58.0/61.7/2.2 & 49.8/52.1/0.4 & 50.9/53.1/1.1 & 56.2/58.5/3.9 \\
    ~~Entropy & 53.6/54.5/1.9 & 50.1/52.2/1.1 & 53.7/54.5/1.1 & 49.6/53.6/3.9 & 46.5/53.9/1.2 & 53.3/58.3/0.0 & 50.3/52.6/1.9 & 50.0/53.5/1.8 & 55.4/58.3/1.9 \\
    ~~LRR & 54.8/54.8/3.9 & 51.5/52.0/3.8 & 55.7/55.3/4.4 & 51.8/55.5/7.5 & 53.5/57.0/\textbf{15.1} & 50.2/55.1/11.6 & 50.3/52.1/\textbf{2.9} & 50.3/52.3/2.8 & 56.1/58.2/1.9 \\
    \rowcolor{gray!15} \multicolumn{10}{l}{\textit{Perturbation}} \\
    ~~NPR & 67.9/65.0/2.8 & 73.0/68.1/4.2 & 75.3/69.7/4.2 & 58.5/62.8/10.2 & 56.1/59.7/11.4 & 58.9/63.7/10.1 & 47.8/50.8/2.0 & 45.2/48.8/1.8 & 53.4/54.4/3.1 \\
    ~~DetectGPT & 58.3/57.3/2.7 & 61.7/59.7/4.5 & 66.8/63.8/4.9 & 55.7/59.2/5.3 & 51.4/56.9/4.9 & 51.3/55.2/7.4 & 51.0/54.1/0.8 & 49.5/51.8/0.0 & 58.4/59.0/1.9 \\
    ~~Fast-DetectGPT & 50.8/51.4/2.9 & 54.2/52.6/5.1 & 52.5/52.2/3.1 & 43.8/53.9/2.8 & 50.3/55.3/1.9 & 41.0/49.9/3.3 & 53.8/55.7/0.4 & 55.8/57.3/1.8 & 58.9/59.5/1.2 \\
    \rowcolor{gray!15} \multicolumn{10}{l}{\textit{LLM-based}} \\
    ~~Binoculars & 68.5/64.4/5.9 & 73.5/68.7/8.2 & 74.0/69.8/7.5 & 52.5/58.9/8.8 & 55.4/60.4/9.2 & 51.0/59.6/8.5 & 53.1/56.0/0.0 & 53.5/54.9/0.4 & 56.5/57.0/1.2 \\
    ~~LLM-Det (0-shot) & 59.5/59.0/1.4 & 60.9/59.5/1.9 & 59.5/59.0/2.2 & 51.1/54.5/0.0 & 51.5/53.6/1.1 & 50.1/52.1/0.0 & 52.2/54.5/2.1 & 53.8/53.9/\textbf{4.2} & 64.1/62.0/\textbf{5.8} \\
    ~~LLM-Det (10-shot) & 53.7/54.5/2.2 & 52.2/53.4/1.8 & 56.5/55.3/2.4 & 47.1/53.7/2.7 & 48.7/56.8/1.2 & 47.3/52.6/3.6 & 52.0/53.4/1.5 & 47.8/50.0/0.7 & 48.5/51.7/0.3 \\
    \rowcolor{gray!15} \multicolumn{10}{l}{\textit{Grammar-based}} \\
    ~~GECScore & 47.6/49.3/0.8 & 52.3/53.2/1.1 & 51.9/52.8/0.3 & 51.1/55.6/7.7 & 50.6/54.3/6.8 & 55.6/58.1/14.1 & 52.5/54.6/1.5 & 53.5/54.7/0.4 & 57.4/58.4/1.5 \\
    \rowcolor{gray!15} \multicolumn{10}{l}{\textit{Neural-based}} \\
    ~~XDAC-D & 58.9/62.1/0.0 & 66.2/68.5/0.0 & 71.9/73.0/0.0 & 58.4/58.7/0.0 & \textbf{63.9}/64.5/0.0 & 66.8/66.0/0.0 & 54.4/55.5/0.0 & 57.4/58.5/0.0 & 67.4/66.7/0.0 \\
    ~~RepreGuard & 73.7/69.2/\textbf{6.5} & 83.4/77.0/\textbf{14.8} & 88.4/81.5/20.7 & 58.4/62.4/6.4 & 63.7/\textbf{67.4}/13.9 & \textbf{78.5}/\textbf{77.6}/24.3 & 64.4/61.8/2.0 & 66.3/64.2/2.4 & 72.8/69.8/5.3 \\
    \rowcolor{gray!15} \multicolumn{10}{l}{\textit{Linguistic-Feature}} \\
    ~~KatFishNet & 57.6/55.9/0.9 & 61.6/58.7/1.9 & 63.7/59.5/1.8 & 50.0/32.0/0.0 & 50.0/33.8/0.0 & 50.0/32.4/0.0 & 52.4/54.4/0.7 & 52.0/53.5/2.8 & 54.2/55.6/2.3 \\
    \cellcolor[HTML]{EAECFF}\textbf{\textsc{KoLingProbe}} & \cellcolor[HTML]{EAECFF}\textbf{79.3}/\textbf{72.9}/4.5 & \cellcolor[HTML]{EAECFF}\textbf{86.7}/\textbf{79.5}/12.5 & \cellcolor[HTML]{EAECFF}\textbf{89.7}/\textbf{82.9}/\textbf{21.7} & \cellcolor[HTML]{EAECFF}\textbf{66.0}/\textbf{66.2}/\textbf{10.6} & \cellcolor[HTML]{EAECFF}62.3/64.2/10.8 & \cellcolor[HTML]{EAECFF}72.8/72.0/\textbf{32.1} & \cellcolor[HTML]{EAECFF}\textbf{68.4}/\textbf{68.5}/0.0 & \cellcolor[HTML]{EAECFF}\textbf{68.5}/\textbf{69.7}/0.0 & \cellcolor[HTML]{EAECFF}\textbf{73.6}/\textbf{71.1}/0.0 \\
    \bottomrule
\end{tabular}}
\caption{HLT-vs-HWT binary results by modification level (Proofread $\to$ Polish $\to$ Paraphrase; increasing LLM intervention): AUROC/F1/TPR@FPR{=}1\% (\%; averaged over 5 generators, and 5 seeds for seeded methods). For each level, the negative class is the same HWT set and the positive class is restricted to HLT texts of that level. Per column and metric, best is \textbf{bolded}.}
\label{tab:app_hltlevel}
\end{table*}

\section{Switching the Training Generator}
\label{sec:app_trainsource}
Tables~\ref{tab:app_train_gpt} and~\ref{tab:app_train_qwen} restore the full main binary table with the training generator switched from Claude-Sonnet-4.6 to GPT-4o and Qwen2.5-32B, respectively. The test pool is unchanged (the same five generator test splits), so zero-shot rows are identical to Table~\ref{tab:exp_binary}; only training-dependent methods ($\dagger$) change.

\begin{table*}[t]
\centering
\resizebox{\textwidth}{!}{
\setlength{\tabcolsep}{2.5pt}
\begin{tabular}{l|ccc|ccc|ccc|ccc|ccc|ccc|ccc|ccc|ccc}
    \toprule
    \multirow{3}{*}{\textbf{Method}} & \multicolumn{12}{c|}{\textbf{HLT vs.\ HWT}} & \multicolumn{12}{c|}{\textbf{LGT vs.\ HWT}} & \multicolumn{3}{c}{\multirow{2}{*}{\textbf{Avg.}}} \\
    \cmidrule(lr){2-13} \cmidrule(lr){14-25}
    & \multicolumn{3}{c|}{Essay} & \multicolumn{3}{c|}{Paper Abs.} & \multicolumn{3}{c|}{Editorial} & \multicolumn{3}{c|}{Avg.} & \multicolumn{3}{c|}{Essay} & \multicolumn{3}{c|}{Paper Abs.} & \multicolumn{3}{c|}{Editorial} & \multicolumn{3}{c|}{Avg.} & \multicolumn{3}{c}{} \\
    \cmidrule(lr){2-4} \cmidrule(lr){5-7} \cmidrule(lr){8-10} \cmidrule(lr){11-13} \cmidrule(lr){14-16} \cmidrule(lr){17-19} \cmidrule(lr){20-22} \cmidrule(lr){23-25} \cmidrule(lr){26-28}
    & AUC & F1 & T@1 & AUC & F1 & T@1 & AUC & F1 & T@1 & AUC & F1 & T@1 & AUC & F1 & T@1 & AUC & F1 & T@1 & AUC & F1 & T@1 & AUC & F1 & T@1 & AUC & F1 & T@1 \\
    \midrule
    \rowcolor{gray!15} \multicolumn{28}{l}{\textit{Confidence-based}} \\
    ~~Log-Likelihood & 54.0 & 52.4 & 1.1 & 56.0 & 58.2 & 5.7 & 52.1 & 53.5 & 2.0 & 54.0 & 54.7 & 2.9 & 47.9 & 51.2 & 1.0 & 55.1 & 60.4 & 6.4 & 52.1 & 59.1 & 4.6 & 51.7 & 56.9 & 4.0 & 52.9 & 55.8 & 3.5 \\
    ~~Rank & 54.9 & 52.5 & 0.7 & 56.6 & 56.7 & 6.5 & 52.1 & 53.6 & 2.4 & 54.5 & 54.3 & 3.2 & 48.5 & 51.1 & 1.7 & 54.8 & 58.5 & 11.1 & 49.2 & 56.6 & 2.4 & 50.8 & 55.4 & 5.1 & 52.7 & 54.8 & 4.1 \\
    ~~Log-Rank & 54.1 & 52.8 & 1.0 & 55.6 & 58.3 & 5.7 & 52.2 & 53.6 & 1.8 & 54.0 & 54.9 & 2.8 & 48.0 & 50.6 & 1.5 & 55.9 & 64.0 & 6.4 & 50.5 & 58.1 & 2.9 & 51.5 & 57.6 & 3.6 & 52.7 & 56.2 & 3.2 \\
    ~~Entropy & 52.5 & 51.3 & 1.4 & 49.5 & 53.3 & 1.7 & 51.8 & 52.6 & 1.9 & 51.3 & 52.4 & 1.7 & 49.5 & 52.3 & 2.6 & 49.6 & 54.7 & 3.8 & 45.0 & 53.9 & 0.9 & 48.0 & 53.6 & 2.4 & 49.6 & 53.0 & 2.0 \\
    ~~LRR & 54.0 & 53.5 & 4.0 & 52.2 & 56.7 & 11.9 & 52.1 & 52.5 & 2.5 & 52.8 & 54.2 & 6.1 & 47.0 & 51.1 & 4.8 & 54.7 & 58.6 & 11.3 & 43.7 & 51.5 & 4.6 & 48.5 & 53.7 & 6.9 & 50.6 & 54.0 & 6.5 \\
    \rowcolor{gray!15} \multicolumn{28}{l}{\textit{Perturbation}} \\
    ~~NPR & 71.9 & 65.3 & 3.7 & 57.9 & 60.7 & 10.7 & 48.7 & 51.4 & 2.3 & 59.5 & 59.1 & 5.6 & 74.5 & 68.9 & 10.4 & 43.0 & 53.4 & 2.8 & 81.9 & 76.5 & 28.4 & 66.5 & 66.3 & 13.9 & 63.0 & 62.7 & 9.7 \\
    ~~DetectGPT & 62.2 & 58.6 & 4.0 & 52.7 & 54.7 & 5.8 & 52.7 & 53.5 & 0.9 & 55.9 & 55.6 & 3.6 & 59.1 & 57.6 & 8.1 & 49.7 & 58.0 & 2.6 & 82.6 & 79.6 & 32.9 & 63.8 & 65.1 & 14.5 & 59.8 & 60.3 & 9.0 \\
    ~~Fast-DetectGPT & 52.5 & 52.3 & 3.7 & 45.4 & 50.0 & 2.7 & 56.1 & 55.4 & 1.1 & 51.3 & 52.6 & 2.5 & 64.5 & 60.8 & 11.3 & 64.8 & 67.1 & 3.4 & 91.2 & 87.7 & 37.7 & 73.5 & 71.9 & 17.5 & 62.4 & 62.2 & 10.0 \\
    \rowcolor{gray!15} \multicolumn{28}{l}{\textit{LLM-based}} \\
    ~~Binoculars & 71.9 & 65.0 & 7.1 & 53.4 & 59.0 & 9.0 & 54.2 & 53.3 & 0.5 & 59.8 & 59.1 & 5.5 & 88.6 & 83.3 & 38.2 & 48.9 & 57.2 & 10.0 & 80.6 & 80.7 & 36.0 & 72.7 & 73.7 & 28.1 & 66.3 & 66.4 & 16.8 \\
    ~~LLM-Det (0-shot) & 60.0 & 58.8 & 1.8 & 50.8 & 51.4 & 0.4 & 56.5 & 54.3 & 4.0 & 55.8 & 54.8 & 2.1 & 67.9 & 63.1 & 4.0 & 57.0 & 58.5 & 2.0 & 67.8 & 66.3 & 15.8 & 64.2 & 62.6 & 7.3 & 60.0 & 58.7 & 4.7 \\
    ~~LLM-Det (10-shot)$^\dagger$ & 49.9 & 49.3 & 1.1 & 51.7 & 54.6 & 2.5 & 51.8 & 52.2 & 2.5 & 51.1 & 52.0 & 2.0 & 59.7 & 57.9 & 2.9 & 50.8 & 55.9 & 1.8 & 53.0 & 57.3 & 2.6 & 54.5 & 57.0 & 2.4 & 52.8 & 54.5 & 2.2 \\
    \rowcolor{gray!15} \multicolumn{28}{l}{\textit{Grammar-based}} \\
    ~~GECScore & 50.5 & 51.1 & 0.8 & 52.5 & 56.5 & 9.4 & 54.5 & 53.9 & 1.1 & 52.5 & 53.8 & 3.8 & 49.8 & 51.0 & 1.2 & 40.6 & 49.6 & 8.0 & 56.8 & 58.5 & 2.2 & 49.1 & 53.0 & 3.8 & 50.8 & 53.4 & 3.8 \\
    \rowcolor{gray!15} \multicolumn{28}{l}{\textit{Neural-based}} \\
    ~~XDAC-D & 65.5 & 61.9 & 0.0 & 63.3 & 62.9 & 0.0 & 59.6 & 58.5 & 0.0 & 62.8 & 61.1 & 0.0 & 68.5 & 69.4 & 0.0 & 58.8 & 58.0 & 0.0 & 67.7 & 69.1 & 0.0 & 65.0 & 65.5 & 0.0 & 63.9 & 63.3 & 0.0 \\
    ~~RepreGuard$^\dagger$ & 82.2 & 73.9 & \textbf{15.9} & 56.1 & 59.5 & 3.3 & 66.8 & 60.6 & \textbf{12.2} & 68.4 & 64.7 & 10.5 & \textbf{90.7} & \textbf{84.7} & \textbf{45.0} & 70.7 & 69.6 & 5.0 & \textbf{98.5} & 95.8 & \textbf{90.3} & 86.6 & 83.4 & \textbf{46.8} & 77.5 & 74.0 & \textbf{28.6} \\
    \rowcolor{gray!15} \multicolumn{28}{l}{\textit{Linguistic-Feature}} \\
    ~~KatFishNet$^\dagger$ & 58.5 & 55.5 & 1.0 & 50.0 & 42.6 & 0.0 & 50.3 & 50.3 & 0.8 & 52.9 & 49.5 & 0.6 & 74.3 & 67.0 & 8.4 & 60.3 & 60.8 & 8.5 & 85.5 & 80.3 & 19.5 & 73.4 & 69.4 & 12.1 & 63.2 & 59.4 & 6.4 \\
    \cellcolor[HTML]{EAECFF}\textbf{\textsc{KoLingProbe}}$^\dagger$ & \cellcolor[HTML]{EAECFF}\textbf{86.2} & \cellcolor[HTML]{EAECFF}\textbf{77.5} & \cellcolor[HTML]{EAECFF}15.1 & \cellcolor[HTML]{EAECFF}\textbf{68.7} & \cellcolor[HTML]{EAECFF}\textbf{65.0} & \cellcolor[HTML]{EAECFF}\textbf{15.5} & \cellcolor[HTML]{EAECFF}\textbf{72.7} & \cellcolor[HTML]{EAECFF}\textbf{68.6} & \cellcolor[HTML]{EAECFF}1.8 & \cellcolor[HTML]{EAECFF}\textbf{75.9} & \cellcolor[HTML]{EAECFF}\textbf{70.4} & \cellcolor[HTML]{EAECFF}\textbf{10.8} & \cellcolor[HTML]{EAECFF}90.3 & \cellcolor[HTML]{EAECFF}84.3 & \cellcolor[HTML]{EAECFF}39.9 & \cellcolor[HTML]{EAECFF}\textbf{84.6} & \cellcolor[HTML]{EAECFF}\textbf{84.6} & \cellcolor[HTML]{EAECFF}\textbf{11.8} & \cellcolor[HTML]{EAECFF}98.1 & \cellcolor[HTML]{EAECFF}\textbf{96.1} & \cellcolor[HTML]{EAECFF}81.2 & \cellcolor[HTML]{EAECFF}\textbf{91.0} & \cellcolor[HTML]{EAECFF}\textbf{88.3} & \cellcolor[HTML]{EAECFF}44.3 & \cellcolor[HTML]{EAECFF}\textbf{83.4} & \cellcolor[HTML]{EAECFF}\textbf{79.4} & \cellcolor[HTML]{EAECFF}27.6 \\
    \bottomrule
\end{tabular}}
\caption{Binary detection with the \textbf{training generator switched to GPT-4o} (instead of Claude-Sonnet-4.6 used in the main table): AUROC (AUC), F1, and TPR@FPR{=}1\% (T@1), in \%, averaged over the same five generator test splits as the main table. Methods marked $^\dagger$ depend on task training and are re-trained on the new source; all other methods are zero-shot, so their scores are identical to Table~\ref{tab:exp_binary} and are repeated for completeness. Per-column best is \textbf{bolded}.}
\label{tab:app_train_gpt}
\end{table*}

\begin{table*}[t]
\centering
\resizebox{\textwidth}{!}{
\setlength{\tabcolsep}{2.5pt}
\begin{tabular}{l|ccc|ccc|ccc|ccc|ccc|ccc|ccc|ccc|ccc}
    \toprule
    \multirow{3}{*}{\textbf{Method}} & \multicolumn{12}{c|}{\textbf{HLT vs.\ HWT}} & \multicolumn{12}{c|}{\textbf{LGT vs.\ HWT}} & \multicolumn{3}{c}{\multirow{2}{*}{\textbf{Avg.}}} \\
    \cmidrule(lr){2-13} \cmidrule(lr){14-25}
    & \multicolumn{3}{c|}{Essay} & \multicolumn{3}{c|}{Paper Abs.} & \multicolumn{3}{c|}{Editorial} & \multicolumn{3}{c|}{Avg.} & \multicolumn{3}{c|}{Essay} & \multicolumn{3}{c|}{Paper Abs.} & \multicolumn{3}{c|}{Editorial} & \multicolumn{3}{c|}{Avg.} & \multicolumn{3}{c}{} \\
    \cmidrule(lr){2-4} \cmidrule(lr){5-7} \cmidrule(lr){8-10} \cmidrule(lr){11-13} \cmidrule(lr){14-16} \cmidrule(lr){17-19} \cmidrule(lr){20-22} \cmidrule(lr){23-25} \cmidrule(lr){26-28}
    & AUC & F1 & T@1 & AUC & F1 & T@1 & AUC & F1 & T@1 & AUC & F1 & T@1 & AUC & F1 & T@1 & AUC & F1 & T@1 & AUC & F1 & T@1 & AUC & F1 & T@1 & AUC & F1 & T@1 \\
    \midrule
    \rowcolor{gray!15} \multicolumn{28}{l}{\textit{Confidence-based}} \\
    ~~Log-Likelihood & 54.0 & 52.4 & 1.1 & 56.0 & 58.2 & 5.7 & 52.1 & 53.5 & 2.0 & 54.0 & 54.7 & 2.9 & 47.9 & 51.2 & 1.0 & 55.1 & 60.4 & 6.4 & 52.1 & 59.1 & 4.6 & 51.7 & 56.9 & 4.0 & 52.9 & 55.8 & 3.5 \\
    ~~Rank & 54.9 & 52.5 & 0.7 & 56.6 & 56.7 & 6.5 & 52.1 & 53.6 & 2.4 & 54.5 & 54.3 & 3.2 & 48.5 & 51.1 & 1.7 & 54.8 & 58.5 & 11.1 & 49.2 & 56.6 & 2.4 & 50.8 & 55.4 & 5.1 & 52.7 & 54.8 & 4.1 \\
    ~~Log-Rank & 54.1 & 52.8 & 1.0 & 55.6 & 58.3 & 5.7 & 52.2 & 53.6 & 1.8 & 54.0 & 54.9 & 2.8 & 48.0 & 50.6 & 1.5 & 55.9 & 64.0 & 6.4 & 50.5 & 58.1 & 2.9 & 51.5 & 57.6 & 3.6 & 52.7 & 56.2 & 3.2 \\
    ~~Entropy & 52.5 & 51.3 & 1.4 & 49.5 & 53.3 & 1.7 & 51.8 & 52.6 & 1.9 & 51.3 & 52.4 & 1.7 & 49.5 & 52.3 & 2.6 & 49.6 & 54.7 & 3.8 & 45.0 & 53.9 & 0.9 & 48.0 & 53.6 & 2.4 & 49.6 & 53.0 & 2.0 \\
    ~~LRR & 54.0 & 53.5 & 4.0 & 52.2 & 56.7 & \textbf{11.9} & 52.1 & 52.5 & 2.5 & 52.8 & 54.2 & 6.1 & 47.0 & 51.1 & 4.8 & 54.7 & 58.6 & 11.3 & 43.7 & 51.5 & 4.6 & 48.5 & 53.7 & 6.9 & 50.6 & 54.0 & 6.5 \\
    \rowcolor{gray!15} \multicolumn{28}{l}{\textit{Perturbation}} \\
    ~~NPR & 71.9 & 65.3 & 3.7 & 57.9 & 60.7 & 10.7 & 48.7 & 51.4 & 2.3 & 59.5 & 59.1 & 5.6 & 74.5 & 68.9 & 10.4 & 43.0 & 53.4 & 2.8 & 81.9 & 76.5 & 28.4 & 66.5 & 66.3 & 13.9 & 63.0 & 62.7 & 9.7 \\
    ~~DetectGPT & 62.2 & 58.6 & 4.0 & 52.7 & 54.7 & 5.8 & 52.7 & 53.5 & 0.9 & 55.9 & 55.6 & 3.6 & 59.1 & 57.6 & 8.1 & 49.7 & 58.0 & 2.6 & 82.6 & 79.6 & 32.9 & 63.8 & 65.1 & 14.5 & 59.8 & 60.3 & 9.0 \\
    ~~Fast-DetectGPT & 52.5 & 52.3 & 3.7 & 45.4 & 50.0 & 2.7 & 56.1 & 55.4 & 1.1 & 51.3 & 52.6 & 2.5 & 64.5 & 60.8 & 11.3 & 64.8 & 67.1 & 3.4 & 91.2 & 87.7 & 37.7 & 73.5 & 71.9 & 17.5 & 62.4 & 62.2 & 10.0 \\
    \rowcolor{gray!15} \multicolumn{28}{l}{\textit{LLM-based}} \\
    ~~Binoculars & 71.9 & 65.0 & 7.1 & 53.4 & 59.0 & 9.0 & 54.2 & 53.3 & 0.5 & 59.8 & 59.1 & 5.5 & 88.6 & 83.3 & 38.2 & 48.9 & 57.2 & 10.0 & 80.6 & 80.7 & 36.0 & 72.7 & 73.7 & 28.1 & 66.3 & 66.4 & 16.8 \\
    ~~LLM-Det (0-shot) & 60.0 & 58.8 & 1.8 & 50.8 & 51.4 & 0.4 & 56.5 & 54.3 & \textbf{4.0} & 55.8 & 54.8 & 2.1 & 67.9 & 63.1 & 4.0 & 57.0 & 58.5 & 2.0 & 67.8 & 66.3 & 15.8 & 64.2 & 62.6 & 7.3 & 60.0 & 58.7 & 4.7 \\
    ~~LLM-Det (10-shot)$^\dagger$ & 50.2 & 50.4 & 1.1 & 50.6 & 56.3 & 3.3 & 48.6 & 50.1 & 1.5 & 49.8 & 52.3 & 2.0 & 52.2 & 53.4 & 1.0 & 51.5 & 58.6 & 5.5 & 46.1 & 51.4 & 2.4 & 49.9 & 54.5 & 3.0 & 49.9 & 53.4 & 2.5 \\
    \rowcolor{gray!15} \multicolumn{28}{l}{\textit{Grammar-based}} \\
    ~~GECScore & 50.5 & 51.1 & 0.8 & 52.5 & 56.5 & 9.4 & 54.5 & 53.9 & 1.1 & 52.5 & 53.8 & 3.8 & 49.8 & 51.0 & 1.2 & 40.6 & 49.6 & 8.0 & 56.8 & 58.5 & 2.2 & 49.1 & 53.0 & 3.8 & 50.8 & 53.4 & 3.8 \\
    \rowcolor{gray!15} \multicolumn{28}{l}{\textit{Neural-based}} \\
    ~~XDAC-D & 65.5 & 61.9 & 0.0 & 63.3 & 62.9 & 0.0 & 59.6 & 58.5 & 0.0 & 62.8 & 61.1 & 0.0 & 68.5 & 69.4 & 0.0 & 58.8 & 58.0 & 0.0 & 67.7 & 69.1 & 0.0 & 65.0 & 65.5 & 0.0 & 63.9 & 63.3 & 0.0 \\
    ~~RepreGuard$^\dagger$ & 83.2 & 73.5 & \textbf{16.8} & 60.7 & 60.4 & 0.0 & 67.7 & 68.0 & 2.2 & 70.5 & 67.3 & 6.3 & \textbf{93.7} & \textbf{87.8} & \textbf{48.2} & \textbf{70.7} & \textbf{72.5} & 6.4 & \textbf{97.1} & \textbf{95.6} & 62.0 & \textbf{87.2} & \textbf{85.3} & 38.9 & 78.9 & 76.3 & 22.6 \\
    \rowcolor{gray!15} \multicolumn{28}{l}{\textit{Linguistic-Feature}} \\
    ~~KatFishNet$^\dagger$ & 58.3 & 55.3 & 1.6 & 49.6 & 53.5 & 2.9 & 53.0 & 52.1 & 0.1 & 53.6 & 53.6 & 1.5 & 74.8 & 67.5 & 7.2 & 61.7 & 63.5 & 10.7 & 81.0 & 75.1 & 25.3 & 72.5 & 68.7 & 14.4 & 63.1 & 61.2 & 8.0 \\
    \cellcolor[HTML]{EAECFF}\textbf{\textsc{KoLingProbe}}$^\dagger$ & \cellcolor[HTML]{EAECFF}\textbf{85.5} & \cellcolor[HTML]{EAECFF}\textbf{77.8} & \cellcolor[HTML]{EAECFF}13.6 & \cellcolor[HTML]{EAECFF}\textbf{68.7} & \cellcolor[HTML]{EAECFF}\textbf{64.1} & \cellcolor[HTML]{EAECFF}7.9 & \cellcolor[HTML]{EAECFF}\textbf{70.5} & \cellcolor[HTML]{EAECFF}\textbf{70.7} & \cellcolor[HTML]{EAECFF}0.4 & \cellcolor[HTML]{EAECFF}\textbf{74.9} & \cellcolor[HTML]{EAECFF}\textbf{70.9} & \cellcolor[HTML]{EAECFF}\textbf{7.3} & \cellcolor[HTML]{EAECFF}92.0 & \cellcolor[HTML]{EAECFF}85.2 & \cellcolor[HTML]{EAECFF}33.4 & \cellcolor[HTML]{EAECFF}66.9 & \cellcolor[HTML]{EAECFF}69.1 & \cellcolor[HTML]{EAECFF}\textbf{19.1} & \cellcolor[HTML]{EAECFF}96.8 & \cellcolor[HTML]{EAECFF}94.3 & \cellcolor[HTML]{EAECFF}\textbf{79.0} & \cellcolor[HTML]{EAECFF}85.2 & \cellcolor[HTML]{EAECFF}82.9 & \cellcolor[HTML]{EAECFF}\textbf{43.8} & \cellcolor[HTML]{EAECFF}\textbf{80.1} & \cellcolor[HTML]{EAECFF}\textbf{76.9} & \cellcolor[HTML]{EAECFF}\textbf{25.6} \\
    \bottomrule
\end{tabular}}
\caption{As Table~\ref{tab:app_train_gpt}, with the training generator switched to \textbf{Qwen2.5-32B}.}
\label{tab:app_train_qwen}
\end{table*}

\section{Cross-Domain Transfer of All Training-Dependent Detectors}
\label{app:cross_domain_full}
Table~\ref{tab:cross_domain_all} reports the full cross-domain transfer protocol of \S\ref{sec:transfer}: each method is trained on one domain with its complete original pipeline (KatFishNet and \textsc{KoLingProbe}: stability selection and stacking; RepreGuard: layer selection and linear probes; XLM-RoBERTa: full fine-tuning) and evaluated on all three domains' test splits.

\begin{table*}[t!]
\centering
\resizebox{0.92\textwidth}{!}{
\setlength{\tabcolsep}{4.5pt}
\begin{tabular}{ll|ccc|ccc}
    \toprule
    \multirow{2.5}{*}{\textbf{Method}} & \multirow{2.5}{*}{\textbf{Train}} & \multicolumn{3}{c|}{\textbf{HLT vs.\ HWT (test)}} & \multicolumn{3}{c}{\textbf{LGT vs.\ HWT (test)}} \\
    \cmidrule(lr){3-5} \cmidrule(lr){6-8}
    & & Essay & Paper Abs. & Editorial & Essay & Paper Abs. & Editorial \\
    \midrule
    \multirow{3}{*}{KatFishNet} & Essay & \cellcolor{gray!12}60.9 & 55.7 & 55.4 & \cellcolor{gray!12}73.7 & 57.3 & 89.6 \\
     & Paper Abs. & 50.0 & \cellcolor{gray!12}50.0 & 50.0 & 55.9 & \cellcolor{gray!12}53.1 & 56.4 \\
     & Editorial & 48.5 & 54.0 & \cellcolor{gray!12}50.0 & 73.7 & 59.3 & \cellcolor{gray!12}91.9 \\
    \midrule
    \multirow{3}{*}{RepreGuard} & Essay & \cellcolor{gray!12}81.6 & 62.7 & 55.8 & \cellcolor{gray!12}89.3 & 65.4 & 89.4 \\
     & Paper Abs. & 64.9 & \cellcolor{gray!12}66.7 & 51.0 & 55.4 & \cellcolor{gray!12}69.4 & 57.7 \\
     & Editorial & 61.3 & 49.7 & \cellcolor{gray!12}68.0 & 68.3 & 54.3 & \cellcolor{gray!12}95.3 \\
    \midrule
    \multirow{3}{*}{XLM-RoBERTa (FT)} & Essay & \cellcolor{gray!12}96.8 & 70.8 & 58.2 & \cellcolor{gray!12}99.5 & 83.0 & 89.2 \\
     & Paper Abs. & 60.7 & \cellcolor{gray!12}53.2 & 52.8 & 70.1 & \cellcolor{gray!12}73.6 & 47.1 \\
     & Editorial & 64.5 & 53.0 & \cellcolor{gray!12}62.0 & 77.7 & 45.5 & \cellcolor{gray!12}99.1 \\
    \midrule
    \multirow{3}{*}{\textsc{KoLingProbe}} & Essay & \cellcolor{gray!12}85.1 & 67.3 & 55.1 & \cellcolor{gray!12}91.2 & 65.9 & 86.9 \\
     & Paper Abs. & 69.3 & \cellcolor{gray!12}66.9 & 53.4 & 69.4 & \cellcolor{gray!12}74.0 & 33.1 \\
     & Editorial & 63.2 & 55.0 & \cellcolor{gray!12}70.1 & 85.0 & 58.1 & \cellcolor{gray!12}99.6 \\
    \bottomrule
\end{tabular}}
\caption{Cross-domain transfer of all training-dependent detectors (AUROC, \%): every detector is trained entirely on the row domain with its full original protocol and evaluated on the column domain's test splits (5 generators $\times$ 5 seeds). Shaded cells are the in-domain reference. For academic-HLT, KatFishNet's stability selection retains no comma feature; following the main-table convention, all features are used as a fallback, yielding chance-level scores.}
\label{tab:cross_domain_all}
\end{table*}

\end{CJK}
\end{document}